%% file: IEEE_jrnl.tex
\let\TeXyear\year
\documentclass{IEEEaccess}
\let\setyear\year
\let\year\TeXyear

\usepackage{microtype}
\usepackage{etoolbox} 

\patchcmd{\appendices}
{\thesectiondis{APPENDIX \Alph{section}}.}
{\thesectiondis{APPENDIX \Alph{section}.}}
{}{}

\setyear{2021}
\vol{V}
\makeatletter
\patchcmd{\ps@titlepage}{2016}{\theyear}{}{}
\patchcmd{\ps@headings}{2016}{\theyear}{}{}
\patchcmd{\ps@headings}{2016}{\theyear}{}{}
\ps@headings
\makeatother

\usepackage{cite}

\usepackage{xr}
\usepackage{graphicx}
\graphicspath{{./img/}}
\DeclareGraphicsExtensions{.pdf,.jpeg,.jpg,.png,.tex,.eps}

\usepackage[labelformat=simple]{subcaption}
\usepackage{tikz}
\NewSpotColorSpace{PANTONE}
\AddSpotColor{PANTONE} {PANTONE3015C} {PANTONE\SpotSpace 3015\SpotSpace C} {1 0.3 0 0.2}
\SetPageColorSpace{PANTONE}%

\usetikzlibrary{%
  calc,
  shapes,
  positioning,
  fit,
  scopes,
  intersections,
  3d,
  plotmarks,
  external,
  arrows.meta,
}
\tikzset{external/only named=true}

\usepackage{pgfplots}
\usepgfplotslibrary{%
  colorbrewer,
  external,
  fillbetween,
  groupplots,
}

\tikzset{%
}

\pgfplotsset{
  compat=1.17,%
  width=\linewidth,height=0.85\linewidth,%
  every axis plot post/.append style={thick},%
  ymajorgrids,
  grid style={dashed}, %
  minimal plot grid/.style={
    y axis line style={opacity=0},
    axis x line*=bottom,
    x axis line style={black},
  },
  minimal plot grid,
  /pgfplots/legend pos/south center/.style={/pgfplots/legend style={at={(0.5,0.03)},anchor=south}},
  /pgfplots/legend pos/north center/.style={/pgfplots/legend style={at={(0.5,0.97)},anchor=north}},
  /pgfplots/legend pos/outer north center/.style={/pgfplots/legend style={at={(0.5,1.02)},anchor=south}},
  y tick label style={
    /pgf/number format/.cd,
    fixed,
    fixed zerofill,
    precision=1,
    /tikz/.cd,
    font=\footnotesize,
  },
  x tick label style={
    /pgf/number format/.cd,
    fixed,
    fixed zerofill,
    precision=1,
    /tikz/.cd,
    font=\footnotesize,
  },
}

\usepackage{amsmath}
\usepackage{amssymb} %
\usepackage{amsfonts}

\let\oldtimes\times
\def\times{{\mkern1mu\oldtimes\mkern1mu}}

\let\given\givenbase

\usepackage{array}
\usepackage{multirow}
\usepackage{booktabs}
\usepackage{siunitx}
\usepackage{xspace}
\makeatletter
\DeclareRobustCommand\onedot{\futurelet\@let@token\@onedot}
\def\@onedot{\ifx\@let@token.\else.\null\fi\xspace}

\def\eg{{e.g}\onedot} \def\Eg{{E.g}\onedot}
\def\ie{{i.e}\onedot} 
\def\cf{{cf}\onedot} 
 \def\vs{{vs}\onedot}

\def\etal{{et al}\onedot}

\usepackage{tkz-kiviat}
\makeatletter
\def\tkz@KiviatGrad[#1](#2){%
  \begingroup
  \pgfkeys{/kiviatgrad/.cd,
    graduation distance= 0 pt,
    prefix ={},
    suffix={},
    unity=1,
    label precision/.store in=\gradlabel@precision,
    label precision=1,
    zero point/.store in=\tkz@grad@zero,
    zero point=0
  }
  \pgfqkeys{/kiviatgrad}{#1}%
  \let\tikz@label@distance@tmp\tikz@label@distance
  \global\let\tikz@label@distance\tkz@kiv@grad
  \foreach \nv in {0,...,\tkz@kiv@lattice}{
    \pgfmathsetmacro{\result}{\tkz@kiv@unity*\nv-\tkz@grad@zero} %
    \protected@edef\tkz@kiv@gd{%
      \tkz@kiv@prefix%
      \pgfmathprintnumber[precision=\gradlabel@precision,fixed]{\result}%
      \tkz@kiv@suffix} 
    \path[/kiviatgrad/.cd,#1] (0:0)--(360/\tkz@kiv@radial*#2:\nv*\tkz@kiv@gap)
    node[label=(360/\tkz@kiv@radial*#2):\scriptsize\tkz@kiv@gd] {}; %
  }
  \let\tikz@label@distance\tikz@label@distance@tmp  
  \endgroup
}%

\def\tkz@KiviatLine[#1](#2,#3){%
  \begingroup
  \pgfkeys{/kiviatline/.cd,
    fill= {},
    opacity=.5,
    zero point/.store in=\tkz@line@zero,
    zero point=0
  }
  \pgfqkeys{/kiviatline}{#1}%
  \ifx\tkzutil@empty\tkz@kivl@fill \else 
  \begin{scope}[on background layer]
    \path[fill=\tkz@kivl@fill,opacity=\tkz@kivl@opacity] (360/\tkz@kiv@radial*0:{(#2+\tkz@line@zero)*\tkz@kiv@gap*\tkz@kiv@step})   
    \foreach \v [count=\rang from 1] in {#3}{%
      -- (360/\tkz@kiv@radial*\rang:{(\v+\tkz@line@zero)*\tkz@kiv@gap*\tkz@kiv@step}) } -- (360/\tkz@kiv@radial*0:{(#2+\tkz@line@zero)*\tkz@kiv@gap*\tkz@kiv@step}); 
  \end{scope}
  \fi       
  \draw[#1,opacity=1,overlay] (0:{(#2+\tkz@line@zero)*\tkz@kiv@gap}) plot coordinates {(360/\tkz@kiv@radial*0:{(#2+\tkz@line@zero)*\tkz@kiv@gap*\tkz@kiv@step})}  
  \foreach \v [count=\rang from 1] in {#3}{%
    -- (360/\tkz@kiv@radial*\rang:{(\v+\tkz@line@zero)*\tkz@kiv@gap*\tkz@kiv@step}) plot coordinates {(360/\tkz@kiv@radial*\rang:{(\v+\tkz@line@zero)*\tkz@kiv@gap*\tkz@kiv@step})}} -- (360/\tkz@kiv@radial*0:{(#2+\tkz@line@zero)*\tkz@kiv@gap*\tkz@kiv@step});   
  \endgroup
}%
\makeatother

\usepackage[]{hyperref}  %
\hypersetup{
  colorlinks,
  linkcolor = red!50!black,
  citecolor = blue!80!black,%
  urlcolor  = magenta!75!black,
}

\begin{document}
\history{Pre-print to appear in IEEE Access}
\doi{10.1109/ACCESS.2021.3085218}

\title{Empirical Study of Multi-Task Hourglass Model for Semantic Segmentation Task}
\author{\uppercase{Darwin Saire,\authorrefmark{1} and Ad\'in Ram\'irez~Rivera,\authorrefmark{2}}  
  \IEEEmembership{Member, IEEE}}
\address[1]{Institute of Computing, University of Campinas, Brazil (e-mail: \texttt{darwin.pilco@ic.unicamp.br})}
\address[2]{Institute of Computing, University of Campinas, Brazil (e-mail: \texttt{adin@ic.unicamp.br})}
\tfootnote{This work was financed in part by the S{\~a}o Paulo Research Foundation (FAPESP) under grants No.~2017/16597-7, 2019/07257-3, and~2019/18678-0, and in part by the Brazilian National Council for Scientific and Technological Development (CNPq) under grant No.~307425/2017-7.  The Laboratoire Lorrain de Recherche en Informatique et ses Applications (LORIA) institute provided, in part, infrastructure for this work.
\newline
Code available at \url{https://gitlab.com/mipl/mtl-ss}.
}

\markboth{Saire and Ram\'irez~Rivera: Empirical Study of Multi-Task Hourglass Model for Semantic Segmentation Task}%
{Saire and Ram\'irez~Rivera: Empirical Study of Multi-Task Hourglass Model for Semantic Segmentation Task}

\corresp{Corresponding author: Ad\'in Ram\'irez~Rivera (e-mail: \texttt{adin@ic.unicamp.br}).}
  

\titlepgskip=-15pt

\begin{abstract}
The semantic segmentation~(SS) task aims to create a dense classification by labeling at the pixel level each object present on images. 
Convolutional neural network~(CNN) approaches have been widely used, and exhibited the best results in this task. 
However, the loss of spatial precision on the results is a main drawback that has not been solved.
In this work, we propose to use a multi-task approach by complementing the semantic segmentation task with edge detection, semantic contour, and distance transform tasks. 
We propose that by sharing a common latent space, the complementary tasks can produce more robust representations that can enhance the semantic labels. 
We explore the influence of contour-based tasks on latent space, as well as their impact on the final results of SS\@.
We demonstrate the effectiveness of learning in a multi-task setting for hourglass models in the Cityscapes, CamVid, and Freiburg Forest datasets by improving the state-of-the-art without any refinement post-processing.
\end{abstract}

\maketitle

\section{Introduction}
\label{sec:introduction}

\IEEEPARstart{H}{umans} possess a remarkable ability to parse images simply by looking at them.
In a blink of an eye, a human can fully analyze an image and separate all its components. 
People can perform several tasks simultaneously by analyzing an image, \eg, object detection and contour detection. 
Furthermore, humans can easily generalize from observing a set of objects to recognizing objects that have never been seen before. 
Although humans enjoy an inherent capacity for generalization, they lack the processing power given by computers. 
That is, to process a large amount of information (\eg, images) in a reduced interval of time.
The separation of an image into its components (\ie, join pixels into regions) according to some features is called image segmentation~\cite{Gonzalez2006}.
Reproducing this process at or above the human level on a computer is not an easy task, and several approaches have been proposed to address it~\cite{Chouhan2019}. 
Nevertheless, the segmentation task continues to be challenging mainly due to variability, \ie, when the visual tasks are performed on a computer there is a considerable variation in pose, appearance, viewpoint, illumination, and occlusion throughout different instances of the same image.
Thus, a type of segmentation commonly used is semantic segmentation~(SS). 
SS is an essential part of the pipeline in computer vision projects. 
It extracts and analyzes useful and meaningful information, in addition to classifying the regions obtained within an image. 
In other words, by improving the segmentation stage, the computer vision's final output is also enhanced.

\begin{figure}[tb]
  \centering
  \subfloat[SegNet]{\includegraphics[width=.33\linewidth,height=0.7in]{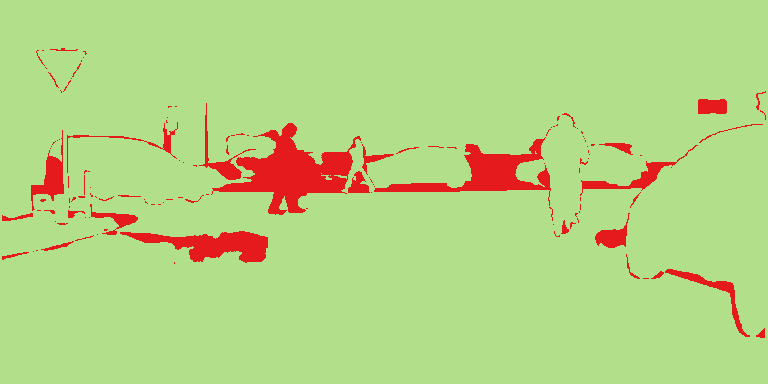}%
    \label{fig:prob_SegNet}}%
  \subfloat[AdapNet++]{\includegraphics[width=.33\linewidth,height=0.7in]{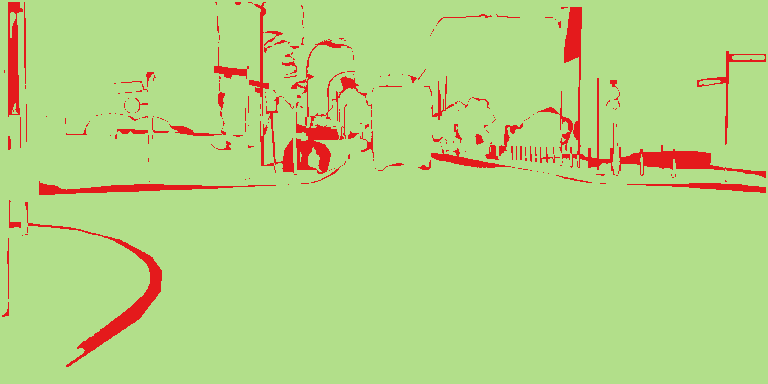}%
    \label{fig:prob_AdapNet}}%
  \subfloat[FastNet]{\includegraphics[width=.33\linewidth,height=0.7in]{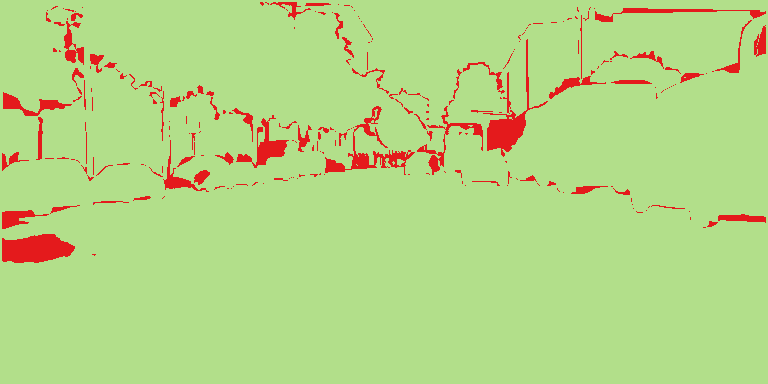}%
    \label{fig:prob_FastNet}}%
  \caption{The main problem of current segmentation methods lie in the loss of spatial precision at the boundaries or small objects. 
    Green and red regions denote correctly and incorrectly segmented regions, respectively.}
  \label{fig:SS_loss_spatial_precision}
\end{figure}

In recent years, Convolutional Neural Networks~(CNNs) have led to several improvements in computer vision.
Fully convolutional networks (FCN)~\cite{Long2016} achieved a significant improvement in the SS task in contrast to the traditional SS techniques~\cite{Zaitoun2015}. 
However, (i)~the low-resolution at the CNNs output and (ii)~the loss of spatial precision of objects within the image are still the main problems that affect the segmentation results~\cite{Chen2017a, Lin2017}, see Fig.~\ref{fig:SS_loss_spatial_precision}.
We believe that these problems are not caused by a specific operation (\eg, down-sampling) but by a set of factors. 
For instance, the absence of reconstruction and refinement methods, the excessive down-sampling, the gradient vanishing problem, or lack of a better extractor of feature maps.

Nowadays, different models~\cite{Garcia-Garcia2018, Lateef2019} have tackled these problems and advancing solutions to the problem of the low resolution on the output maps. 
For better refinement, previous models~\cite{Zheng2015} post-process the results to enhance them, \eg, Conditional Random Fields (CRF)~\cite{Kraehenbuehl2011}. 
For global feature extraction with more information, new architectures~\cite{AmirulIslam2017, Valada2019, Alshammari2019} were created as well as sparse convolution operations~\cite{Yu2016, Chen2017a}. 
Commonly, methods in SS use hourglass models~\cite{Ronneberger2015, Noh2015, Badrinarayanan2017} that comprise a coding and decoding stages to recover the pixel-wise position of the segmented objects.

Other models use Multi-Task Learning (MTL) based on the idea that simultaneously learning related tasks can improve the performance on all of them~\cite{Zhang2017,Ruder2017,Thung2018}.
Hence, related tasks facilitate the transfer of shared knowledge among them.
\Eg, edge detection improves segmentation by adjusting the edge at each level of a CNN~\cite{Marmanis2018,Ding2018,Ding2020} or by learning edge-aware features~\cite{Ding2019}.
Although several MTL approaches~\cite{Dai2016,Kendall2018,Chen2019,Alshammari2019} were applied to the SS task, it is still difficult to say which auxiliary tasks most beneficial for the final SS results. 
Even more, which of these auxiliary tasks provide the necessary complementary information so that the models can address the loss of spatial precision.
We deduce that by reinforcing the object contours' information (through auxiliary tasks), we will be able to force greater attention (in the training phase) on the segmented objects' contours through the MTL approach.

In this work, we use a multi-task approach with complementary contour-based tasks for rich and robust feature extraction and address the problem of spatial precision loss.  
We work specifically with hourglass (encoder-decoder) models because we (empirically) discovered that a multi-task setup helps to adjust the latent space in these models (\ie, it exhibits a clustering behavior).
We also show that the improved results of different hourglass (encoder-decoder) architectures are directly related to this clustering behavior.
We present four different studies on the latent space for SS: 
(i)~visualization of the latent space behavior, 
(ii)~activation maps used by the models to predict the segmentation, 
(iii)~reduction of over-fitting of the models, and 
(iv)~the ablation studies on loss functions and complementary tasks.
The latent space's visualizations show different induced clusters when influenced by adding or removing the different complementary contour-based tasks.
Fig.~\ref{fig:fig_our_model} depicts the contour-based tasks used in our study.
Namely, they are edge detection~\cite{Gonzalez2006}, semantic contours~\cite{Hariharan2011}, semantic segmentation~\cite{Arnab2016}, and distance transform~\cite{Borgefors1986}.

In summary, our main contributions are:
\begin{itemize}
  \item Address the problem of spatial precision loss by obtaining hourglass models with better generalization by focusing on segmented objects' contours.
  \item Improve the SS results in hourglass models by using complementary information from contour-based tasks, and thus induce a clustering behavior in the latent space.
  \item Extensively evaluate common hourglass models, namely SegNet~\cite{Badrinarayanan2017} and UNet~\cite{Ronneberger2015}, on Cityscapes~\cite{Cordts2016}, and CamVid~\cite{Brostow2009} datasets. 
  Moreover, we show that the use of complementary tasks improves the state-of-the-art in hourglass (encoder-decoder) models.
\end{itemize}

\section{Related Work}
\label{sec:related_works}

Over the past years, the SS task has been done with deep learning as the preferred option due to deep neural networks'~(DNNs) extraordinary ability for feature extraction. 
The initial layers learn the low-level features (\eg, edges and texture), while the last layers learn the higher-level ones (\eg, identify different objects). 
The feature extraction of DNNs can be improved by adding complementary information (though a multi-task approach) in the training phase.
This section presents a review of relevant literature approaches that focus on SS tasks, using several approaches, such as deep learning and MTL\@.

\subsection{Semantic Segmentation}
\label{sec:sem_seg}
Semantic segmentation refers to the process of linking each pixel in an image to a class label.
In the deep models, FCN showed to be useful in this task. 
However, the first SS models produced low-resolution maps with a loss in spatial precision. 
Here we discuss different models created to deal with these problems.

Some researchers~\cite{Zheng2015} used FCN with CRFs as a post-processing step, but it is computationally expensive.
Consequently, embedding the post-processing steps within a network~\cite{Chen2017a, Li2019} was a viable solution. 
In contrast, we improve the hourglass (encoder-decoder) models without the need to use post-processing steps by introducing additional tasks that refine the latent space.

Other models~\cite{He2017a, Liu2018, Chen2018} adjusted the bounding boxes.
The intuition was to do object detection first and then refine the instances' contours.
Mask R-CNN~\cite{He2017a} used a feature pyramid network~\cite{Lin2017b} to extract a feature hierarchy in-network and an FCN to get a segmentation mask in each region of interest.
This region-based approach had proper segmentation but depended on the accurate detection of objects (bounding box).
Hourglass models, on the other hand, do not have this constraint.
Other models~\cite{Pinheiro2016} require more delimited boundaries for the segmentation as masks instead of just a box. 
They also use sliding operations to obtain better adjustment~\cite{Liu2016a} to the final targets.

Instead of an abrupt prediction of the last layer, the hourglass approach~\cite{Liu2015,Oliveira2016,AmirulIslam2017,Valada2019} (\ie, models with encoder-decoder stages, such as U-Net~\cite{Ronneberger2015}, DeconvNet~\cite{Noh2015}, SegNet~\cite{Badrinarayanan2017}) created a decoder stage to gradually recover the spatial information by combining multi-level feature maps from the encoder.
Thus, the flow of information from a lower scale to a higher one is done by an upsampling operation, \ie, bilinear interpolation~\cite{Gonzalez2006}, unpooling~\cite{Noh2015}, or DUpsampling~\cite{Tian2019}.
We consider hourglass models to have a robust decoding stage for the reconstruction of the pixel-wise predicted image.
Although hourglass models proved efficient in SS, they still need a more significant transfer of information between its stages, \eg, FC-DenseNet~\cite{Jegou2017} or UPSNet~\cite{Xiong2019}.
For this reason, we add complementary information to the models by adding the MTL approach (\ie, auxiliary task).

Though the previous models improved the objects' boundary, we need models that observe larger regions.
Thus, multi-scales models emerged~\cite{Lin2017a, Li2019, Tao2020}. 
They obtain a full semantic map in low-resolution (coarse prediction map), then refine it with different fusion operations, \eg, fusion cascade~\cite{Zhao2018}, attention blocks~\cite{Yu2018, Huang2019}, layer aggregation~\cite{Yu2018b}, residual units~\cite{Paszke2016}, and gated fusion~\cite{Li2020}.
These models are unnecessarily complex to extract robust features.
Instead, we use auxiliary tasks to reinforce the gradient and achieve better information extraction.

Current models (\eg, HRNetv2~\cite{Wang2020}, HRNet+OCR~\cite{Yuan2020}) perform multi-scale feature extraction by sharing feature maps across their different levels (scales), \ie, broadcasting context information at various resolutions.
Contrary to multi-scale models and to capture high-resolution feature maps, PSP-Net~\cite{Zhao2017} performs pooling operations at multiple grid scales. 
Simultaneously, DeepLabv3+~\cite{Chen2018a} and CasiNet~\cite{Jin2021} use Atrous Spatial Pyramid Pooling (ASPP)~\cite{Chen2017a, Chen2017} (\ie, several sparse filters) to modify the filters' size instead of the images' size~\cite{Ziegler2019}.
Later experiments showed that there are still limitations to get global features~\cite{Wang2018}. 
Moreover, the introduced dilated convolutions bring heavy computation complexity and a memory footprint, thus limiting many applications' usage.

The first attempt to address the high resource consumption of ASPP was FastFCN~\cite{Wu2019}, which performs a new method of ascending pyramidal sampling.
Besides, AdapNet++~\cite{Valada2019,Valada2017} proposed cascaded and parallel Atrous convolutions to capture long-range context using fewer parameters.
However, the problem of spatial precision loss persisted. 
These results lead us to believe that we need models that make use of inductive biases, \ie, more specific features from prior information. 
We address this problem by using well-behaved hourglass models paired with multi-task learning to improve the learned features.

\subsection{Multi-Task Learning}
\label{sec:MTL}
In machine learning, we generally train a single model to perform a specific task.
By focusing on a single task, we risk ignoring additional information that could help us learn a better representation of the desired task.
Instead, MTL~\cite{Caruana1997} aims to solve multiple related tasks simultaneously.
Thus, it facilitates the transfer of shared knowledge across relevant tasks~\cite{Long2017, Thung2018}.

In this literature review, we focus on supervised learning tasks due to similarity with our work.
Currently, machine learning models share knowledge in two ways~\cite{Thung2018,Zhang2018}:
(i)~feature-based MTL that distributes knowledge across training representative features, and
(ii)~parameter-based MTL that uses the model parameters trained in a specific task to fit the related tasks.
In this work, we are interested in studying the latent space shared among all tasks while focusing on SS as the main task restricted to hourglass models.

The previous models~\cite{Obozinski2006, Jebara2011, Zhang2010} used handcrafted features and assume that the data-to-target has a direct relationship.
Many times the data exhibit a complex data-to-target relationship~\cite{Thung2018}.
This assumption can restrict the models' performance.
For this reason, deep learning with MTL is used due to its capacity to learn nonlinear complex latent representations.
Deep MTL is grouped into two types~\cite{Thung2018}: hard (\ie, sharing parameters between all tasks) and soft (\ie, each task has its model, hidden layers, and parameters).

Previous models~\cite{Misra2016, Fang2017} used two separate architectures with soft parameter sharing.
They used cross-stitch units or task transfer connections to leverage the knowledge of the task-specific networks.
In contrast, MRN~\cite{Long2017} learned a Bayesian transfer relationship (both the last layers).
The previous models, having their own parameters for each task, made it easy to increase the number of required resources.

Unlike soft models, the hard ones do not need any assumption for the tasks' relation; they do this internally. 
Thus, some MTL models~\cite{Dai2016,Pinheiro2016} use a cascade-based approach to learn a task from the previous one.
However, this approach restricts the feature space.
Accordingly, models~\cite{Liao2016,Teichmann2018} (with independent tasks) focus on merging multiple loss functions (depending on the task) to ensure the convergence of the models and robustness to noise~\cite{Klingner2020}.
Some models~\cite{Hayder2017, Tan2018, Bischke2019} combined semantic segmentation with geometric information and others~\cite{Kendall2018, Kong2018} with depth.
Other models~\cite{Kendall2017, Kendall2018, Takikawa2019} measured and adjusted the degree of uncertainty of the samples along with the segmentation. 
We noted that the uncertainty (\ie, either due to noise at the capture or due to the prediction's degree of confidence) is related to the segmented objects' edges.
Similarly, previous works~\cite{Cheng2017,Liu2020} also used this line of reasoning by merging semantic contours with edge detection using the multi-scale feature in Acuna \etal's~\cite{Acuna2019} work or adjusting the contours across the entire network as done by Cheng \etal~\cite{Cheng2017} (\ie, multiple losses for detection).
Finally, the combination of SS with edge detection (in models~\cite{Ding2019}) showed significant improvements for the segmentation task.
These models (CCL~\cite{Ding2018}, CGBNet~\cite{Ding2020}) generally present an hourglass architecture with skip connections for the edge detection map.

Despite the useful latent space for related tasks obtained by the deep MTL approaches~\cite{Ding2020}, it is not yet explored or understood how this latent space behaves to improve the target task.
In other words, understanding what parts of the latent space improve SS is an open problem.
Moreover, information is even scarcer when the target task is SS on images (\ie, multi-label pixel-wise classification).

\section{Overview}
\label{sec:overview}
The idea of using CNNs as feature extractors and generators is not new. 
It has been widely used and achieved better results against traditional methods~\cite{Thoma2016}. 
Previous works (see Section~\ref{sec:sem_seg}) use a CNN for SS tasks and bring up challenging problems like the loss of spatial precision as the main problem.
Besides, we discussed in Section~\ref{sec:MTL} that deep MTL models obtain additional information from related tasks and learn at some level a new feature space shared across all tasks, specifically in hourglass models.
However, there is still no analysis of deep MTL models specifically for the SS task.
In particular, there is no indication of what happens with the shared features, how they behave, nor what are the most relevant related tasks for SS to enhance them.
In this work, we are interested in a particular type of behavior in the latent space, \ie, clustering, which improves the SS results in hourglass (encoder-decoder) architectures.
We empirically analyze how clustering in the latent space is influenced by different contour-based auxiliary tasks. 
We highlight that the clustering behavior is only observed in hourglass architectures. 
These models (\eg, UNet, SegNet, ParseNet) depend largely on the latent space to perform the reconstruction (a stream or up-down-up route). 
In contrast, other models (\eg, HRNet) perform feature extraction across multi-resolution (\ie, multi-stream). 
They distribute more contextual information but deprive the intermediate representation spaces of internal interpretability (\ie, decreasing the interpretability in latent representation).
In this section, we introduce our learning framework.
Also, we introduce the datasets and define the metrics we used in this work.
Recall that our study is entirely empirical. 
Our objectives are 
(i)~to propose and evaluate the use of contour-based auxiliary tasks to address the problem of loss of spatial precision, and 
(ii)~to show how the addition or deletion of these contour-based auxiliary tasks helps improve the SS results for hourglass models.

\subsection{Learning a Multi-Task Approach}
\label{sec:learning_MTL}
Deep MTL approaches learn features that might not be easy or possible to learn within original task.
We want to know if we can leverage the information in the training signals of other related SS tasks during the learning phase.
An effective way to achieve this is by giving cues to the model from other related tasks, \ie, predicting the features with an auxiliary task.

\makeatletter
\begin{figure}[tb]
\centering
\resizebox{\linewidth}{!}{%
  \begin{tikzpicture}[
  pics/named code/.style={code={\tikz@fig@mustbenamed%
      \begin{scope}[local bounding box/.expanded=\tikz@fig@name]#1\end{scope}%
  }},
  coder long height/.store in=\longheight,
  coder long height=1.5,
  coder short height/.store in=\shortheight,
  coder short height=.75,
  coder width/.store in=\width,
  coder width=1.3,
  coder fill/.store in=\coderfill,
  coder text/.store in=\codertext,
  coder label/.store in=\coderlabel,
  coder style hidden/.style={#1},
  coder style hidden/.default={coder label=, coder text=black, coder fill=white,},
  pics/encoder/.style = {named code={%
      \tikzset{coder style hidden, #1}%
      \coordinate (-center) at (0, 0);
      \coordinate (-east) at (\width/2, 0);
      \coordinate (-west) at (-\width/2,0);
      \coordinate (-north east) at (\width/2, -\shortheight/2);
      \coordinate (-south east) at (\width/2, \shortheight/2);
      \coordinate (-north west) at (-\width/2, -\longheight/2);
      \coordinate (-south west) at (-\width/2, \longheight/2);
      \draw[fill=\coderfill] (-west) -- (-north west) -- (-north east) -- (-south east) -- (-south west) -- (-west);
      \node[text=\codertext, anchor=center] at (-center)  {\coderlabel};%
  }},
  pics/decoder/.style = {named code={%
      \tikzset{coder style hidden, #1}%
      \coordinate (-center) at (0, 0);
      \coordinate (-east) at (\width/2, 0);
      \coordinate (-west) at (-\width/2,0);
      \coordinate (-north east) at (\width/2, -\longheight/2);
      \coordinate (-south east) at (\width/2, \longheight/2);
      \coordinate (-north west) at (-\width/2, -\shortheight/2);
      \coordinate (-south west) at (-\width/2, \shortheight/2);
      \draw[fill=\coderfill] (-west) -- (-north west) -- (-north east) -- (-south east) -- (-south west) -- (-west);
      \node[text=\codertext, anchor=center] at (-center)  {\coderlabel};%
  }},
  node distance=1cm,
  edge/.style={
    ->,
    >=Latex,
    shorten <= 2pt,
    shorten >= 2pt,
    rounded corners,
  },
  ]
  \node (input) {\includegraphics[width=2cm]{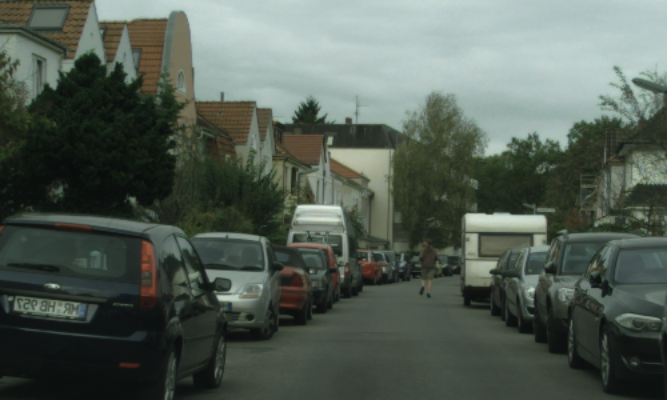}};
  \pic[right=1.5cm of input] (enc)  {encoder={coder label={}}};
  \pic[right=2cm of enc.east] (dec)  {decoder={coder label={}}};
  \node[draw, rectangle, rounded corners, minimum height=1cm, fill=black!50] at ($(enc)!.5!(dec)$) (lat) {};
  \node[below=5pt of lat, circle, draw,minimum size=1cm, path picture ={
    \foreach \i in {1,...,100}
    \path let \p1=(path picture bounding box.south west),
    \p2=(path picture bounding box.north east),
    \n1={rnd}, \n2={rnd} in
    ({\n1*\x1+(1-\n1)*\x2},{\n2*\y1+(1-\n2)*\y2}) node[circle, draw, fill, minimum size=1pt, inner sep=0pt] {};
  }] (spc) {};
  
  \pic[above right=.125cm and 1.5cm of dec] (dec-s)  {decoder={coder label={$\mathcal{T}_S$}, coder width=.75}};
  \pic[below right=.125cm and 1.5cm of dec] (dec-c)  {decoder={coder label={$\mathcal{T}_C$}, coder width=.75}};
  
  \pic[above=1cm of dec-s] (dec-e)  {decoder={coder label={$\mathcal{T}_E$}, coder width=.75}};
  \pic[below=1cm of dec-c] (dec-d)  {decoder={coder label={$\mathcal{T}_D$}, coder width=.75}};
  
  \node[right=1.cm of dec-s] (o-s) {\includegraphics[width=2cm]{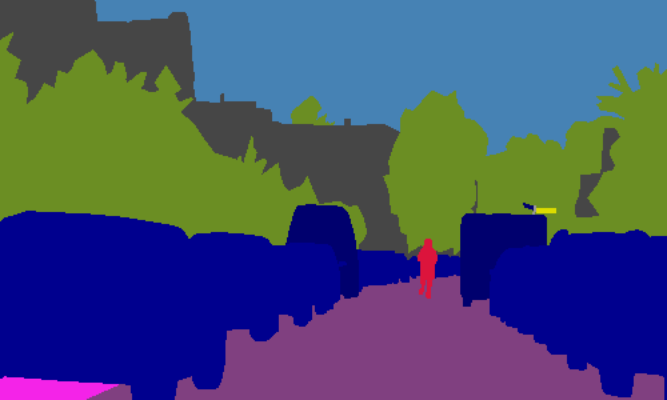}};
  \node[right=1.cm of dec-e] (o-e) {\includegraphics[width=2cm]{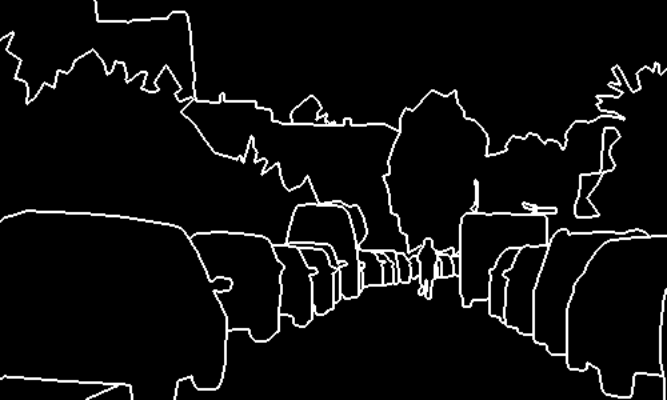}};
  \node[right=1.cm of dec-c] (o-c) {\includegraphics[width=2cm]{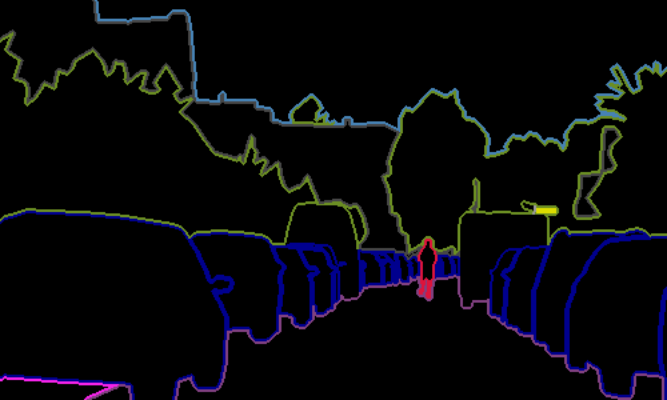}};
  \node[right=1.cm of dec-d] (o-d) {\includegraphics[width=2cm]{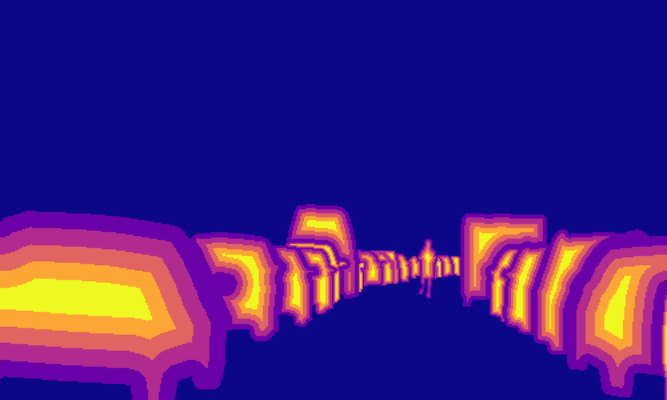}};
  
  \draw[edge] (input) -- (enc);
  \draw[edge] (enc) -- (lat);
  \draw[edge] (lat) -- (dec);
  
  \draw[edge] (dec) -| ($(dec)!.5!(dec-e)$) |- (dec-e);
  \draw[edge] (dec) -| ($(dec)!.5!(dec-s)$) |- (dec-s);
  \draw[edge] (dec) -| ($(dec)!.5!(dec-c)$) |- (dec-c);
  \draw[edge] (dec) -| ($(dec)!.5!(dec-d)$) |- (dec-d);
  
  \draw[edge] (dec-e) -- (o-e);
  \draw[edge] (dec-s) -- (o-s);
  \draw[edge] (dec-c) -- (o-c);
  \draw[edge] (dec-d) -- (o-d);
  
  \draw (lat.south west) -- (spc.north west);
  \draw (lat.south east) -- (spc.north east);
  
  \node[below=5pt of spc] {What is happening here?};
  \node[above=5pt of dec-e] {Task-dependent};
  \end{tikzpicture}
}
  \caption{Illustration of a multi-task hourglass model, for tasks of edge detection~(E), semantic segmentation~(S), semantic contours~(C), and distance transform~(D), from top to bottom. 
  Note that the model share weights in the first layers (encoder and decoder), and the specific features for each task are obtained in the last layers (specific task decoders $\mathcal{T}_\cdot$).}
  \label{fig:fig_our_model}
\end{figure}
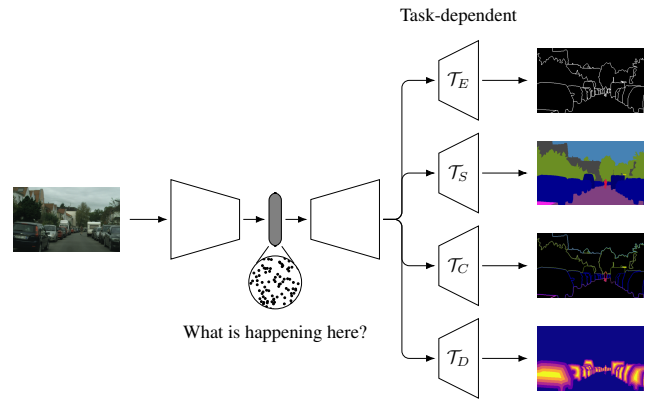
\makeatother

The goal of an auxiliary task in MTL is to learn useful shared representations for the main task (\ie, add a regularizing factor~\cite{CheolSong2019}). 
They are closely related to the main task, so adding them allows the model to learn beneficial representations.
However, finding an auxiliary task that helps improve the SS task is not trivial~\cite{Guo2020}.
At first glance, tasks that seem different can use similar representations, and tasks that seem related can adjust different internal functions~\cite{Caruana1997}.
We still do not know which auxiliary tasks will help in practice for SS\@.
Finding an auxiliary task is largely based on the assumption that they should be related to the main task in some way.
We perceive, from Fig.~\ref{fig:SS_loss_spatial_precision}, that spatial precision loss is generally produced on the edge of segmented objects. 
So we use tasks related to the gradient or edge regions. 
That is, we give more attention to the contours of the objects.
With this in mind, we propose to employ three types of contour-based auxiliary tasks to improve the boundary of segmented object and, therefore, the SS task.
We choose auxiliary tasks to reinforce and complement the information obtained from the edges of objects. 
Thereby, we address the problem of spatial precision loss, generally reflected in the segmented objects' contours, \cf Fig.~\ref{fig:SS_loss_spatial_precision}.

We propose to use the additional tasks of edge detection~(E), semantic segmentation~(S), semantic contours~(C), and distance transform~(D), \cf Fig.~\ref{fig:fig_our_model}.
Edge detection~\cite{Gonzalez2006} aims to extract object boundaries. 
Distance transform~\cite{Borgefors1986}, in our case, is a distance function to the objects' edges.
Semantic contour~\cite{Hariharan2011} produces a pixel-wise level dense classification on objects' contour. 
Initially, we tried to use a continuous distance transform (\ie, without quantification). 
However, we discard it because of the longer training time to convergence, and the results were comparable with the quantized distance transform.
We intuit that this behavior is due to the higher degrees of freedom when fitting a regressor.
In Appendix~\ref{sec:apx_final_representation}, we detail how to quantize the distance transform.

Although these auxiliary tasks were previously used in MTL models~\cite{Cheng2017, Yu2018, Bischke2019, Liu2020}, they were not used together in the same model.
Besides, the impact produced by adding each of the auxiliary tasks has not yet been studied.
Consequently, we show how each of them improves the latent space separation (Section~\ref{sec:visualization_LS}). 
We evaluated each of these auxiliary tasks' contribution (quantitative results) to the SS task (see our ablation study in Section~\ref{sec:ablation_studies}).

Unlike the previous hourglass models (encoder-decoder), the hourglass with MTL adds specific heads for each task (\ie, deconvolution layers in the last decoder stage). 
Our hourglass model with MTL (Fig.~\ref{fig:fig_our_model}) has two types of hidden layers, the shared layers, and task-specific layers. 
The shared layers learn a low-level representation of the data, influenced by all tasks, while the task-specific layer learns the parameters for the pixel-wise classification network. 
These specific layers map the learned latent representations from the previously shared layers to the task-specific output layers (\ie, target for each task). 

Consider that the hourglass models with MTL work over a set of images~$X$, and has a corresponding ground truth per task $\{Y^t\}_{t \in T}$, \ie, set of pixel-wise labeled images by task. 
Each $i$th sample has a corresponding ground truth image $y_i^t$ for the corresponding task~$t$.
Thus, the hourglass model is represented by $f(x; \theta^h, \theta^t) = y^t$ such that some parameters, $\theta^h$, are shared between the contour-based tasks, and some, $\theta^t \in \{\theta^E, \theta^S, \theta^C, \theta^D\}$, are particular to each specific task.

The hourglass parameters are learned by solving an optimization problem that minimizes a weighted sum of the losses for each task. 
It is defined by 
\begin{equation} 
\label{eq:opt_MTL}
\mathcal{L}_\mathit{final} = 
\min\limits_{\theta^h,\theta^E,\theta^S,\theta^C,\theta^D} \frac{1}{|T|N}\sum_{t\in T}  \sum_{i=1}^N \lambda_t \mathcal{L}_t \left(\theta^h, \theta^t \right),
\end{equation}
where we used four tasks $T=\{E, S, C, D\}$, $N$ is the number of samples, and the loss of the auxiliary task $t$ is defined as 
\begin{equation} 
\label{eq:task-loss}
\mathcal{L}_t(\theta^h, \theta^t) = \mathcal{L}_t( f^t(x_i; \theta^h, \theta^t), y_i^t ),
\end{equation}
and where $\mathcal{L}_t = \{\mathcal{L_\mathit{E}},\mathcal{L}_{\mathit{S}},\mathcal{L_\mathit{C}},\mathcal{L_\mathit{D}}\}$ represents the loss functions of each task.
To moderate each task's importance on the model loss~\eqref{eq:opt_MTL}, we use a scalar $\lambda_i$ to weigh each task loss~\eqref{eq:task-loss}.
Each loss function helps to adjust the latent space into a useful representation for each task. 
In this work, we use for each task's loss the cross-entropy and soft IoU loss functions (see details in Appendix~\ref{sec:apx_training}).

\begin{figure}[tb]
  \centering
  \begin{tikzpicture}[
    >=Latex,
  ]
  \node (latent_space) [inner sep=0pt] at (0,0)  {\includegraphics[width=2cm]{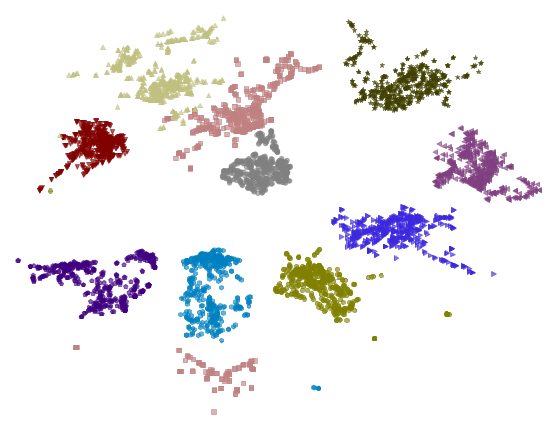}};
  \node (circle_LS) [circle, draw=black, minimum size=55pt, inner sep=0pt, line width=0.2pt] at (0,0) {};
  \node (S) [inner sep=0pt] at (2,1.5) {\includegraphics[width=1.5cm]{ss-seg}};
  \node (E) [inner sep=0pt] at (2,-1.5) {\includegraphics[width=1.5cm]{ss-edge}};
  \node (C) [inner sep=0pt] at (-2,1.5) {\includegraphics[width=1.5cm]{ss-contours}};
  \node (D) [inner sep=0pt] at (-2,-1.5) {\includegraphics[width=1.5cm]{ss-distance}};
  
  \coordinate (bbsw) at (current bounding box.south west);
  \coordinate (bbne) at (current bounding box.north east);
  
  \path[->] 
  (S) edge [bend left, font=\scriptsize, pos=0.3]  node[below right] {$\nabla_\theta \mathcal{L}_S\left(\theta^{h},\theta^S\right)$} (circle_LS)
  (E) edge [bend left, font=\scriptsize, pos=0.3]  node[below left] {$\nabla_\theta \mathcal{L}_E\left(\theta^{h},\theta^E\right)$} (circle_LS)
  (C) edge [bend left, font=\scriptsize, pos=0.3]  node[above right] {$\nabla_\theta \mathcal{L}_C\left(\theta^{h},\theta^C\right)$} (circle_LS)
  (D) edge [bend left, font=\scriptsize, pos=0.3]  node[above left] {$\nabla_\theta \mathcal{L}_D\left(\theta^{h},\theta^D\right)$} (circle_LS);
  
  \pgfresetboundingbox
  \path[use as bounding box] (bbsw) rectangle (bbne); 
  \end{tikzpicture}
  \caption{%
    The contour-based auxiliary tasks influence the latent space through their gradients (by backpropagation).  
    Due to the contour origin of the task their corresponding losses penalize the contours of the objects to improve.
  }
  \label{fig:influence_aux_task}
  \vspace{-10pt}
\end{figure}

In Fig.~\ref{fig:fig_our_model}, the $\mathcal{T}_E$, $\mathcal{T}_S$, $\mathcal{T}_C$, and $\mathcal{T}_D$ blocks represent the layers that extract specific information to discriminate each task. 
In other words, each distinct decoding stage for each task have independent parameters.
Since we work with the latent space, we analyze how the auxiliary tasks influence the latent space (feature representation).
We do not need to use large networks for the specific tasks. 
Thus, each specific task-block contains two layers of convolutions, ensuring that the enhancement is performed on the shared parameters (\ie, a more robust feature extraction).
The gradients of the separate tasks carry out the influence of the additional tasks on latent space (see Fig.~\ref{fig:influence_aux_task}). 
Due to the chosen tasks, gradients are prone to bring more significant changes to the edges of objects. 
That is, to provide further attention to the edges of the objects when performing the segmentation.

We are aware of the existence of non-deep-learning-based methods to do edge detection (\eg, Canny~\cite{Canny1986}, Sobel~\cite{Gonzalez2006}, hierarchical method~\cite{Arbelaez2010}) or distance transform (\eg, mathematical morphology~\cite{Gonzalez2006}).
We plan to use these auxiliary tasks only in the training phase, and not as final tasks that would replace the existing methods.
The multi-task setup is to provide complementary information to the latent space and adjust the SS task.
Additionally, we evaluate the impact of each auxiliary task.

\subsection{Experiments Description}
\label{sec:exp_descrip}
We describe a set of empirical studies in order to show how the addition or removal of contour-based auxiliary tasks helps improve the semantic segmentation task. 
We also demonstrate that the use of auxiliary tasks diminishes the loss of spatial precision in the segmented objects.

\textit{Ablative Studies} (Section~\ref{sec:ablation_studies}): 
We performed two ablation studies. 
The first study is on the loss functions.
We want to know which of the loss functions (cross-entropy and soft IoU) trains a better model for the SS task.
We determined the best results according to both loss functions and a data augmentation technique explained in Section~\ref{sec:datasets}.
We did a second study to know which contour-based task helps improve the prediction.
In other words, we evaluated quantitatively how the addition or removal of related tasks impacts the final segmentation result.
We obtained the best results by training the models using all tasks together.

\textit{Visualization of Latent Space Behavior} (Section~\ref{sec:visualization_LS}):
Here, we show the latent space behavior in the well-known SegNet model using complementary information from contour-based auxiliary tasks.
To plot the samples in this study, we used the multidimensional projection method t-SNE~\cite{Maaten2008}.
Experiments show that the latent space exhibits clustering behavior, improving dissimilarity and segmentation results by adding auxiliary tasks.

\textit{Showing Activation Maps} (Section~\ref{sec:activation_maps}): 
With the previous experiments, we obtained the best results using all the auxiliary tasks. 
In this study, we plot the activation maps used by the hourglass models to predict the segmentation.
In this work, the activation maps are the regions that the network use for dense classification at the pixel-wise level.
Therefore, we observe that by training hourglass models with contour-based auxiliary tasks, the model employs activation maps not previously used to improve the contours of segmentation.

\textit{Reducing the Over-Fitting} (Section~\ref{sec:reduce_overfitting}): 
In this study, we investigated whether there is a segmentation improvement on the segmented object's edge. 
To do so, we evaluated the classification errors of the segmented object's edge. 
We performed these experiments for various hourglass models for binary and multi-label segmentation.
The empirical results show that there is an improvement at the segmented object's edges.
This improvement appears due to the having a more robust latent space that better defines the objects.
By using complementary information, we learn models that generalize better than the traditional ones.
Thus, we address the problem of spatial precision loss.

\textit{Comparing Results} (Section~\ref{sec:comparing_results}): 
Previously, we carried out extensive studies on CamVid dataset due to the shorter required training time.
We present final results on a set of hourglass models with and without MTL for the SS task and comparison tables for the CamVid, Cityscape, and Freiburg Forest datasets.
We improve the final segmentation results when using MTL with contour-based tasks.
The improvement may seem modest.
However, the amount of pixels at the objects' edges is small in comparison to the image total amount of pixels.

\subsection{Datasets}
\label{sec:datasets}
We evaluated our proposed methodology on Cityscapes~\cite{Cordts2016}, CamVid~\cite{Brostow2009}, and Freiburg Forest~\cite{Valada2016} datasets. 
They contain several types of urban/forest scenarios.

\textbf{Cityscapes:} The dataset has \num{5000} samples with $2048 \times 1024$ size images and pixel-level labels of \num{19} semantic classes.
There are \num{2979}, \num{500}, and \num{1525} images in the training, validation, and test set, respectively.
We do not use coarse data in our experiments.
For this work, we required a wide variety of samples; for this reason, we employ data augmentation.
We applied a random crop of $300 \times 500$ and some random transformations of contrast, brightness, and horizontal flip; thus, we generated \num{17500} training samples.
We use the original validation set to compare the MTL models with a resolution of $768 \times 384$ pixels (resize). 
For this, we employ bilinear interpolation (for RGB images) and the nearest-neighbor interpolation (for the labels).
To facilitate comparison with previous approaches, we report results on the reduced \num{11} class label set consisting of: \textit{sky, building, road, sidewalk, fence, vegetation, pole, car/truck/bus, traffic sign, person, rider/bicycle/motorbike, and background}.

\textbf{CamVid:} It is road scene understanding dataset for SS with \num{11} classes: \textit{building, tree, sky, car, sign, road, pedestrian, fence, pole, sidewalk, and cyclist}. 
The dataset has \num{367}, \num{101}, and \num{233} samples for training, validation, and test set, respectively, with images size of $360 \times 480$.
We apply the same transformations of Cityscapes for the data augmentation, with a random crop size of $260 \times 346$, generating \num{5616} samples for the training set.
We report results on the original test set.

\textbf{Freiburg Forest:} It is a dataset on forests with six classes: \textit{sky, trail, grass, vegetation, obstacle, and void}.
Note, forested environments are unstructured (\eg, trails), unlike urban scenes that are highly structured (rigid and geometric objects, \eg, buildings).
We do data augmentation using transformations of Cityscape, generating \num{1840} samples for the training set.
We conserve the original testing set (\num{136} images).
Note, all the images are resized at $768 \times 384$ pixels.

\subsection{Evaluation Metrics}
\label{sec:metrics}
The success achieved by the SS methods must be measured by the achievements of the final applications.
They are generally too difficult to evaluate because graphical applications often require an expert user. 
For this reason, it is necessary to use application-independent measures of accuracy. 
Thus, to evaluate our results on segmentation, we chose accuracy, intersection-over-union, precision, and recall metrics as
validation measures (from Csurka \etal~\cite{Csurka2013}).
The intersection-over-union (IoU) is defined by
\begin{equation} 
  \label{eq:metric_iou}
  \text{IoU} = \sum_{i}^N \frac{P_i \cap Y_i}{P_i \cup Y_i} = \sum_{i}^N \frac{\mathit{TP}_i}{\mathit{TP}_i + \mathit{FP}_i + \mathit{FN}_i},
\end{equation}
the accuracy (Acc), \ie, pixel-wise accuracy is
\begin{equation} 
  \label{eq:metric_acc}
  \text{Acc} = \sum_{i}^N \frac{P_i \cap Y_i}{Y_i} = \sum_{i}^N \frac{\mathit{TP}_i + \mathit{TN}_i}{\mathit{TP}_i + \mathit{TN}_i + \mathit{FP}_i + \mathit{FN}_i},
\end{equation}
the precision (Prec) is 
\begin{equation} 
\label{eq:metric_prec}
\text{Prec} = \sum_{i}^N \frac{\mathit{TP}_i}{\mathit{TP}_i + \mathit{FP}_i},
\end{equation}
and the recall (Rec) is 
\begin{equation} 
\label{eq:metric_rec}
\text{Rec} = \sum_{i}^N \frac{\mathit{TP}_i}{\mathit{TP}_i + \mathit{FN}_i}.
\end{equation}
We assume that $P_i$ is the set of pixels predicted as the $i$th class, $Y_i$ is pixels set belonging to the $i$th class, and $N$ is the number of classes.
Besides, $\mathit{TP}_i$, $\mathit{FP}_i$, $\mathit{TN}_i$, and $\mathit{FN}_i$ represent True/False Positives and True/False Negatives, respectively, for a given class $i$.
Note, these metrics are widely used in SS\@.

Furthermore, to measure the behavior of the latent space, we use metrics for clustering. 
Thus, we utilize the Silhouette Coefficient (SSI)~\cite{Rousseeuw1987} defined by
\begin{equation} 
\label{eq:metric_ssi}
\text{SSI} = \sum_{i}^N \frac{b_i - a_i}{\max\{a_i, b_i\}},
\end{equation}
where $a_i$ is the mean intra-cluster distance from sample $i$, and $b_i$ is the mean nearest-cluster distance from $i$ to each sample. 
Note that a higher value is related to better-defined clusters.
The Calinski-Harabasz Index (CHI)~\cite{Calinski1974} is given by
\begin{equation} 
\label{eq:metric_chi}
\text{CHI} = \frac{\text{SS}_M}{\text{SS}_W} \frac{N-k}{k-1},
\end{equation}
where $k$ is the number of clusters, and $N$ is the total number of observations (\ie, data points), $\text{SS}_W$ is the overall within-cluster variance and, $\text{SS}_M$ is the overall between-cluster variance.
Note that a higher value is associated with dense and well-distributed clusters.
Finally, we employ the Davies-Bouldin Index (DBI)~\cite{Davies1979} denoted by
\begin{equation} 
\label{eq:metric_dbi}
\text{DBI} = \frac{1}{k}\sum_i^k \max_{j \ne i}\left(\frac{s_i + s_j}{d_{ij}}\right),
\end{equation}
where $s_i$ is the average distance between each point of cluster $i$ and its centroid, and $d_{ij}$ is the distance between cluster centroids $i$ and $j$.
Note that a lower value is related to better separation between the clusters.

\begin{figure*}[tb]
  \centering
  \newlength{\wsz} \setlength{\wsz}{.2\linewidth}
  \newlength{\hsz} \setlength{\hsz}{1.0in}
  \captionsetup[subfloat]{justification=centering}
  \subfloat[S \protect\\ $\text{SSI}=0.384$, $\text{DBI}=1.360$\label{fig:latent-S}]{\includegraphics[width=\wsz,height=\hsz]{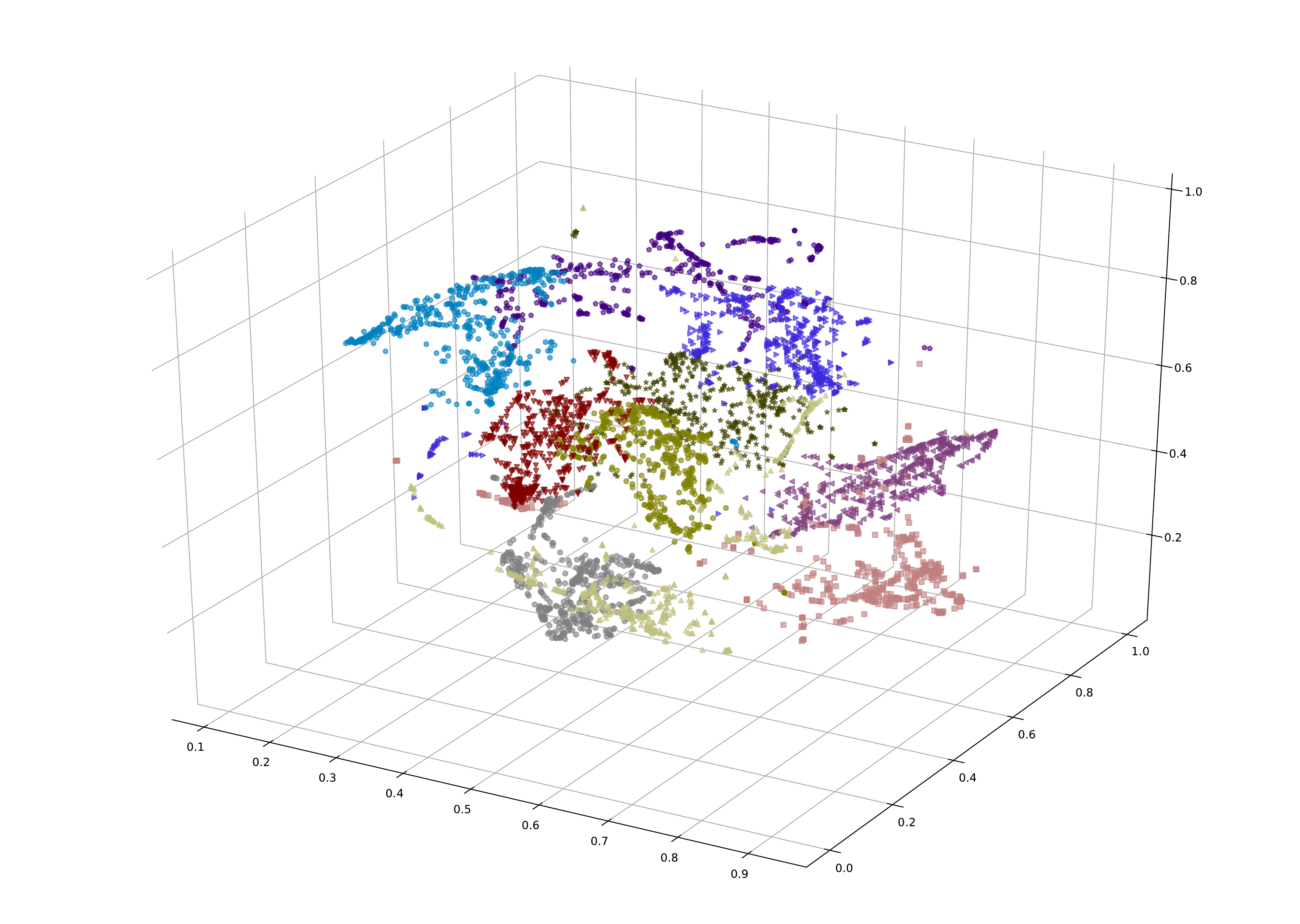}}
  \subfloat[S+E \protect\\ $\text{SSI}=0.391$, $\text{DBI}=1.141$\label{fig:latent0-SB}]{\includegraphics[width=\wsz,height=\hsz]{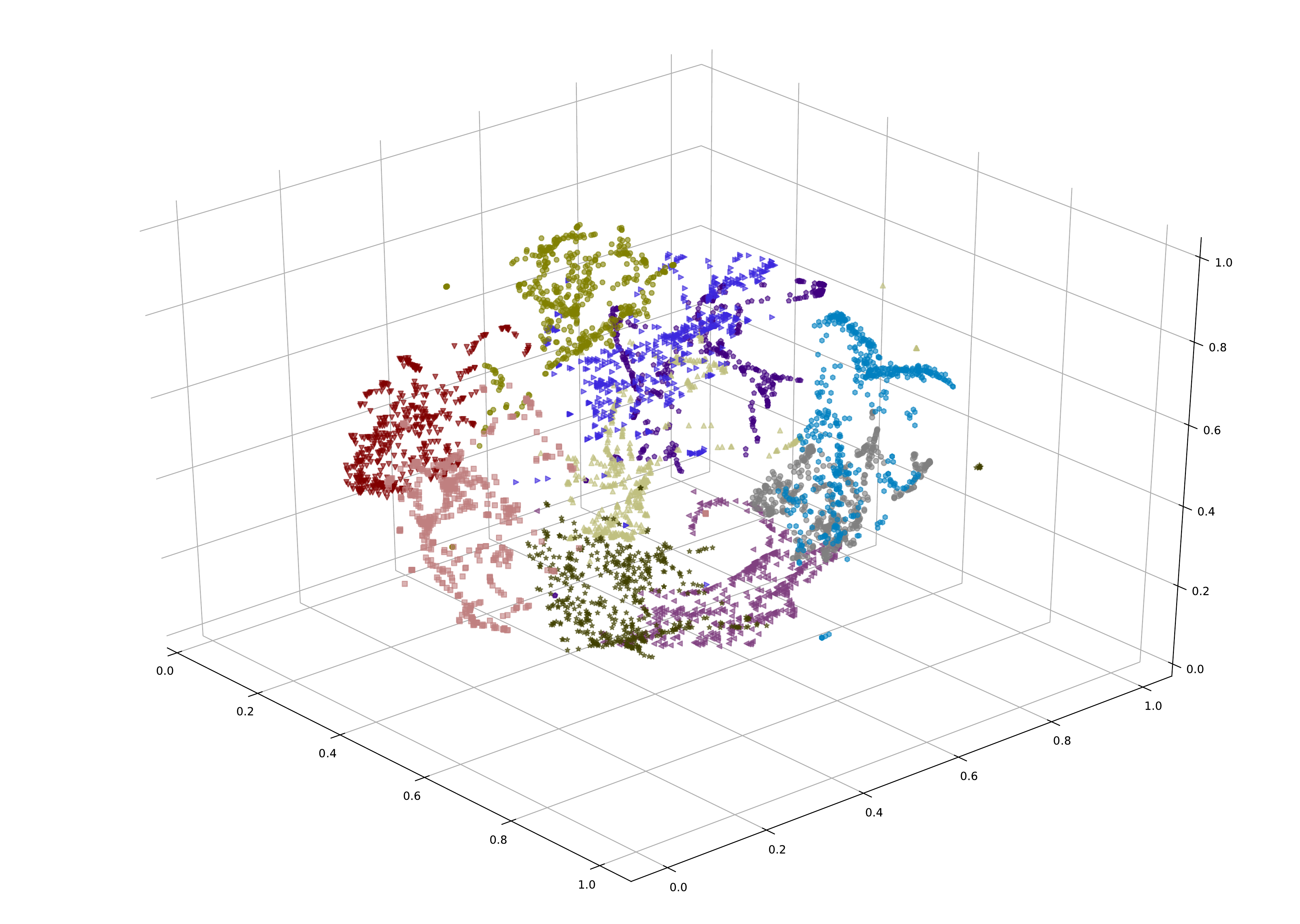}}
  \subfloat[S+D \protect\\ $\text{SSI}=0.394$, $\text{DBI}=1.275$\label{fig:latent0-SE}]{\includegraphics[width=\wsz,height=\hsz]{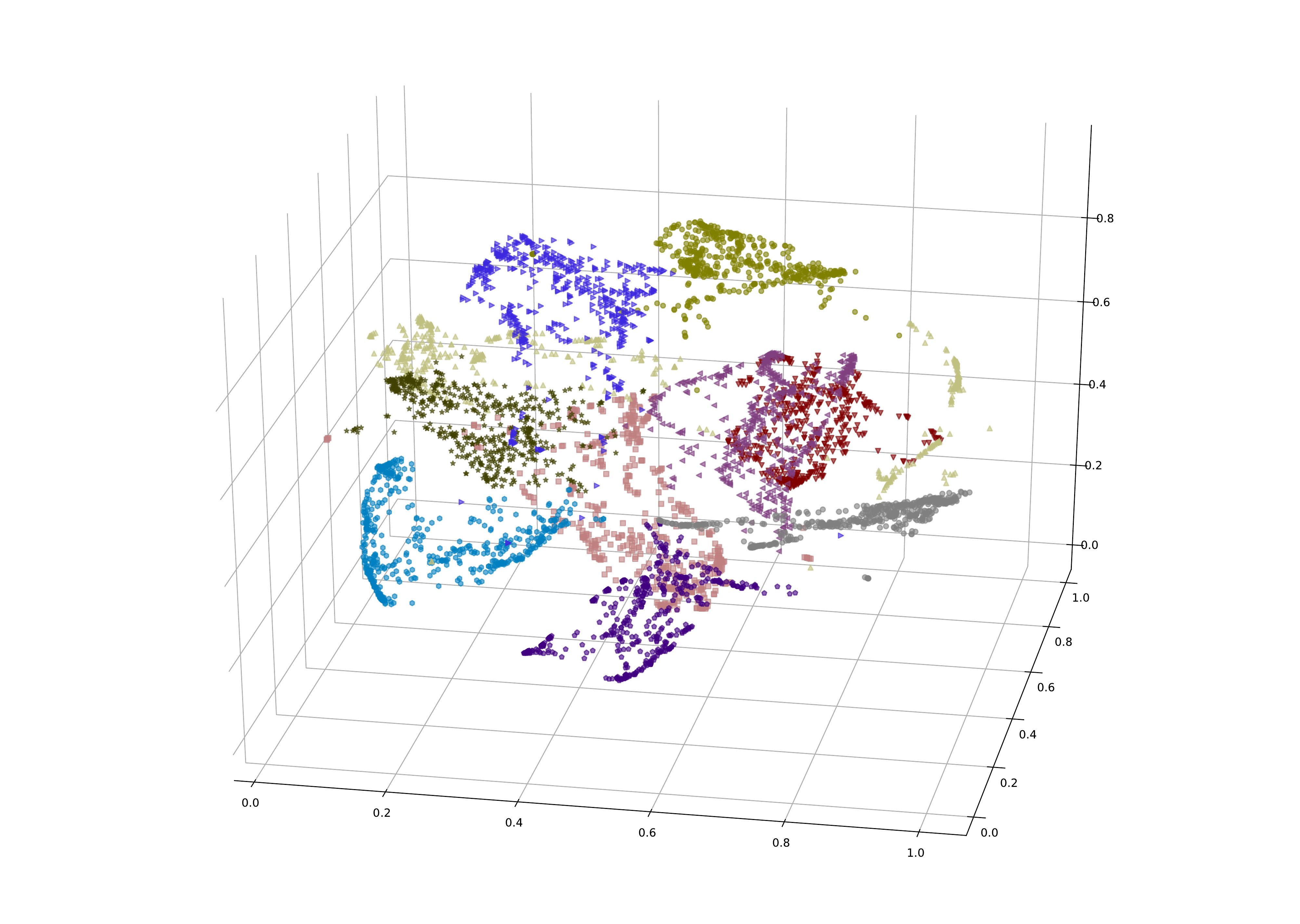}}
  \subfloat[S+E+C \protect\\$\text{SSI}=0.437$, $\text{DBI}=1.150$\label{fig:latent-SBC}]{\includegraphics[width=\wsz,height=\hsz]{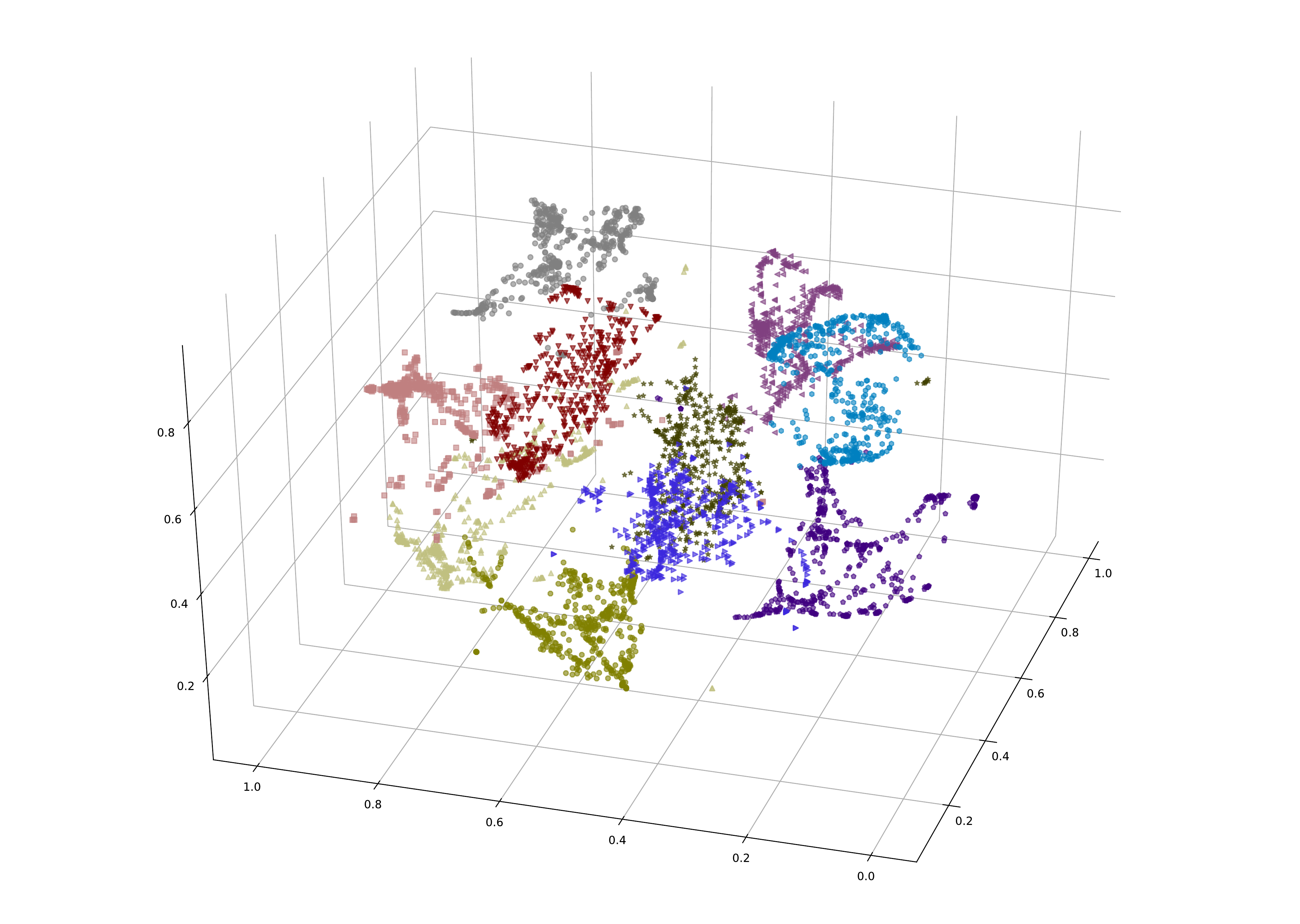}}
  \subfloat[S+E+C+D \protect\\ $\text{SSI}=0.636$, $\text{DBI}=0.774$\label{fig:latent-SBCE}]{\includegraphics[width=\wsz,height=\hsz]{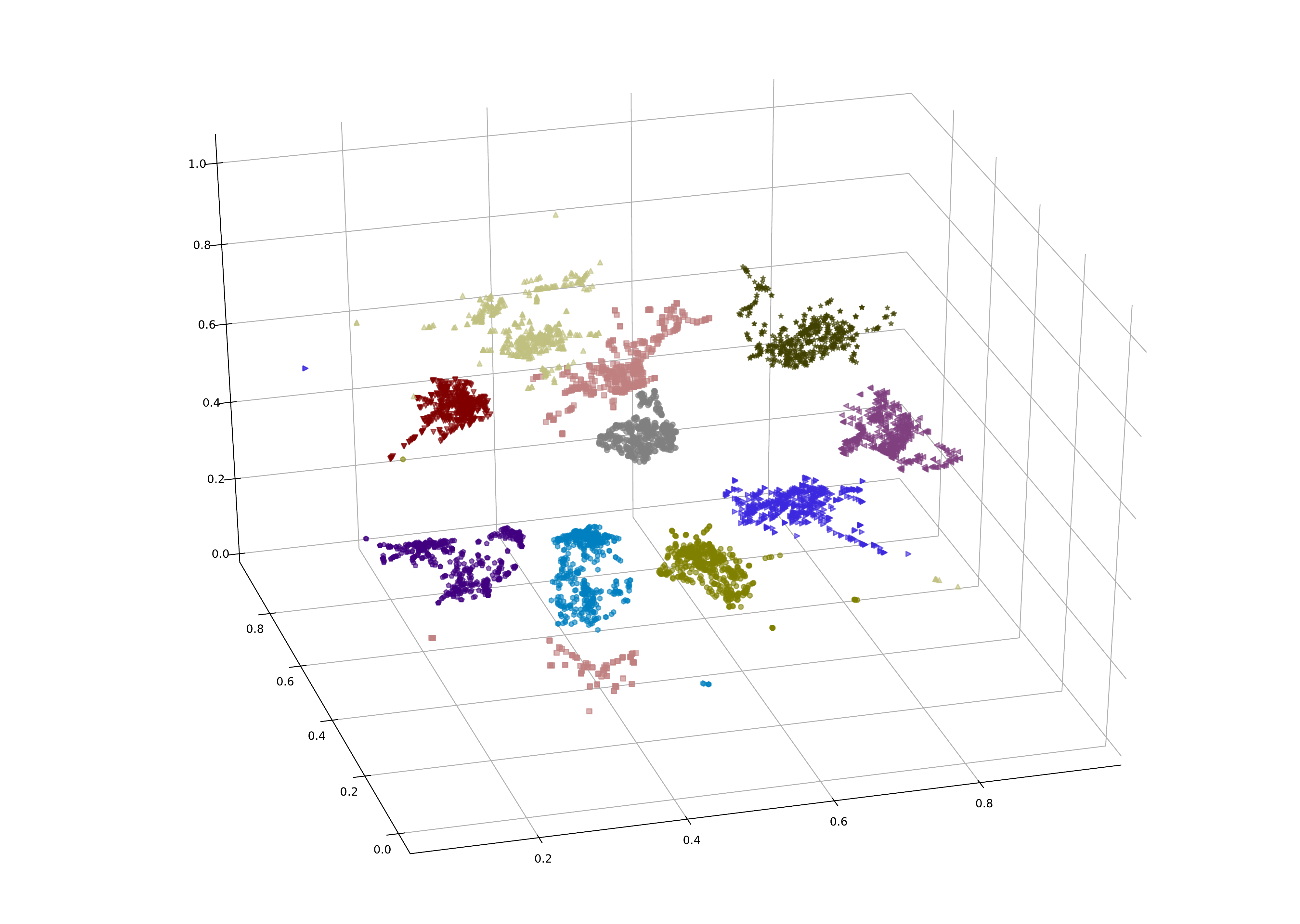}}
  \caption{
    We show the shared latent space on the Camvid testing dataset.  
    We combine the different tasks of edge detection~(E), semantic segmentation~(S), semantic contour~(C), and distance transform~(D).
    When adding tasks related to semantic segmentation, \ie, by providing complementary information, maps of similar features (within a unimodal multi-task hourglass model) are clustered together in a similar latent space, and they are not arbitrarily placed.
    We confirm this behavior by using a set of metrics for clustering shown in Table~\ref{tab:ablation2}.
  }
  \label{fig:latent-space-tasks}
\end{figure*}

\section{Experiments}
\label{sec:Experiments}

This section presents a set of empirical studies of the latent space on hourglass models based on the MTL approach.
Then we perform a series of ablation experiments on the CamVid dataset using the well-known SegNet model (see Fig.~\ref{fig:fig_our_model}).
Also, we present comparisons between the different hourglass models with and without multi-task (contour-based tasks) for the Camvid, Cityscape, and Freiburg Forest datasets.

\subsection{Ablative Studies}
\label{sec:ablation_studies}
We present two types of ablative analysis on the CamVid dataset, using the well-known hourglass model, SegNet, for the semantic segmentation task.
The first study, presented in Table~\ref{tab:ablation1}, shows the distinct behavior of the SegNet model by using the different loss functions (cross-entropy and loss-IoU) and data augmentation. 
The reported results focus on a single semantic segmentation task on the CamVid test set. 
The best performance of the model presented in Table~\ref{tab:ablation1} was achieved using both loss functions and data augmentation.
So we opted for this configuration for the following experiments.

\begin{table} [tb]
  \centering
  \sisetup{
    table-format = 1.4,
  }
  \newrobustcmd{\B}{\bfseries}
  \caption[Ablative study on objective functions]{Ablative study on loss functions of the SegNet~\cite{Badrinarayanan2017} model with a multi-task approach on the CamVid test set. 
  }
  \label{tab:ablation1}
  \scriptsize
  \newlength{\colsep}
  \setlength{\colsep}{7pt}
  \begin{tabular}{%
      @{ }S
      @{\hspace{\colsep}}%
      S@{\hspace{\colsep}}%
      S@{\hspace{\colsep}}%
      S@{\hspace{5pt}}%
      S@{\hspace{\colsep}}%
      S@{\hspace{\colsep}}%
      S@{\hspace{\colsep}}%
      S@{\hspace{\colsep}}%
      S@{ }%
    }
    \toprule
    \multicolumn{3}{c}{\textbf{ W/O }} & & \multicolumn{4}{c}{\textbf{Metrics Segmentation}} \\
    \cmidrule{1-3} \cmidrule{5-8}
    \textbf{Cross} & \textbf{IoU} & \textbf{Aug} & & \textbf{Acc}$\uparrow$ & \textbf{IoU}$\uparrow$  & \textbf{Prec}$\uparrow$ & \textbf{Rec}$\uparrow$\\
    \midrule
    {\checkmark} & {--} & {--} & & 0.6240563636363636 & 0.5206054545454545 & 0.6793700000000003 & 0.6240563636363636\\
    {--} & {\checkmark} & {--} & & 0.6475249785834231 & 0.5401836363636363 & 0.7049187706970556 & 0.6475249785834231\\
    {\checkmark} & {\checkmark} & {--} & & 0.6581575601230989 & 0.5490536363636364 & 0.7164937779264005 & 0.6581575601230989\\
    {\checkmark} & {--} & {\checkmark} & & 0.666546369459209 & 0.5569418181818183 & 0.7256261347626717 & 0.666546369459209\\
    {--} & {\checkmark} & {\checkmark} & & 0.692171577505276 & 0.5774290909090909 & 0.7535226495723512 & 0.692171577505276\\
    {\checkmark} & {\checkmark} & {\checkmark} & & \B 0.7066977958022812 & \B 0.5895472727272727 & \B 0.7693364085522801 & \B 0.7066977958022812\\
    \bottomrule
  \end{tabular}
\end{table}

\begin{table} [tb]
  \centering
  \sisetup{
    table-format = 1.4,
  }
  \newrobustcmd{\B}{\bfseries}
  \caption[Ablative study on tasks]{Ablative study on tasks of edge detection~(E), semantic segmentation~(S), semantic contours~(C), and distance transform~(D) of the SegNet~\cite{Badrinarayanan2017} on the CamVid test set. 
  }
  \label{tab:ablation2}
  \scriptsize
  \setlength{\colsep}{6pt}
  \begin{tabular}{%
      @{ }S
      @{\hspace{\colsep}}%
      S@{\hspace{\colsep}}%
      S@{\hspace{\colsep}}%
      S@{\hspace{\colsep}}%
      S@{\hspace{5pt}}%
      S@{\hspace{\colsep}}%
      S@{\hspace{\colsep}}%
      S@{\hspace{\colsep}}%
      S@{\hspace{\colsep}}%
      S@{\hspace{5pt}}%
      S@{\hspace{13pt}}%
      S[table-format=1.2]@{\hspace{3pt}}%
      S@{\hspace{\colsep}}%
      S@{ }%
    }
    \toprule
    \multicolumn{4}{c}{\textbf{ Task }} & & \multicolumn{4}{c}{\textbf{Metrics Segmentation}} & \multicolumn{4}{c}{\textbf{Metrics Clustering}} \\
    \cmidrule{1-4} \cmidrule{6-9} \cmidrule{11-13}
    \textbf{S} & \textbf{E} & \textbf{C} & \textbf{D} & & \textbf{Acc}$\uparrow$ & \textbf{IoU}$\uparrow$  & \textbf{Prec}$\uparrow$ & \textbf{Rec}$\uparrow$ & & \textbf{SSI}$\uparrow$ & \textbf{CHI}$\uparrow$ & \textbf{DBI}$\downarrow$\\
    \midrule
    {\checkmark} & {--} & {--} & {--} & & 0.7066977958 & 0.5895472727 & 0.7693364085 & 0.7066977958 & & 0.38425976303141018 & 1847.3994384097211 & 1.3601770879105035\\
    {\checkmark} & {\checkmark} & {--} & {--} & & 0.708715135634 & 0.5912301946 & 0.771532556545 & 0.708715135634 & & 0.39057391438134792 & 2262.2252961128643 & 1.1411086606582739\\
    {\checkmark} & {--} & {--} & {\checkmark} & & 0.711739987608 & 0.5937536116 & 0.774825518215 & 0.711739987608 & & 0.39396657599774471 & 2009.0362826279143 & 1.2751765978457497\\
    {\checkmark} & {\checkmark} & {\checkmark} & {--} & & 0.732902862737 & 0.6114082801 & 0.797864178364 & 0.732902862737 & & 0.43689662862361123 & 2507.3270600642659 & 1.1494541530014966\\
    {\checkmark} & {\checkmark} & {--} & {\checkmark} & & 0.730620008642 & 0.6095038587 & 0.795378982031 & 0.730620008642 & & 0.44368765567597529 & 3023.7583123330578 & 0.92003561489275787\\
    {\checkmark} & {\checkmark} & {\checkmark} & {\checkmark} & & \B 0.75032213165 & \B 0.6259399265 & \B 0.816827415357 & \B 0.75032213165 & & \B 0.63597228088027968 & \B 4060.1852001405282 & \B 0.77434128358149601\\
    \bottomrule
  \end{tabular}
\end{table}

\begin{figure*}[tb]
  \centering  
  \newlength{\colfig}
  \setlength{\colfig}{1pt}
  \setlength{\hsz}{0.15\linewidth}
  {\renewcommand{\arraystretch}{0}
    \resizebox{\textwidth}{!}{%
      \begin{tabular}{%
          @{}%
          c@{\hspace{\colfig}}
          c@{\hspace{\colfig}}
          c@{\hspace{\colfig}}
          c@{\hspace{2pt}}
          c@{\hspace{\colfig}}
          c@{\hspace{\colfig}}
          c@{}
        }
        \rotatebox{90}{\scriptsize \ SegNet+MTL} &
        \includegraphics[width=\hsz, height=0.08\linewidth]{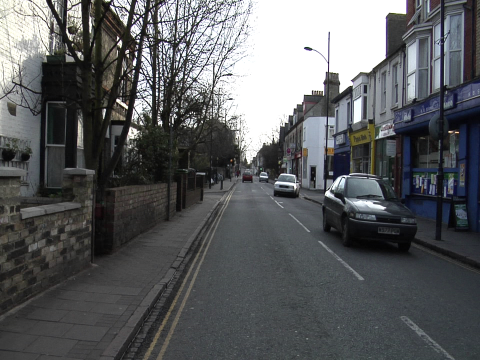} &
        \includegraphics[width=\hsz, height=0.08\linewidth]{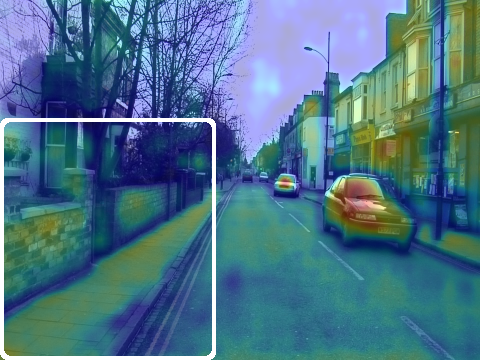} &
        \includegraphics[width=\hsz, height=0.08\linewidth]{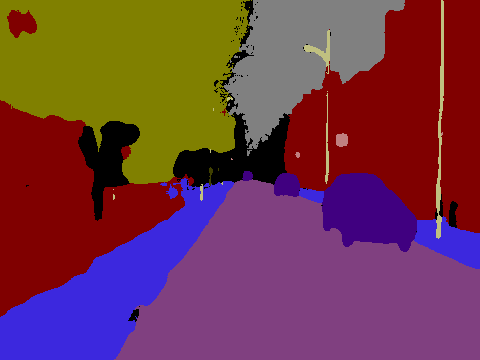} &
        \includegraphics[width=\hsz, height=0.08\linewidth]{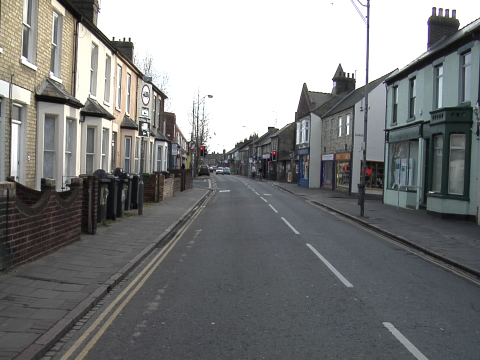} &
        \includegraphics[width=\hsz, height=0.08\linewidth]{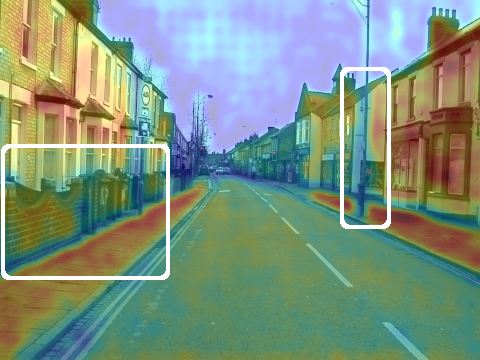} &
        \includegraphics[width=\hsz, height=0.08\linewidth]{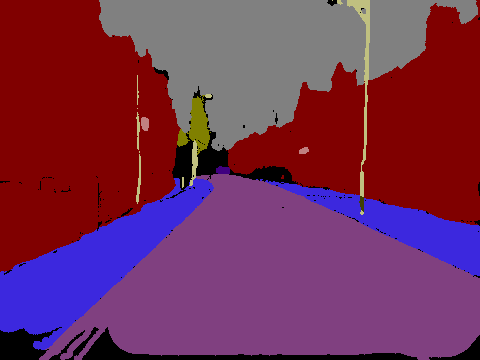} \\[.2mm]
        {\tiny \rotatebox{90}{\scriptsize \ \ \ \ \ SegNet}} &
        \includegraphics[width=\hsz, height=0.08\linewidth]{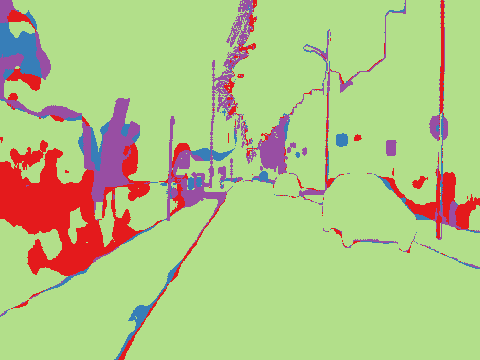} &
        \includegraphics[width=\hsz, height=0.08\linewidth]{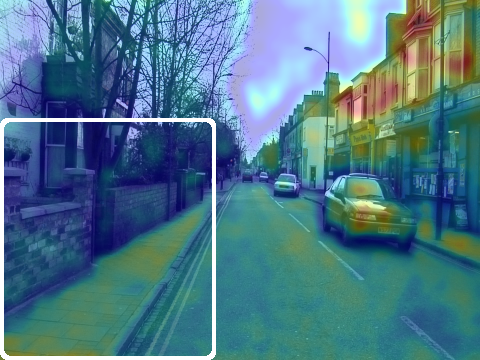} &
        \includegraphics[width=\hsz, height=0.08\linewidth]{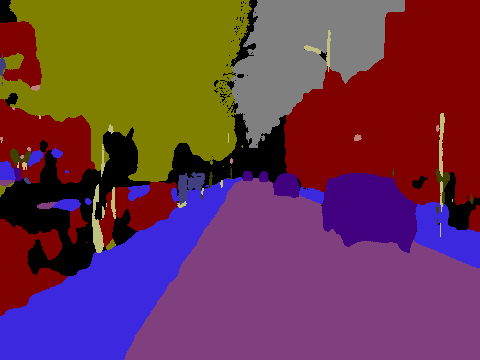} &
        \includegraphics[width=\hsz, height=0.08\linewidth]{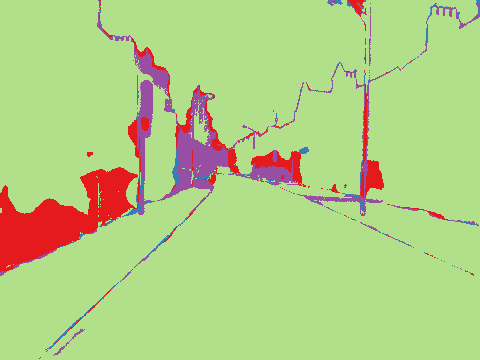} &
        \includegraphics[width=\hsz, height=0.08\linewidth]{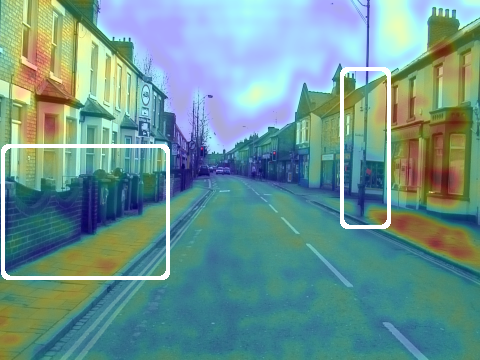} &
        \includegraphics[width=\hsz, height=0.08\linewidth]{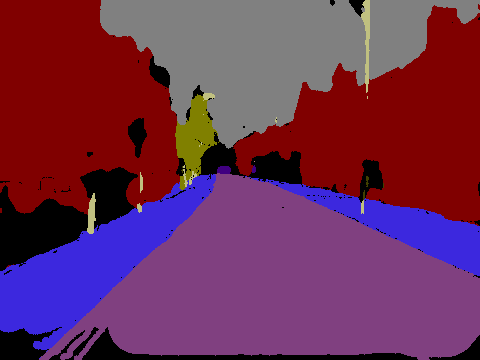} \\[.2mm]
        \rotatebox{90}{\scriptsize \ \ UNet+MTL} &
        \includegraphics[width=\hsz, height=0.08\linewidth]{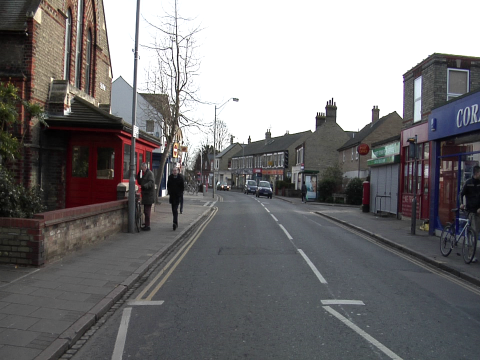} &
        \includegraphics[width=\hsz, height=0.08\linewidth]{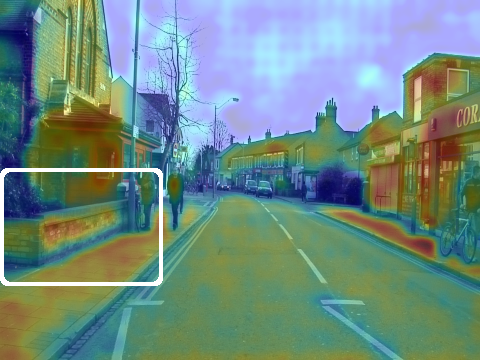} &
        \includegraphics[width=\hsz, height=0.08\linewidth]{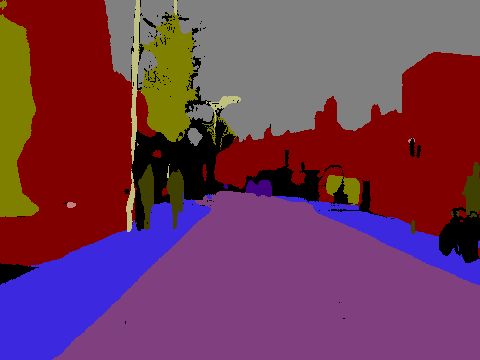} &
        \includegraphics[width=\hsz, height=0.08\linewidth]{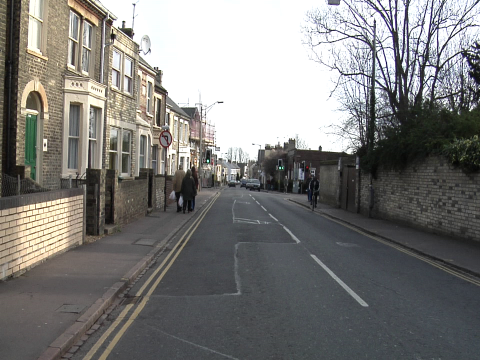} &
        \includegraphics[width=\hsz, height=0.08\linewidth]{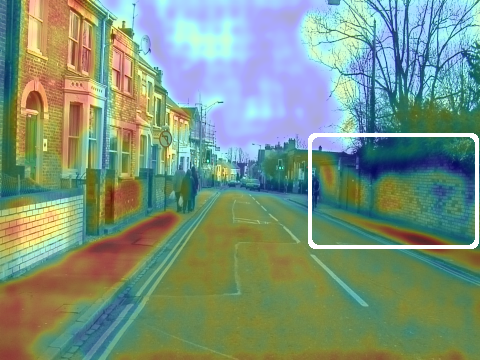} &
        \includegraphics[width=\hsz, height=0.08\linewidth]{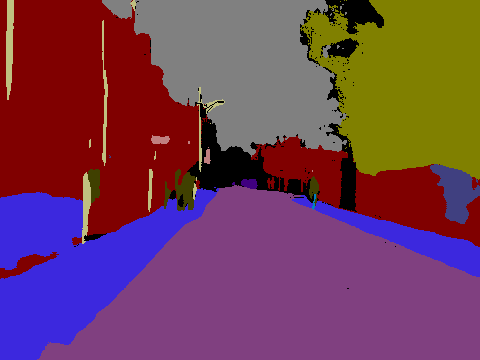} \\[.2mm]
        {\tiny \rotatebox{90}{\scriptsize \ \ \ \ \ UNet}} &
        \includegraphics[width=\hsz, height=0.08\linewidth]{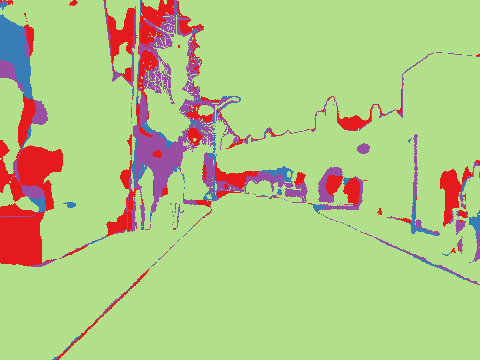} &
        \includegraphics[width=\hsz, height=0.08\linewidth]{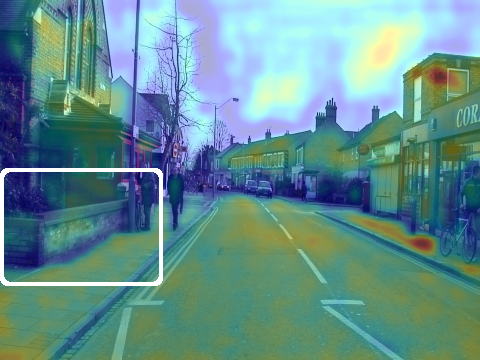} &
        \includegraphics[width=\hsz, height=0.08\linewidth]{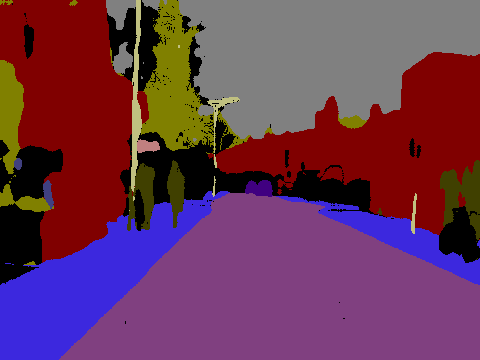} &
        \includegraphics[width=\hsz, height=0.08\linewidth]{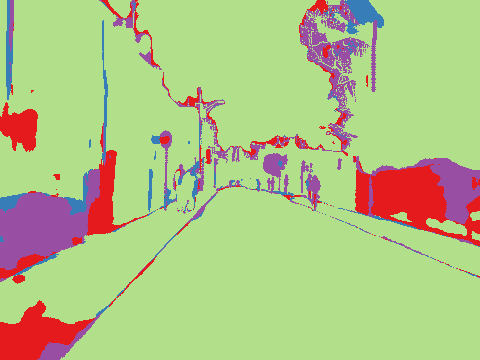} &
        \includegraphics[width=\hsz, height=0.08\linewidth]{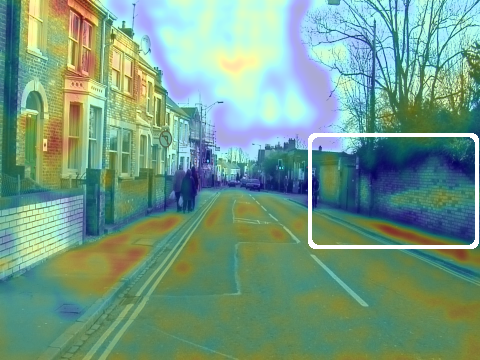} &
        \includegraphics[width=\hsz, height=0.08\linewidth]{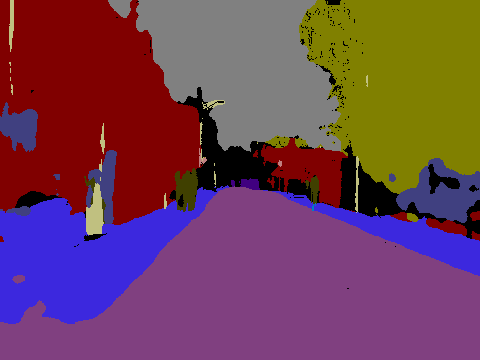} \\[1.2mm]
        &
        \scriptsize (a) Comparison &
        \scriptsize (b) Activation map &
        \scriptsize (c) Prediction &
        \scriptsize (d) Comparison &
        \scriptsize (e) Activation map &
        \scriptsize (f) Prediction\\
      \end{tabular} 
    }
  }
  \caption{%
    A comparison of activation maps (\ie, regions that are responsible for CNN's prediction) produced by the SegNet~\cite{Badrinarayanan2017} and UNet~\cite{Ronneberger2015} models with and without a multi-task approach.
    By using related tasks (\ie, by adding complementary information), the activation maps are better delimited (white bounding box).
    In the comparison images, correctly segmented regions are green while incorrectly segmented ones by the original models (SegNet and UNet) are red, by the multi-task models (SegNet+MTL and UNet+MTL) are blue; and by both are purple.
  }
  \label{fig:discriminative-regions}
\end{figure*}

The second study, presented in Table~\ref{tab:ablation2}, focuses on the internal behavior of the SegNet model's latent space when adding contour-based auxiliary tasks.
These tasks are edge detection (E)~\cite{Gonzalez2006}, semantic contours (C)~\cite{Hariharan2011}, quantized distance transform (D)~\cite{Hayder2017}, and semantic segmentation (S)~\cite{Arnab2016} as the main task.
Here, we use the segmentation metrics to evaluate the predicted regions. 
To evaluate the behavior (distribution), we use several clustering metrics. 
We notice a direct correlation between clustering behavior and segmentation results.
These quantitative results on CamVid are complementary results to those presented in Section~\ref{sec:visualization_LS}.
In Table~\ref{tab:ablation2}, our best results are produced by using both loss (cross-entropy and loss-IoU), data augmentation, and all tasks. 
We replicate this setting in Cityscapes and Freiburg Forest datasets.

\subsection{Visualization of Latent Space Behavior}
\label{sec:visualization_LS}

One way to understand the latent space in the hourglass model is to look at it, how it behaves, and its influence on the segmentation predictions.
Thus, we plot the latent space in which all the tasks are involved. 
We illustrate this space for the segmentation task using t-SNE in Fig~\ref{fig:latent-space-tasks}. 
Note that by using more related tasks, the space is better delimited.

We see that the SS task by itself presents a poorly distributed latent space; see Fig.~\ref{fig:latent-S}. 
By adding the edge detection task, the latent space improves its clusters per class, although it could be improved, see Fig.~\ref{fig:latent0-SB}. 
A particular task that supports segmentation is the distance transform quantified by adding geometric information to the feature maps, see Fig.~\ref{fig:latent0-SE}. 
On the other hand, with the semantic contours task, the semantic information is reinforced, achieving a better distribution in the latent space, see Fig.~\ref{fig:latent-SBC}. 
Thus, by adding geometric information and a higher quality of semantic information, the latent space presents a better delimitation and, therefore, better quantitative results, see Fig.~\ref{fig:latent-SBCE}.

We deduced that by adding complementary information (\ie, auxiliary tasks) in the MTL stage, the features that stimulate the activation of the same neurons on the network are reinforced across tasks.
Moreover, these features are clustered together.
Based on the set of clustering metrics (SSI, CHI, and DBI), shown in Table~\ref{tab:ablation2}, we can say that maps of similar features within an MTL hourglass model are correctly grouped, and they are not placed arbitrarily.

\subsection{Visualization of Activation Maps}
\label{sec:activation_maps}
Another way to understand CNNs is to look at the important image regions that influence their SS predictions.
In this study, we analyze the regions used by the hourglass models to make the best prediction of the segmentation (activation map) when applying the latent space adjusted by the multi-tasks.
The proposed visualization of activation maps is typically performed during inference (testing) to provide visual explanations for the network's prediction.

We present, in Fig.~\ref{fig:discriminative-regions}, a comparison between the image regions that were responsible for CNNs prediction (\ie, activation maps) of SegNet~\cite{Badrinarayanan2017} and UNet~\cite{Ronneberger2015} models with and without MTL\@.
We notice that the activation maps are better adjusted to the objects' contours (white bounding box). 
This behavior happens due to the better distributed latent space (\ie, decoder stage) on the models trained with MTL, see Fig.~\ref{fig:latent-SBCE}. 
The latent space's clustering behavior provides the networks' ability to use regions they did not use before to make the segmentation prediction.

The prediction columns in Fig.~\ref{fig:discriminative-regions} show the predictions made by both models; in addition, the rows show the models with and without MTL\@.
In the comparison columns, we show the correct and incorrect segmented regions color coded. 
The green regions are correctly segmented.
Red and blue color regions indicate the regions incorrectly segmented for models without and with MTL, respectively.. 
Lastly, the purple regions are the incorrectly segmented ones produced by both models.

We show different activation for SegNet and UNet with and without MTL in Fig.~\ref{fig:discriminative-regions}.
First, we reaffirm the spatial precision loss as the main problem of SS due to the incorrect segmentation in the objects' boundary.
Second, adding tasks focused on object contours helps hourglass models highlight regions that were not adequately delimited (white bounding box). 
In conclusion, we can say that the improvement of SegNet+MTL and UNet+MTL (quantitative results in Section~\ref{sec:ablation_studies}) happens mainly due to a higher flow of information provided by the contour-based auxiliary tasks.
It better delimits the activation maps used for dense pixel prediction.

\subsection{Reducing the Over-Fitting}
\label{sec:reduce_overfitting}

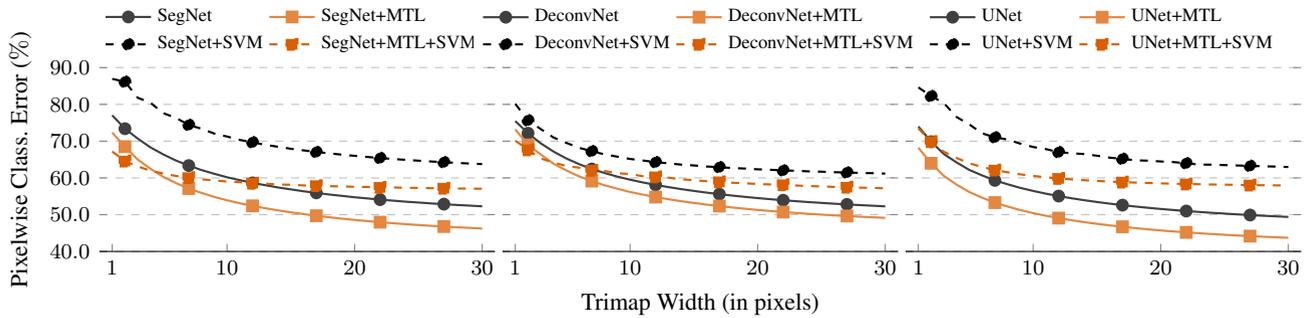
\begin{figure*}[tb]
  \centering
  \begin{tikzpicture}%
  \begin{groupplot}[
  group style={
    group size=3 by 1, 
    xlabels at=edge bottom,
    ylabels at=edge left,
    x descriptions at=edge bottom,
    y descriptions at=edge left,
    horizontal sep=.1cm,
  }, 
  footnotesize,
  height=4cm,
  width=6.8cm,
    cycle list/Paired,
    cycle multiindex* list={%
     [2 of]mark list\nextlist
     solid, solid, dashed, dashed\nextlist
     black!75, Dark2-B!75, black, Dark2-B\nextlist
    },
  /tikz/mark repeat=5,
  /tikz/mark phase=2,
  ymajorgrids,
  major grid style={dashed},
  ylabel={Pixelwise Class.\ Error ($\%$)},
  xtick={1,10,20,30},
  xmax=31,
  xmin=0,
  ytick={40,50,60,70, 80, 90},
  ymax=90,
  ymin=40,
  x tick label style={/pgf/number format/precision=0},
  legend pos=outer north center,
  legend columns=2,
  legend style={
    cells={anchor=west},
    font=\scriptsize,
    draw=none,
  },
  ]
  
  \nextgroupplot[%
    legend style={at={(0.49,1.02)},anchor=south},
  ]%
  \addplot table[x=width, y=trimap, col sep=comma, header=true]{img/trimap_SegNet_S.csv};%
  \addlegendentry{SegNet}%
  
  \addplot table[x=width, y=trimap, col sep=comma, header=true]{img/trimap_SegNet_SBCE.csv};%
  \addlegendentry{SegNet+MTL}%
  
  \addplot table[x=width, y=trimap, col sep=comma, header=true]{img/trimap_SegNet_S_svm_car.csv};%
  \addlegendentry{SegNet+SVM}%
  
  \addplot table[x=width, y=trimap, col sep=comma, header=true]{img/trimap_SegNet_SBCE_svm_car.csv};%
  \addlegendentry{SegNet+MTL+SVM}%

  \nextgroupplot[%
  xlabel={Trimap Width (in pixels)}, %
  ]%
  \addplot table[x=width, y=trimap, col sep=comma, header=true]{img/trimap_DeconvNet_S.csv};%
  \addlegendentry{DeconvNet}%
  
  \addplot table[x=width, y=trimap, col sep=comma, header=true]{img/trimap_DeconvNet_SBCE.csv};%
  \addlegendentry{DeconvNet+MTL}%
  
  \addplot table[x=width, y=trimap, col sep=comma, header=true]{img/trimap_DeconvNet_S_svm_car.csv};%
  \addlegendentry{DeconvNet+SVM}%
  
  \addplot table[x=width, y=trimap, col sep=comma, header=true]{img/trimap_DeconvNet_SBCE_svm_car.csv};%
  \addlegendentry{DeconvNet+MTL+SVM}%

  \nextgroupplot[%
  ]%
  \addplot table[x=width, y=trimap, col sep=comma, header=true]{img/trimap_UNet_S.csv};%
  \addlegendentry{UNet}%
  
  \addplot table[x=width, y=trimap, col sep=comma, header=true]{img/trimap_UNet_SBCE.csv};%
  \addlegendentry{UNet+MTL}%
  
  \addplot table[x=width, y=trimap, col sep=comma, header=true]{img/trimap_UNet_S_svm_car.csv};%
  \addlegendentry{UNet+SVM}%
  
  \addplot table[x=width, y=trimap, col sep=comma, header=true]{img/trimap_UNet_SBCE_svm_car.csv};%
  \addlegendentry{UNet+MTL+SVM}%
  \end{groupplot}%
  \end{tikzpicture}%
  \caption[Pixelwise classification error \vs trimap width]{%
    Pixelwise classification error \vs trimap width for the hourglass models focused on semantic segmentation (SegNet~\cite{Badrinarayanan2017}, DeconvNet~\cite{Noh2015}, and UNet~\cite{Ronneberger2015}) on the CamVid dataset. 
    Circle marks represent the base network, and square ones represent the addition of MTL\@. 
    The dashed-line lighter-tone denotes multi-label segmentation, while the solid-line darker-tone represents binary segmentation (also denoted with a +SVM).
  }
  \label{fig:study_trimap}
  \vspace{-10pt}
\end{figure*}

In this study, analyzed the contours of the segmented objects.
We evaluate whether there is improvement in the objects' edges, and if the proposal addresses the problem of spatial precision loss.

Our study (in Fig.~\ref{fig:study_trimap}) shows that there is improvement in the segmented objects' boundary.
We compare the popular SegNet~\cite{Badrinarayanan2017}, DeconvNet~\cite{Noh2015}, and UNet~\cite{Ronneberger2015} hourglass models for SS on the CamVid dataset.
Note, these models share similar operations to avoid using models with additional operations (\eg, atrous convolutions) and ensure that the improvement is not due to the use of these operations.

\begin{figure}[tb]
  \centering
  \subfloat[Ground truth]{\includegraphics[width=0.33\linewidth,height=0.7in]{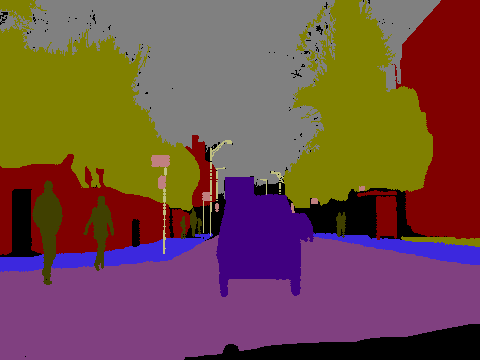}%
    \label{fig:trimap_gt}}%
  \subfloat[Trimap (8\,px)]{\includegraphics[width=0.33\linewidth,height=0.7in]{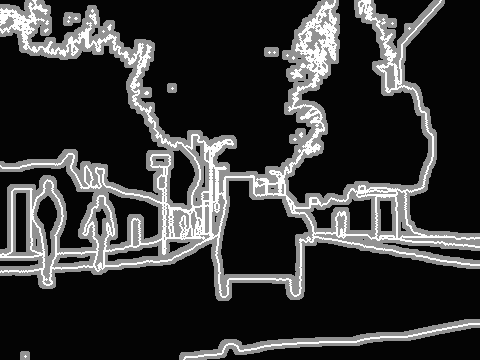}%
    \label{fig:trimap_8px}}%
  \subfloat[Evaluation region]{\includegraphics[width=0.33\linewidth,height=0.7in]{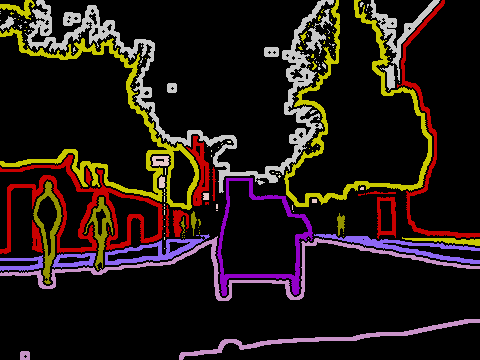}%
    \label{fig:trimap_app}}%
  \caption{Illustration of boundary accuracy evaluation using Trimap~\cite{Csurka2013, Kraehenbuehl2011}. 
    (a)~The image ground-truth from the Camvid dataset. 
    (b)~The Trimap used for measuring the pixel boundary labeling accuracy (gray region) with a width of $8$ pixels.
    And, (c)~an example of the evaluation region.
    }
  \label{fig:trimap_camvid}
\end{figure}

For comparison, we report experiments employing Trimap~\cite{Csurka2013, Kraehenbuehl2011}, which focuses on boundary regions of segmentation; see evaluation region in Fig~\ref{fig:trimap_app}. 
The Trimap is a rough image segmentation in the foreground, background, and unknown regions, shown in Fig.~\ref{fig:trimap_8px} with white, black, and gray regions, respectively.
The idea is to define a narrow band (gray region defined using a width pixel) around each contour and compute pixel-wise accuracy in the given band. 
The error curve comparison (plots in Fig.~\ref{fig:study_trimap}) shows that learning the contour-based auxiliary tasks did not allow the network to overfit, enabling the networks to generalize better.

From the previous analyzes, we conclude that the IoU improvement is especially due to better performance near the objects' boundary.
Qualitatively (overlapping of segmented regions in Fig.~\ref{fig:discriminative-regions}) and quantitatively (error curve comparison in Fig.~\ref{fig:study_trimap}), on SS task, we find an improved performance near boundaries by adding a multi-task approach to the hourglass models.
Besides, auxiliary tasks in hourglass models encourage clustering behavior in similar feature maps (\ie, latent space). 
This behavior is reflected in Fig.~\ref{fig:latent-space-tasks}, where we visualize that the latent space influenced by the multi-task approach (contour-based tasks) is not spaced arbitrarily.

\subsection{Comparing Results}
\label{sec:comparing_results}
Finally, we report comparative results (models with and without MTL approach) of several hourglass models existing in the literature.
We present our quantitative results for the CamVid, Cityscape, and Freiburg Forest datasets in Tables~\ref{tab:result_camvid}, \ref{tab:result_cityscape}, and~\ref{tab:result_forest}, respectively. 
We use the IoU metric (higher is better) for each class on all datasets. 
Note that the second to last and last columns (in all tables) show the mean IoU of the classes for each model, with (w/) and without (w/o) MTL\@.

\begin{table}[tb]
  \centering
  \sisetup{
    table-format = 1.2,
  }
  \caption[Results on CamVid test]{IoU results on the CamVid test set for semantic segmentation. 
  }
  \label{tab:result_camvid}
  
  \definecolor{cv-sky}{RGB}{128,128,128}
  \definecolor{cv-build}{RGB}{128,0,0}
  \definecolor{cv-pole}{RGB}{192,192,128}
  \definecolor{cv-road}{RGB}{128,64,128}
  \definecolor{cv-sidewalk}{RGB}{60,40,222}
  \definecolor{cv-tree}{RGB}{128,128,0}
  \definecolor{cv-sign}{RGB}{192,128,128}
  \definecolor{cv-fence}{RGB}{64,64,128}
  \definecolor{cv-car}{RGB}{64,0,128}
  \definecolor{cv-pedestrian}{RGB}{64,64,0}
  \definecolor{cv-cyclist}{RGB}{0,128,192}
  
  \newrobustcmd{\B}{\bfseries}
  \setlength{\colsep}{10pt}
  \scriptsize 
  \renewcommand{\arraystretch}{1.}
  \resizebox{\linewidth}{!}{%
  \begin{tabular}{@{\hspace{5pt}}
      l@{\hspace{\colsep}} %
      S@{\hspace{\colsep}} %
      S@{\hspace{\colsep}} %
      S@{\hspace{\colsep}} %
      S@{\hspace{\colsep}} %
      S@{\hspace{\colsep}} %
      S@{\hspace{\colsep}} %
      S@{\hspace{\colsep}} %
      S@{\hspace{\colsep}} %
      S@{\hspace{\colsep}} %
      S@{\hspace{\colsep}} %
      S@{\hspace{5pt}} %
      @{\hspace{8pt}}
      S@{\hspace{\colsep}} %
      S@{\hspace{5pt}}}    %
    \toprule
    \B{Model} & \B{\rotatebox[origin=l]{90}{\colorbox{cv-build}{} Building}} 
    & \B{\rotatebox[origin=l]{90}{\colorbox{cv-tree}{} Tree}} & \B{\rotatebox[origin=l]{90}{\colorbox{cv-sky}{} Sky}} 
    & \B{\rotatebox[origin=l]{90}{\colorbox{cv-car}{} Car}} & \B{\rotatebox[origin=l]{90}{\colorbox{cv-sign}{} Sign}} 
    & \B{\rotatebox[origin=l]{90}{\colorbox{cv-road}{} Road}} & \B{\rotatebox[origin=l]{90}{\colorbox{cv-pedestrian}{} Pedestrian}} 
    & \B{\rotatebox[origin=l]{90}{\colorbox{cv-fence}{} Fence}} & \B{\rotatebox[origin=l]{90}{\colorbox{cv-pole}{} Pole}} 
    & \B{\rotatebox[origin=l]{90}{\colorbox{cv-sidewalk}{} Sidewalk}} & \B{\rotatebox[origin=l]{90}{\colorbox{cv-cyclist}{} Cyclist}} 
    & \B{\rotatebox[origin=l]{90}{mIoU w/ MTL}} & \B{\rotatebox[origin=l]{90}{mIoU w/o MTL}}\\
    \midrule    
    \text{ENet~\cite{Paszke2016}} & 72.95334 & 63.58063 & 83.16372 & 75.57105 & 31.03089 & 92.93005 & 41.83943 & 15.35347 & 24.19076 & 76.28467 & 42.67589 & 56.32489 & 51.35547\\    
    \text{DeconvNet~\cite{Noh2015}} & 76.88499 & 67.98886 & 86.90582 & 78.66608 & 27.73828 & 93.54952 & 41.79856 & 26.5194  & 25.82796 & 78.21863 & 46.56049 & 59.15078 & 48.93486\\    
    \text{SegNet~\cite{Badrinarayanan2017}} & 78.16914 & 71.04894 & 88.41527 & 80.65418 & 39.38215 & 93.74948 & 46.88451 & 34.4612  & 28.14128 & 78.94282 & 48.68494 & 62.59399 & 55.69418\\    
    \text{UNet~\cite{Ronneberger2015}} & \B 79.60943 & \B 73.21305 & \B 89.1677  & 81.37496 & \B 42.41067 & 93.8112  & 58.00092 & 32.64772 & 31.34379 & 79.94234 & 47.98074 & 64.50022 & 56.12073\\    
    \text{FCN8~\cite{Long2016}} & 78.84189 & 71.81983  & 85.13701 & \B 84.60183 & 40.69326 & 94.11377 & 54.19309 & \B 40.47746 & 29.34844 & 80.60897 & \B 52.19043 & 64.72963 & 57.09838\\
        \text{CGBNet~\cite{Ding2020}} & 79.34564 & 72.01548 & 85.96541 & 82.43298 & 40.86345 & 94.29751 & 56.48227 & 38.48245 & 31.10543 & 80.72065 & 50.84530 & 64,77787 & 58,86452\\
    \text{FC-DenseNet67~\cite{Jegou2017}} & 79.06759 & 71.38042 & 86.47661 & 84.59334 & 40.4429  & \B 94.4125  & \B 58.09791 & 39.8477  & \B 36.74752 & \B 82.62042 & 50.44985 & \B 65.83061 & \B 65.81933\\
    \bottomrule
  \end{tabular}}
\end{table}

\begin{table}[tb]
  \centering
  \sisetup{
    table-format = 1.2,
  }
  \caption[Results on Cityscapes validation]{IoU results on Cityscapes validation set for semantic segmentation, using $11$ classes and with crop size of $384 \times 768$.
  }
  \label{tab:result_cityscape}
  \definecolor{cs-sky}{RGB}{70,130,180}
  \definecolor{cs-build}{RGB}{70,70,70}
  \definecolor{cs-road}{RGB}{128,64,128}
  \definecolor{cs-sidewalk}{RGB}{244,35,232}
  \definecolor{cs-fence}{RGB}{190,153,153}
  \definecolor{cs-vege}{RGB}{107,142, 35}
  \definecolor{cs-pole}{RGB}{153,153,153}
  \definecolor{cs-car}{RGB}{0,0,142}
  \definecolor{cs-sign}{RGB}{220,220,0}
  \definecolor{cs-person}{RGB}{220,20,60}
  \definecolor{cs-cyclist}{RGB}{119,11,32}
  
  \newrobustcmd{\B}{\bfseries}
  \setlength{\colsep}{10pt}
  \scriptsize 
  \renewcommand{\arraystretch}{1.}
  \resizebox{\linewidth}{!}{%
  \begin{tabular}{@{\hspace{5pt}}
      l@{\hspace{\colsep}} %
      S@{\hspace{\colsep}} %
      S@{\hspace{\colsep}} %
      S@{\hspace{\colsep}} %
      S@{\hspace{\colsep}} %
      S@{\hspace{\colsep}} %
      S@{\hspace{\colsep}} %
      S@{\hspace{\colsep}} %
      S@{\hspace{\colsep}} %
      S@{\hspace{\colsep}} %
      S@{\hspace{\colsep}} %
      S@{\hspace{5pt}} %
      @{\hspace{8pt}}
      S@{\hspace{\colsep}} %
      S@{\hspace{5pt}}}    %
    \toprule
    \B{Model} & \B{\rotatebox[origin=l]{90}{\colorbox{cs-sky}{} Sky}} & \B{\rotatebox[origin=l]{90}{\colorbox{cs-build}{} Building}} & 
    \B{\rotatebox[origin=l]{90}{\colorbox{cs-road}{} Road}} & \B{\rotatebox[origin=l]{90}{\colorbox{cs-sidewalk}{} Sidewalk}} & 
    \B{\rotatebox[origin=l]{90}{\colorbox{cs-fence}{} Fence}} & \B{\rotatebox[origin=l]{90}{\colorbox{cs-vege}{} Vegetation}} & 
    \B{\rotatebox[origin=l]{90}{\colorbox{cs-pole}{} Pole}} & \B{\rotatebox[origin=l]{90}{\colorbox{cs-car}{} Car}} & 
    \B{\rotatebox[origin=l]{90}{\colorbox{cs-sign}{} Sign}} & \B{\rotatebox[origin=l]{90}{\colorbox{cs-person}{} Person}} & 
    \B{\rotatebox[origin=l]{90}{\colorbox{cs-cyclist}{} Cyclist}} 
    & \B{\rotatebox[origin=l]{90}{mIoU w/ MTL}} & \B{\rotatebox[origin=l]{90}{mIoU w/o MTL}}\\
    \midrule  
    \text{ParseNet~\cite{Liu2015}} & 92.68293827404352 & 89.16041572953266 & 96.64929344202108 & 78.68159572042823 & 38.8946520805525 & 90.31107497240238 & 51.25962939829274 & 92.22531410130854 & 69.6257540181677 & 72.5078099997721 & 71.13040897806083 & 76.64808061041657 & 71.02214 \\
    \text{DeconvNet}~\cite{Noh2015} & 93.38068506558454 & 89.30000689030125 & 96.88428559437489 & 77.74536506815247 & 47.09711705555088 & 90.94015731128758 & 53.386026294237446 & 92.32749112042848 & 62.895704180888536 & 70.13062867767482 & 69.2939112131625 & 76.67103440651304 & 62.0281\\
    \text{FCN8~\cite{Long2016}} & 92.48669736904394 & 89.23579597502945 & 96.9382594309139 & 77.98272166561223 & 49.744548065731294 & 90.23699581309369 & 49.0955098719256 & 91.90855149250817 & 65.96284903753138 & 70.76164399020723 & 70.78263914402699 & 76.83056471414763 & 59.9772\\
    \text{FastNet~\cite{Oliveira2016}} & 93.0441852916049 & 89.37268355237003 & 96.95447663381437 & 78.78835404554599 & 48.76638364912157 & 90.3099611595817 & 53.635451996403006 & 92.0867166047084 & 68.71895401365256 & 71.24763462321243 & 69.61882485850919 & 77.50396603895675 & 68.5236\\
    \text{AdapNet++~\cite{Valada2019}} & 93.06606033717834 & 89.46445452619719 & 97.05622901868213 & 80.02680655465022 & 49.46480016553393 & 90.57898007902187 & 52.099869078274374 & 92.22088910324935 & 66.2616099406663 & 72.88264400519273 & 70.6234837933447 & 77.61325696381738 & \B 72.78740\\
    \text{CGBNet~\cite{Ding2020}} & 92.97664103982117 & 89.39958276438676 & 96.66423105831508 & 77.59557837050379 & 42.80119490412156 & 91.88485470080798 & 57.52385700042729 & 91.13894547133451 & 73.28850836068722 & 75.2576623028372 & 71.1815062265886 & 78.15568747271192 & \B 73.2542485117273\\
    \text{FC-DenseNet67~\cite{Jegou2017}} & \B 93.88313998276449 & 89.73299438760934 & 96.81290253288601 & 77.80647258331066 & 49.13706449171094 & 89.94146975843269 & \B 58.74235499720015 & 92.33820177878768 & 66.76975260475419 & \B 75.19007740288495 & 69.6320658658321 & 78.18059058056122 & 72.4968\\
    \text{SegNet}~\cite{Badrinarayanan2017} & 93.74626719289523 & \B 90.08717473914852 & \B 97.37211631563157 & \B 81.57706328313445 & \B 51.8361304877961 & \B 91.75380542403174 & 56.85244795193234 & \B 92.81677417451446 & \B 67.52481330639286 & 72.61162786243042 & \B 72.12037519847499 & \B 78.9362359942166 & 52.1732\\
    \bottomrule
  \end{tabular}}
\end{table}

\begin{table}[tb]
  \centering
  \sisetup{
    table-format = 1.2,
  }
  \caption[Results on Freiburg Forest test]{IoU results on Freiburg Forest test set for semantic segmentation, using $5$ classes and with crop size of $384 \times 768$.
  }
  \label{tab:result_forest}
  \definecolor{cs-trail}{RGB}{170,170,170}
  \definecolor{cs-grass}{RGB}{0,255,0}
  \definecolor{cs-vegetation}{RGB}{102,102,51}
  \definecolor{cs-sky}{RGB}{0,120,255}
  \definecolor{cs-obstacle}{RGB}{255,255,0}
  
  \newrobustcmd{\B}{\bfseries}
  \setlength{\colsep}{10pt}
  \scriptsize 
  \renewcommand{\arraystretch}{1.}
  \begin{tabular}{@{\hspace{5pt}}
      l@{\hspace{\colsep}} %
      S@{\hspace{\colsep}} %
      S@{\hspace{\colsep}} %
      S@{\hspace{\colsep}} %
      S@{\hspace{\colsep}} %
      S@{\hspace{5pt}} %
      @{\hspace{8pt}}
      S@{\hspace{\colsep}} %
      S@{\hspace{5pt}}}    %
    \toprule
    \B{Model} & \B{\rotatebox[origin=l]{90}{\colorbox{cs-trail}{} Trail}} & \B{\rotatebox[origin=l]{90}{\colorbox{cs-grass}{} Grass}} & 
    \B{\rotatebox[origin=l]{90}{\colorbox{cs-vegetation}{} Vegetation}} & \B{\rotatebox[origin=l]{90}{\colorbox{cs-sky}{} Sky}} & 
    \B{\rotatebox[origin=l]{90}{\colorbox{cs-obstacle}{} Obstacle}} 
    & \B{\rotatebox[origin=l]{90}{mIoU w/ MTL}} & \B{\rotatebox[origin=l]{90}{mIoU w/o MTL}}\\
    \midrule    
  \text{FC-DenseNet67~\cite{Jegou2017}} & 79.89327392696255 & 80.37944863230267 & 85.41303105992351 & 92.04258556910936 & 34.61129583135417 & 74.46792700393046 & 73.75705\\
    \text{FCN8~\cite{Long2016}} & 85.1216742415122 & 87.4328769258108 & 89.82046206254512 & 91.89093359920273 & 45.97794104995092 & 80.04877757580434 & 77.49172\\
    \text{ParseNet~\cite{Liu2015}} & 86.29782003254194 & 87.72805334136773 & 90.196326801931 & 91.97294852349603 & \B 47.4098525049095 & 80.72100024084924 & 78.97639\\
    \text{FastNet~\cite{Oliveira2016}} & 86.9035795041595 & \B 88.07845006053368 & \B 90.77433819417125 & \B 92.81743713075622 & 45.88196991314835 & 80.89115496055379 & \B 79.67255\\
    \text{DeconvNet}~\cite{Noh2015} & 87.15929784327722 & 87.41532108960926 & 90.47793020911796 & 92.78637767896278 & 47.16089479254092 & 80.99996432270163 & 78.04483\\
    \text{CGBNet~\cite{Ding2020}} & 87.58712369625532 & 87.61782649884515 & 90.63366521452878 & 92.78478998953157 & 46.58378462516348 & 81.03574632156925 & 77.89452\\
    \text{SegNet}~\cite{Badrinarayanan2017} & \B 88.04198964046398 & 88.04034292874486 & 90.60757490231275 & 92.68127623818356 & 46.215345994324345 & \B 81.1173059408059 & 74.58112\\
    \bottomrule
  \end{tabular}
\end{table}

\begin{figure*}[tb]
  \centering
  \setlength{\hsz}{.12\linewidth}
  \setlength{\colfig}{1pt}
  {\renewcommand{\arraystretch}{0}
    \begin{tabular}{%
        @{}%
        c@{\hspace{\colfig}}
        c@{\hspace{\colfig}}
        c@{\hspace{\colfig}}
        c@{\hspace{\colfig}}
        c@{\hspace{\colfig}}
        c@{}
      }
      \footnotesize\multirow{3}{*}{\raisebox{-1.75\normalbaselineskip}[0pt][0pt]{\rotatebox[origin=c]{90}{Cityscapes dataset}}}
      &\includegraphics[width=\hsz]{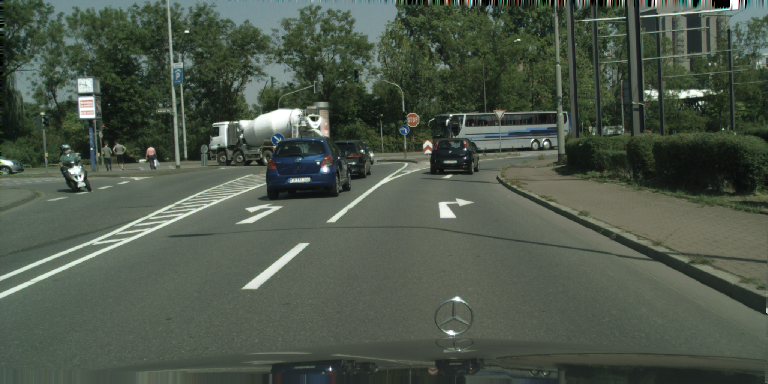} &
      \includegraphics[width=\hsz]{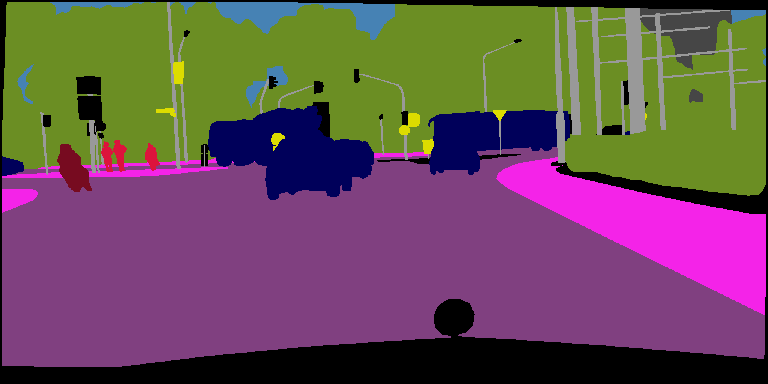} &
      \includegraphics[width=\hsz]{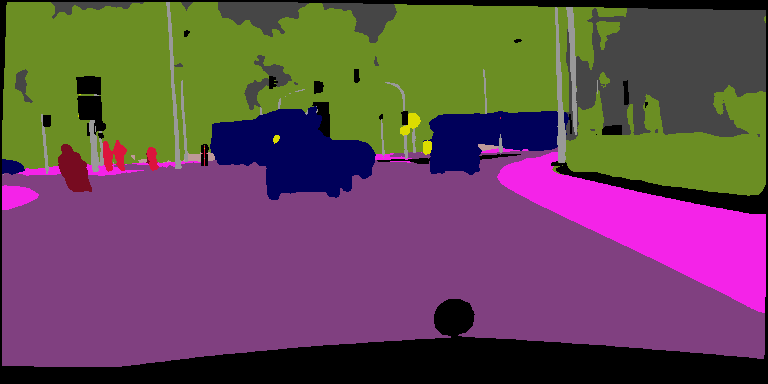} &
      \includegraphics[width=\hsz]{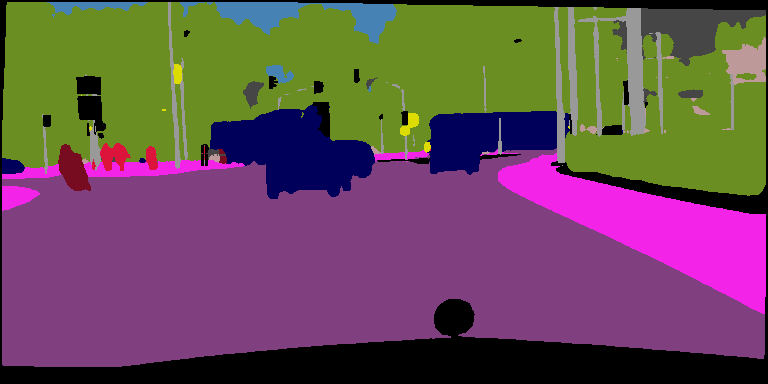} &
      \includegraphics[width=\hsz]{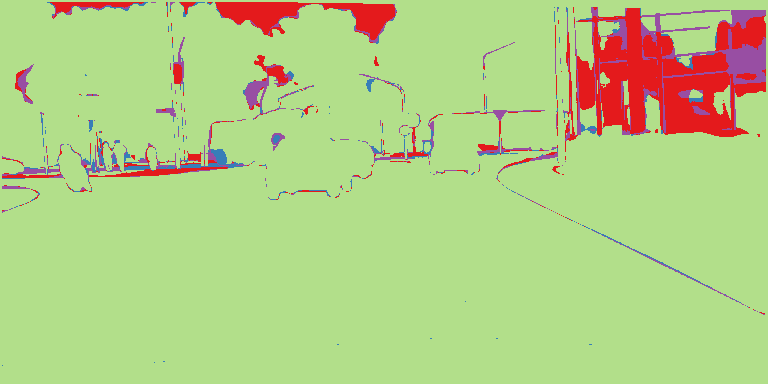}\\%[.2mm]
      &\includegraphics[width=\hsz]{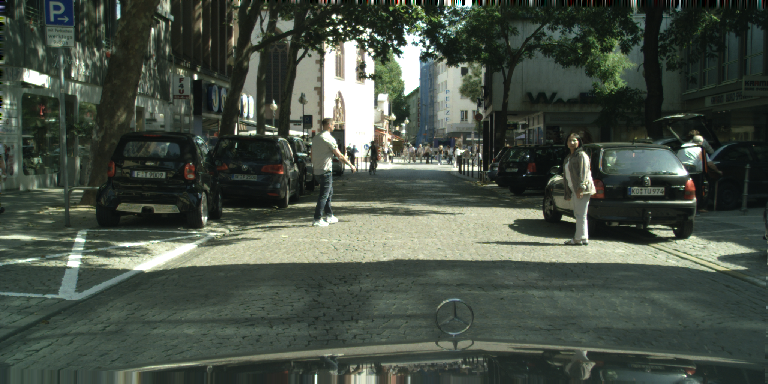} &
      \includegraphics[width=\hsz]{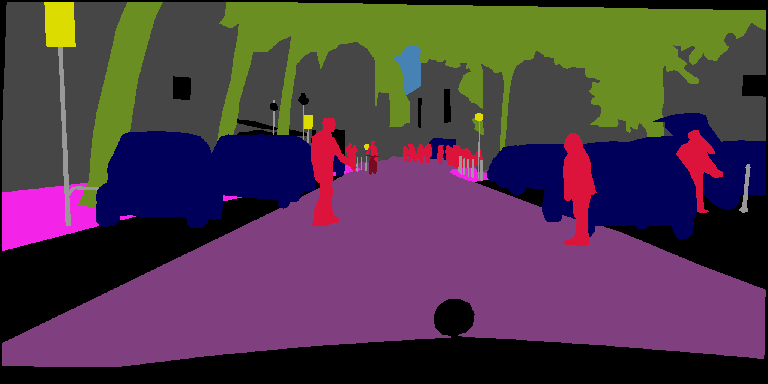} &
      \includegraphics[width=\hsz]{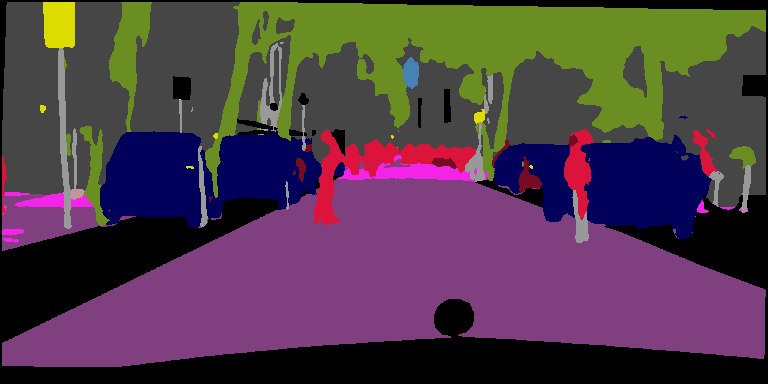} &
      \includegraphics[width=\hsz]{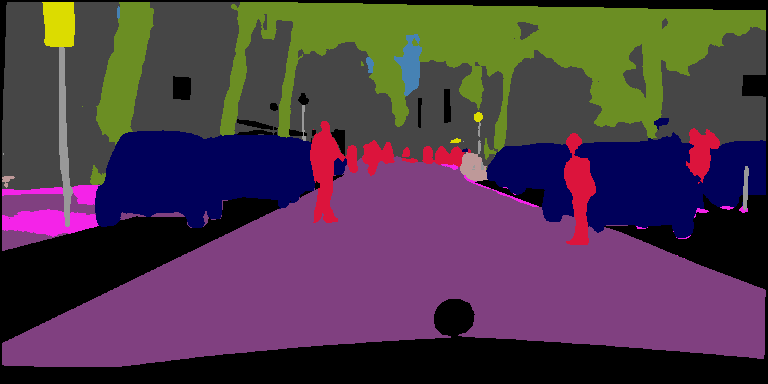} &
      \includegraphics[width=\hsz]{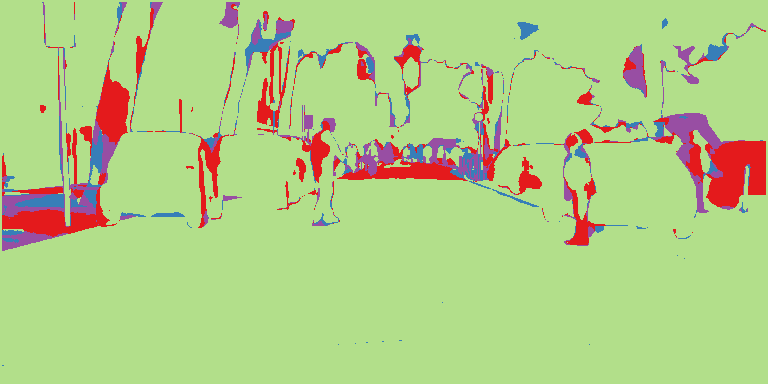}\\%[.2mm]
      &\includegraphics[width=\hsz]{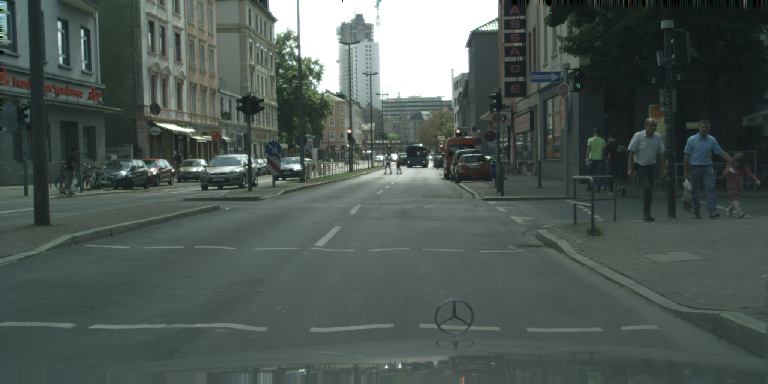} &
      \includegraphics[width=\hsz]{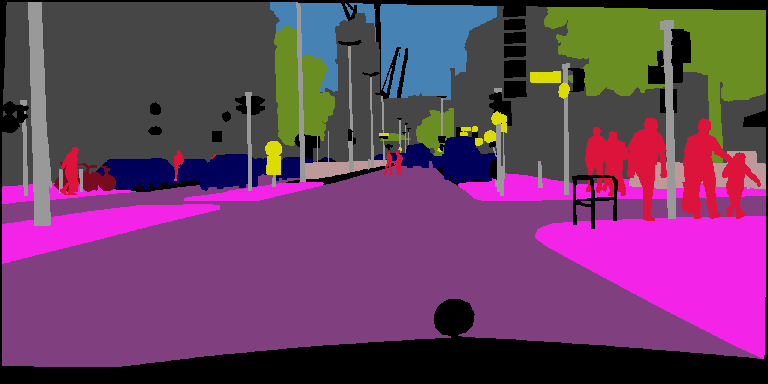} &
      \includegraphics[width=\hsz]{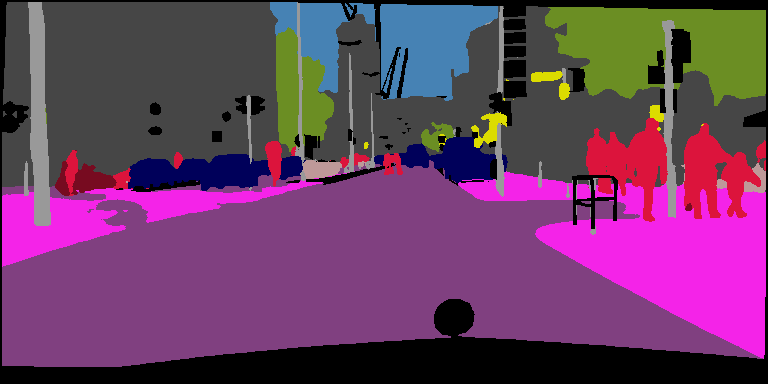} &
      \includegraphics[width=\hsz]{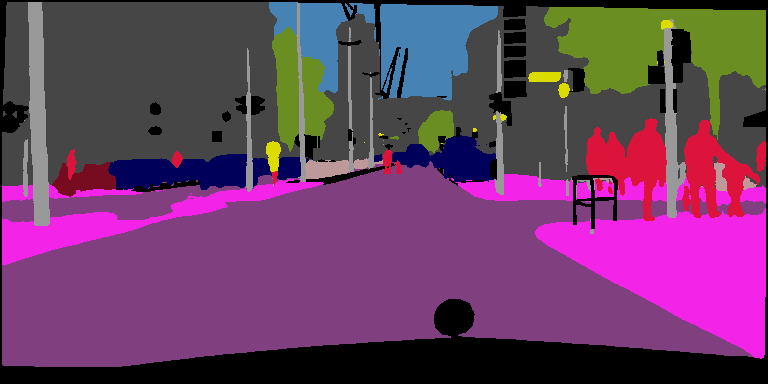} &
      \includegraphics[width=\hsz]{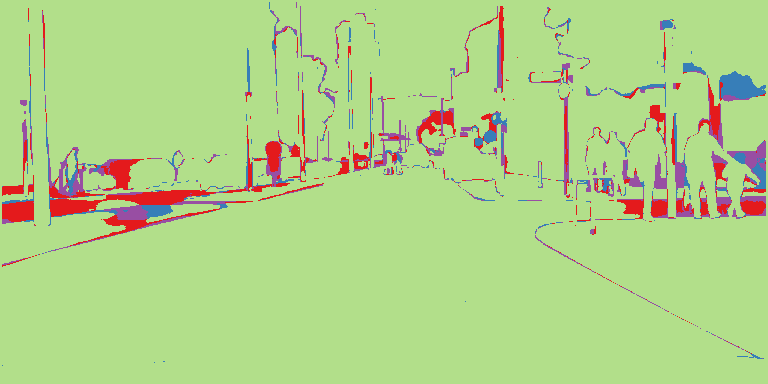}\\[.9mm]
      \footnotesize\multirow{3}{*}{\raisebox{-2.2\normalbaselineskip}[0pt][0pt]{\rotatebox[origin=c]{90}{CamVid dataset}}}
      &\includegraphics[width=\hsz]{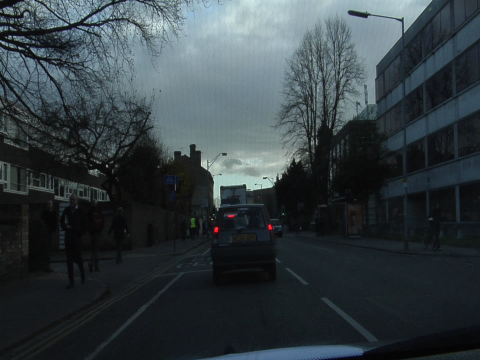} &
      \includegraphics[width=\hsz]{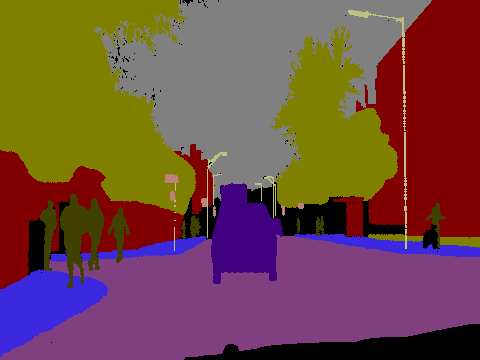} &
      \includegraphics[width=\hsz]{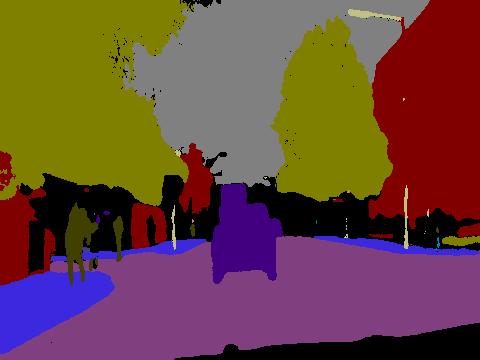} &
      \includegraphics[width=\hsz]{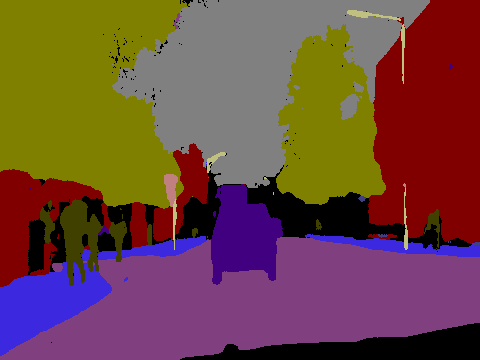} &
      \includegraphics[width=\hsz]{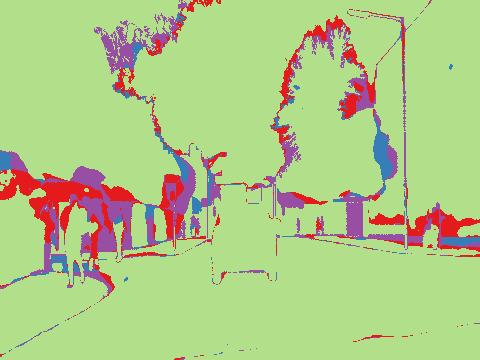}\\%[.2mm]
      &\includegraphics[width=\hsz]{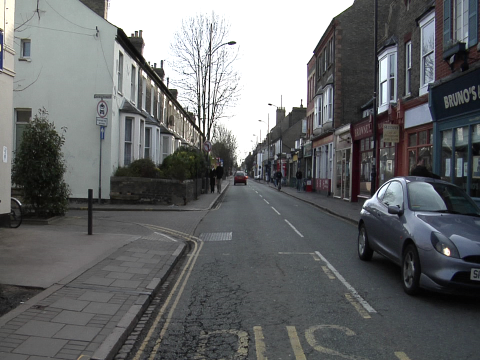} &
      \includegraphics[width=\hsz]{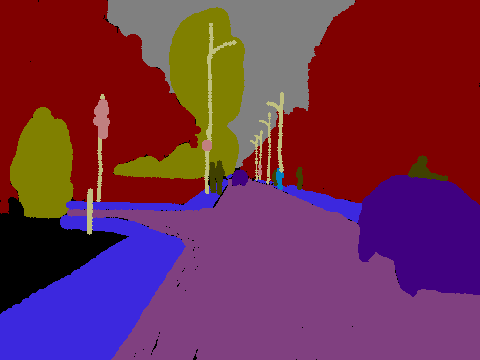} &
      \includegraphics[width=\hsz]{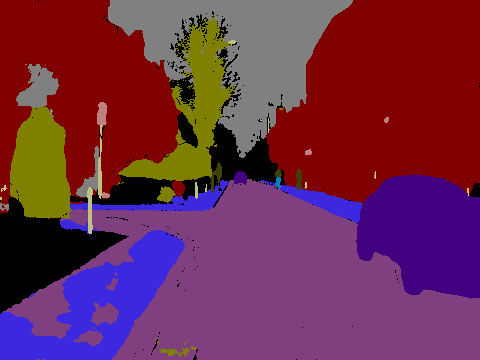} &
      \includegraphics[width=\hsz]{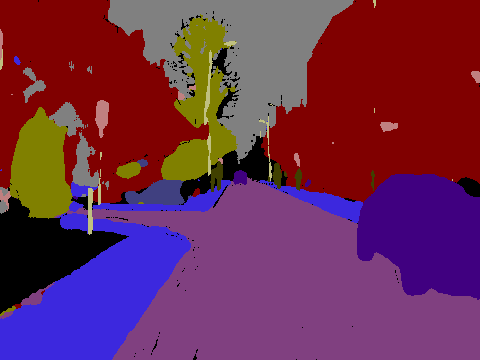} &
      \includegraphics[width=\hsz]{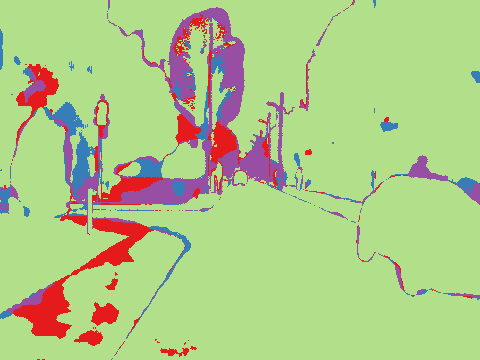}\\%[.2mm]
      &\includegraphics[width=\hsz]{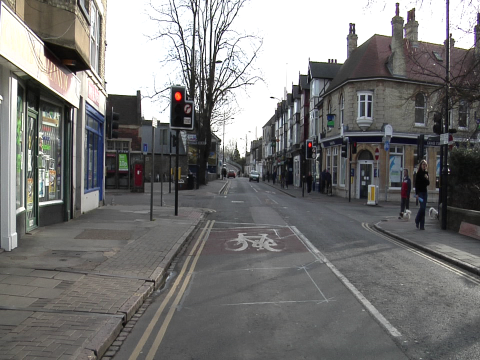} &
      \includegraphics[width=\hsz]{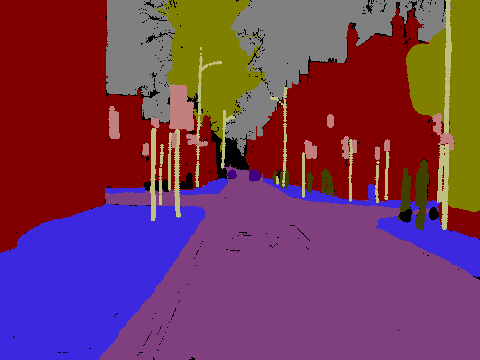} &
      \includegraphics[width=\hsz]{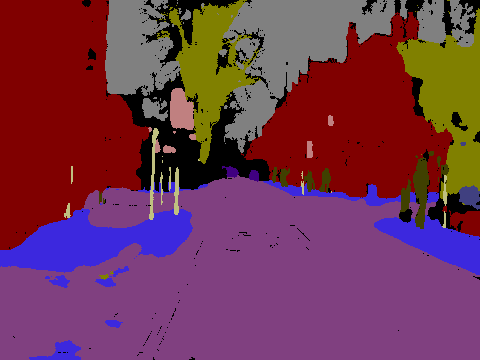} &
      \includegraphics[width=\hsz]{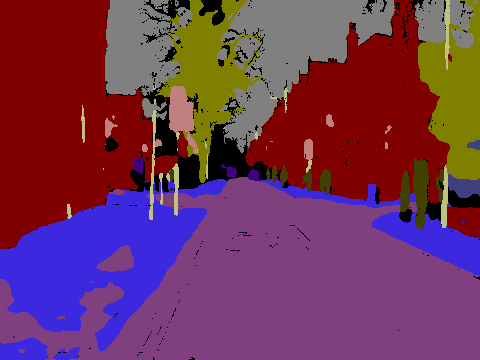} &
      \includegraphics[width=\hsz]{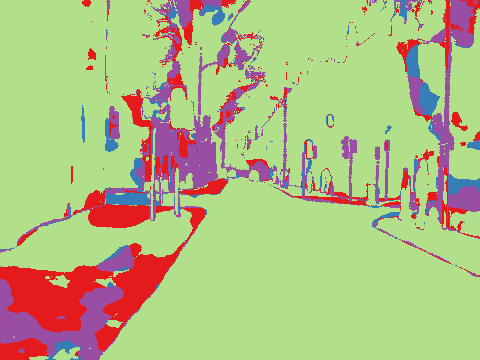}\\[.9mm]
      \footnotesize\multirow{3}{*}{\raisebox{-1.5\normalbaselineskip}[0pt][0pt]{\rotatebox[origin=c]{90}{Freiburg Forest dataset}}}%
      &\includegraphics[width=\hsz]{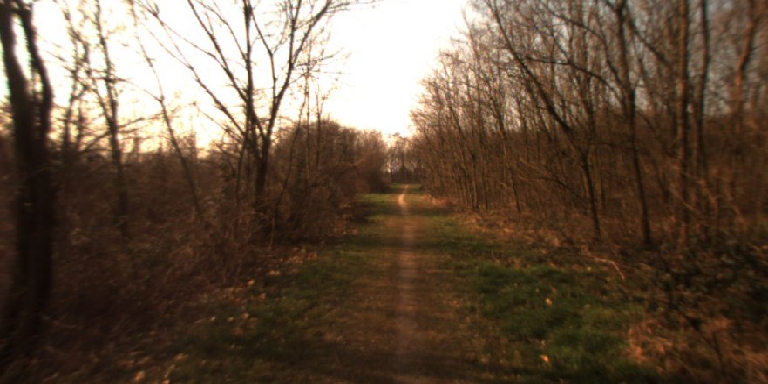} &
      \includegraphics[width=\hsz]{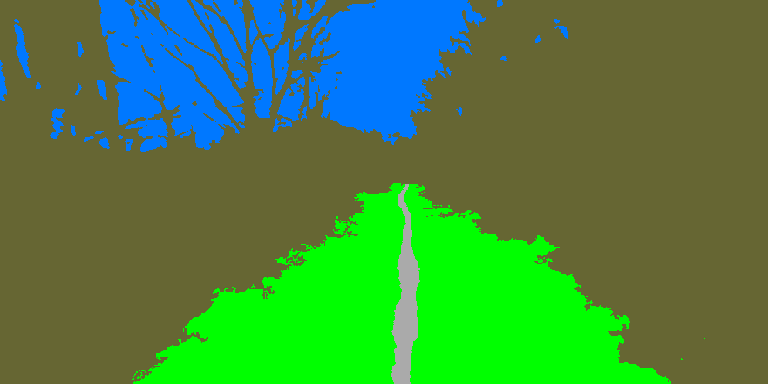} &
      \includegraphics[width=\hsz]{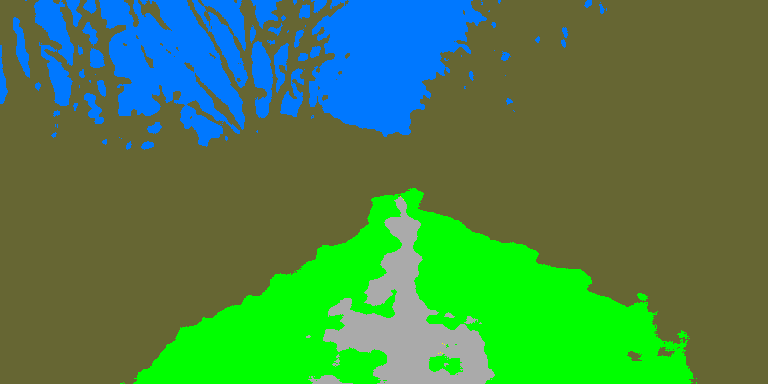} &
      \includegraphics[width=\hsz]{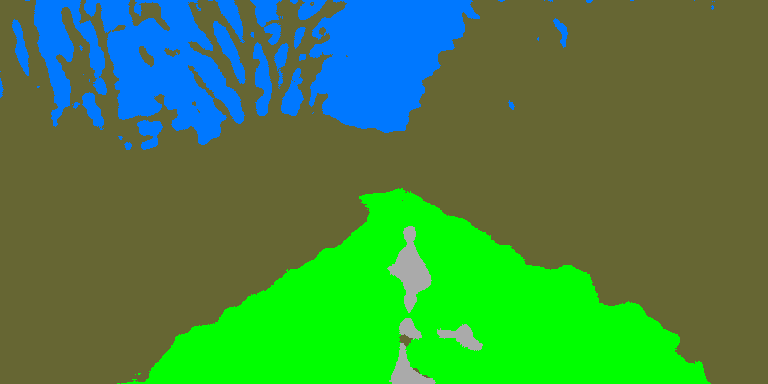} &
      \includegraphics[width=\hsz]{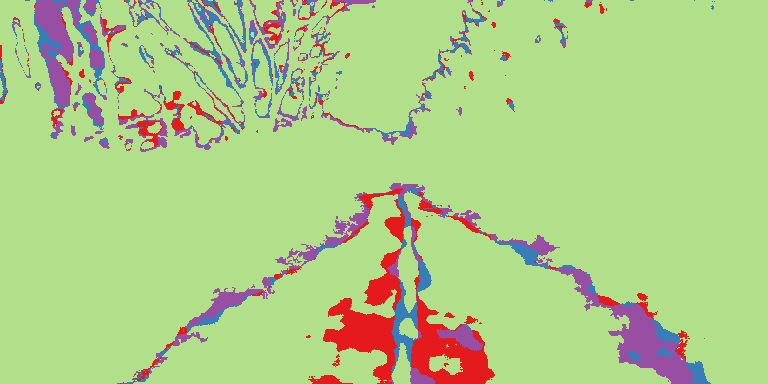}\\%[.2mm]
      &\includegraphics[width=\hsz]{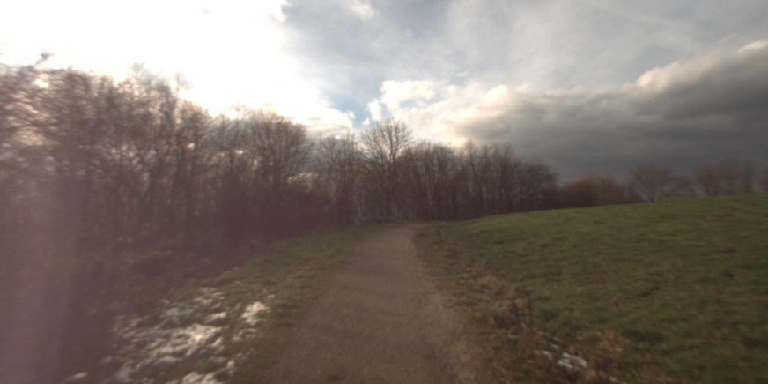} &
      \includegraphics[width=\hsz]{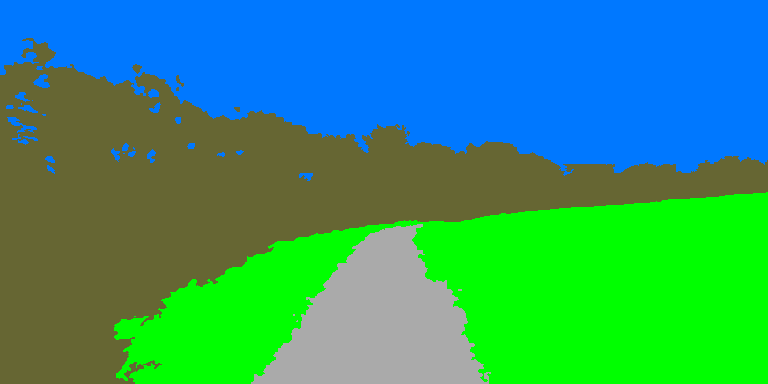} &
      \includegraphics[width=\hsz]{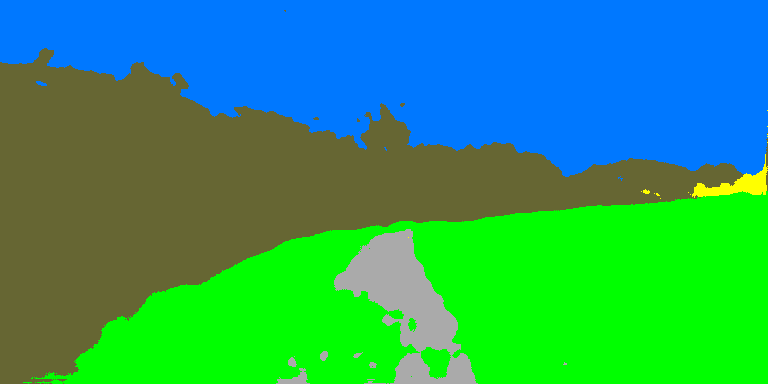} &
      \includegraphics[width=\hsz]{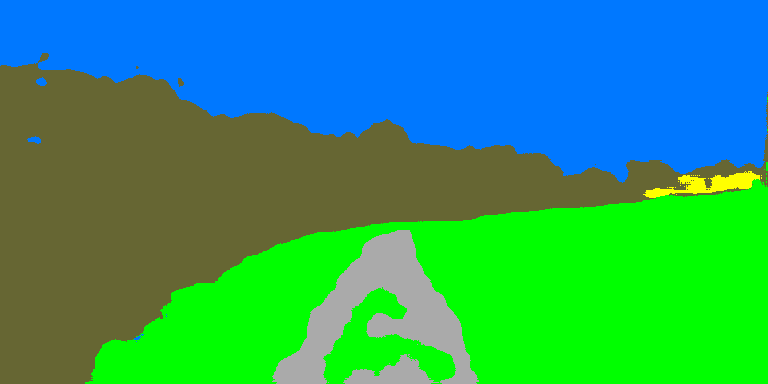} &
      \includegraphics[width=\hsz]{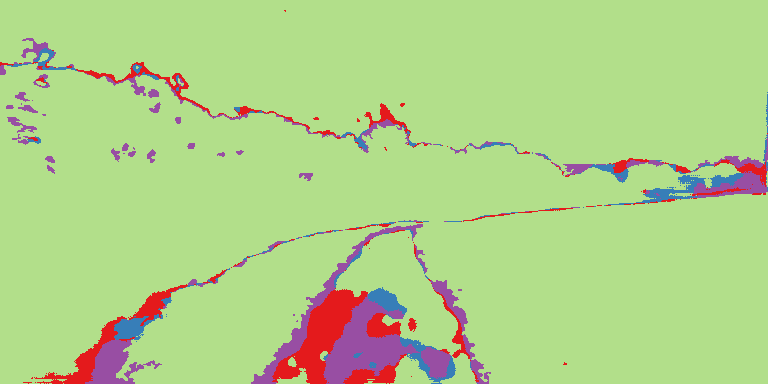}\\%[.2mm]
      &\includegraphics[width=\hsz]{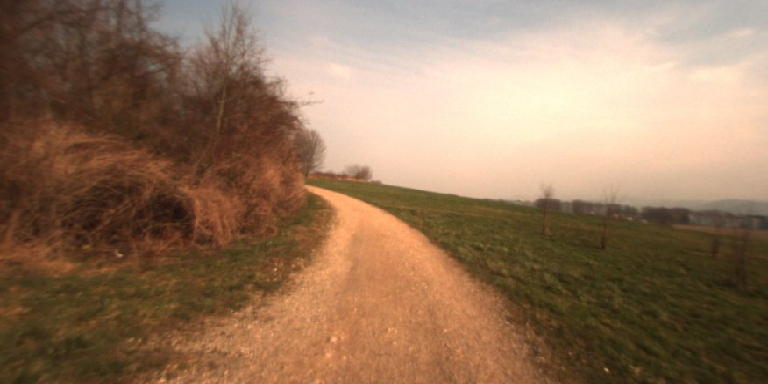} &
      \includegraphics[width=\hsz]{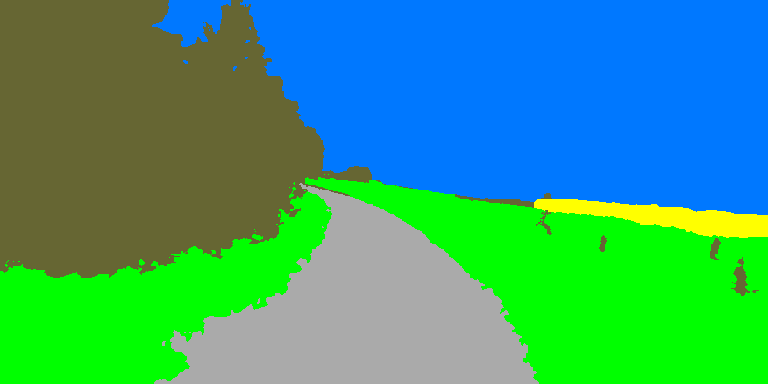} &
      \includegraphics[width=\hsz]{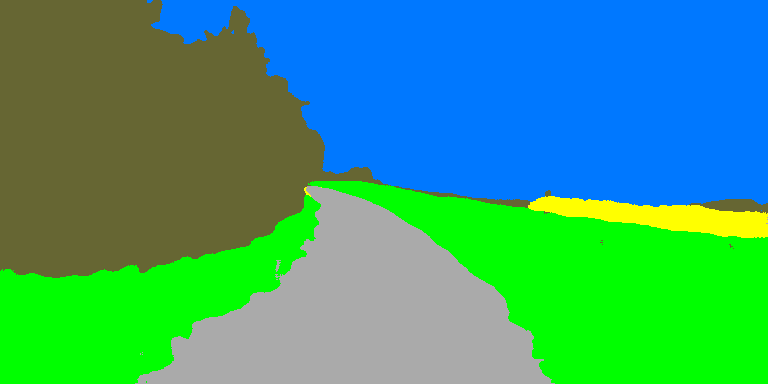} &
      \includegraphics[width=\hsz]{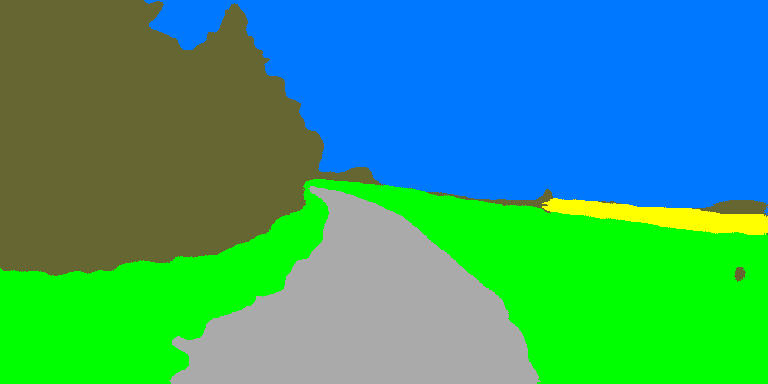} &
      \includegraphics[width=\hsz]{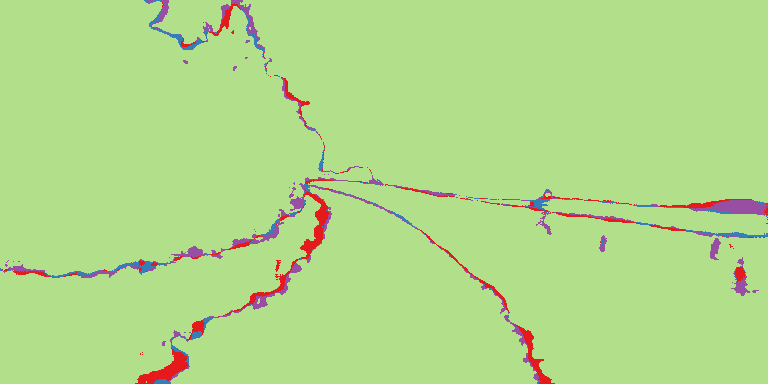}\\[1.2mm]
      &\scriptsize (a) Image &
      \scriptsize (b) Ground truth &
      \scriptsize (c) Prediction &
      \scriptsize (d) Prediction+MTL &
      \scriptsize (e) Comparison\\
    \end{tabular} 
  }
  \caption{Qualitative comparison of the predictions of the AdapNet++~\cite{Valada2019}, UNet~\cite{Ronneberger2015} and FastNet~\cite{Oliveira2016} models (without and with multi-task) against the ground-truth. 
    The comparison column shows an overlap of the predictions against the ground-truth. 
    Note, the green regions are correctly segmented. 
    The red color represents regions erroneously segmented by original models (AdapNet++, UNet, and FastNet), and the blue ones are erroneously segmented by models with multi-task. 
    The regions incorrectly segmented by both predictions are purple.}
  \label{fig:qualitative_cityscape_camvid}
\end{figure*}

In Fig.~\ref{fig:qualitative_cityscape_camvid}, we show qualitative results using the AdapNet++, UNet, and FastNet models.
The columns from left to right represent the ground-truth image, the prediction without and with MTL, and a comparison (overlap) of both prediction maps. 
In the comparison image, the green regions are correctly segmented.
The red color represents regions erroneously segmented by original models (AdapNet++, UNet, or FastNet), and the MTL models erroneously segment the blue ones.
The regions incorrectly segmented by both predictions are purple.

Our experimental results show that adding a multi-task approach to the already defined hourglass models improves the SS task's performance. 
It is important to note that we add contour-based auxiliary tasks because the original models still exhibit the problem of spatial precision loss.
As we saw, this problem is reflected in the boundary of the segmented objects (see Fig.~\ref{fig:SS_loss_spatial_precision}).

\makeatletter
\begin{figure}[tb]
\includegraphics[width=\linewidth]{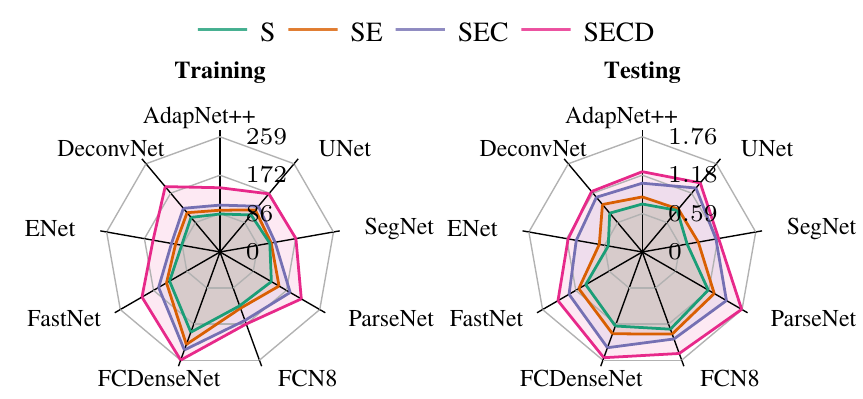}
\caption{%
  Comparison of the execution time of hourglass models (for a growing combination of tasks, \ie, S, SE, SEC, SECD) in the training phase (left) and in the testing phase (right), both in minutes.
  We use the Cityscapes dataset with \num{17500} and \num{500} samples for training and testing, respectively.
  Note that the auxiliary tasks are only used in the training phase. 
  In the testing phase, we only use the time shown in blue (right).
}
\label{fig:time_exec}
\end{figure}
\makeatother

Also, we present an efficiency comparison of the hourglass models (see Fig.~\ref{fig:time_exec}) in the training and testing phase.
We used the Cityscapes dataset with \num{17500} training samples and \num{500} testing samples (in our case, they are the validation samples) for these experiments.
The results show the execution time of one epoch (a forward-pass over the entire dataset) in a single GPU\@. 
Note that we only do this for an increasing sequence of combinations: S, SE, SEC, and SECD, due to our computational limitations.
We executed the process five times and report the averages of the training (left) and testing phase (right), both in minutes.
Our graphics on training using multitasking show an increase in the time required to train hourglass models. 
This increment is directly related to the complexity of the contour-based auxiliary tasks, which proved to be challenging enough to fit a latent space.
Remember, for testing; we only use a single task: semantic segmentation. 
In other words, for \num{500} samples, the models need the time presented by the segmentation plot~(S) in the right of Fig.~\ref{fig:time_exec}.
Finally, the previous experiments have been conducted on NVIDIA GTX Titan~X with \SI{12}{\giga\byte} of memory and four GPUs (multi-GPU).

\subsection{Discussion}
\label{sec:discussions}
In the previous experiments, we showed that by learning multiple contour-based tasks on the hourglass models the redundant information needed to solve the tasks improves the models' learning.
We noticed that the tasks increase the models' ability to accommodate noise during the training phase.
Consequently, the tasks reduce the model's overfitting risk by providing a gradient that tends to keep the latent space from overfitting.
This advantage in the feature space is largely due to the clustering behavior of latent space that we achieved by a more robust feature extraction.

The addition of auxiliary tasks changes the weight updating dynamics (\ie, gradient updating) such that the model learns robust features that work in all the used tasks. 
The robustness of the space comes from adding related tasks to the blob prediction one (\ie, the SS task). 
Hence, by unsupervisedly restricting the latent space through more tasks we robustify the latent space, as show by our experiments.

\begin{figure}[tb]
  \centering
  \begin{tikzpicture}
    \def\ca{( 45:.7) circle (1)}
    \def\cb{(135:.7) circle (1)}
    \def\cc{(225:.7) circle (1)}
    \def\cd{(315:.7) circle (1)}
    
    \path[name path=tsp, draw] \ca node[above right] (ts) {$\mathcal{T}_S$};
    \draw \cb node[above left] (te) {$\mathcal{T}_E$};
    \draw \cc node[below left] (td) {$\mathcal{T}_D$};
    \draw \cd node[below right] (tc) {$\mathcal{T}_C$};
    
    \node[below left=20pt and 10pt of td, font=\footnotesize, anchor=west] (lbl) {Learned feature space};
    
    \coordinate (bbsw) at (current bounding box.south west);
    \coordinate (bbne) at (current bounding box.north east);
    
    \begin{scope}%
      \clip \ca;
      \clip \cb;
      \clip \cc;
      \fill[gray] \cd;
      \node (int) {};
    \end{scope}
    
    \draw (int) -- ($(ts)!.5!(tc)+(1,0)$) node[font=\footnotesize, anchor=west] {Best features};
    
    \pgfresetboundingbox
    \path[use as bounding box] (bbsw) rectangle (bbne); 
  \end{tikzpicture}
  \caption{Example of the latent space (\ie, feature representations) learned by backpropagation when using several auxiliary tasks.
  The representations at the intersection of all the tasks are better since they satisfy several tasks simultaneously.
  }
  \label{fig:latent_representation_space}
  \vspace{-10pt}
\end{figure}
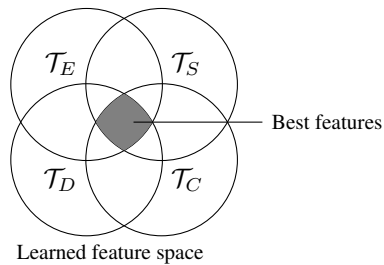

For example, one of the main problems with SS methods is to correctly predict the boundaries of the objects since most of the accuracy comes from correctly detecting the main blob.
By adding a contour-based auxiliary task, we increase the learning rate's effectiveness for this case.
This increment happens since the same model is forced to understand the boundaries of the objects to predict the contours while been asked to predict the blobs as well (through the other task).
We found that by simultaneously learning to solve related tasks the learned features improve with the tasks added (\cf Fig.~\ref{fig:latent-space-tasks}).
This result is intuitive if we assume that the different tasks have a common latent space, as show in Fig.~\ref{fig:latent_representation_space}.
Then, by simultaneously optimizing in this shared space, our multi-task learning problem is finding solutions in the intersection of the tasks which in turn improves the others.

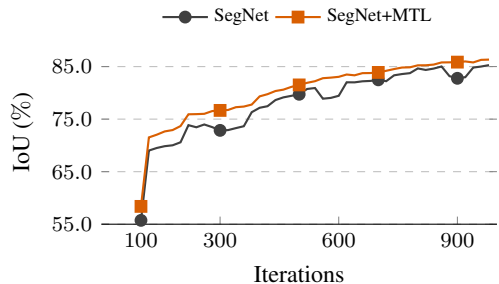
\begin{figure}[tb]
  \centering
  \begin{tikzpicture}%
  \begin{groupplot}[
  group style={
    group size=3 by 1, 
    xlabels at=edge bottom,
    ylabels at=edge left,
    x descriptions at=edge bottom,
    y descriptions at=edge left,
    horizontal sep=.1cm,
  }, 
  footnotesize,
  height=4cm,
  width=6.8cm,
  cycle list/Paired,
  cycle multiindex* list={%
    [2 of]mark list\nextlist
    black!75, Dark2-B\nextlist
  },
  /tikz/mark repeat=10,
  /tikz/mark phase=1,
  ymajorgrids,
  major grid style={dashed},
  ylabel={IoU ($\%$)},
  xlabel={Iterations},
  xtick={100,300,600,900},
  xmax=1000,
  xmin=0,
  ytick={55,65,75,85},
  ymax=90,
  ymin=55,
  x tick label style={/pgf/number format/precision=0},
  legend pos=outer north center,
  legend columns=2,
  legend style={
    cells={anchor=west},
    font=\scriptsize,
    draw=none,
  },
  ]
  
  \nextgroupplot[%
  ]%
  \addplot table[x=iter, y=iou, col sep=comma, header=true]{img/stable_SegNet_S.csv};%
  \addlegendentry{SegNet}%
  
  \addplot table[x=iter, y=iou, col sep=comma, header=true]{img/stable_SegNet_SBCE.csv};%
  \addlegendentry{SegNet+MTL}%
  
  \end{groupplot}%
  \end{tikzpicture}%
  \caption[stable training]{%
    IoU over the iterations in the training phase of the hourglass model SegNet~\cite{Badrinarayanan2017} focused on the semantic segmentation task in the CamVid dataset. 
    By adding contour-based auxiliary tasks in the training phase, we achieve a robust feature extraction that increases the ability to treat noise and reduces the risk of overfitting.
    In consequence, the model that uses the MTL achieves a more stable learning than its counterpart.
  }
  \label{fig:stable_training}
  \vspace{-10pt}
\end{figure}

When training multiple tasks simultaneously, from the model point of view,  the hidden units in the hourglass models improve two-fold:
(i)~There is a more significant number of parameters involved in updating the weights (\ie, backpropagation) using MTL, and 
(ii)~the most relevant parameters (\ie, the most frequently influenced by all tasks) achieve a better (robust) feature extraction. 
This adjustment on the parameters of the MTL hourglass models has a regularization effect on the parameters in addition to a more stable training (\ie, the variance in the loss function plot shown in Fig.~\ref{fig:stable_training}).

Unlike other architectures (\eg, DeepLabV3), hourglass models have a decoding stage that is complex enough (\ie, a set of reconstruction operations) to highlight the changes produced in the latent space. 
For this reason our research is focused on hourglass models.
Note that each addition to the auxiliary tasks has a different effect on what is learned in the latent space. 
Changes in architecture can alter, in different ways, the way backpropagation benefits from contour-based auxiliary tasks.

\section{Future Work}
\label{sec:Future_work}
In our work, we observed the clustering behavior of the latent space. Future work may focus on using a clustering framework to impose particular biases to learn the latent representations.
By forcing the latent space into clusters, we intuit that we will need fewer tasks in the training phase.
Another venue to explore is the influence of the auxiliary contour-based tasks on architectures other than hourglass-based.

Regarding extending our work to videos, on one hand, we need to extend the architectures to work with 3D data.  Thus, we will need to use bigger models that rely on 3D convolutions to perform the segmentation on volumetric data.  On the other hand, we will need to maintain not only spatial consistency but also temporal one.  This new constraint will be akin the problems we face today trying to reduce the instabilities in the segmentation boundaries.  Hence, future work will need to find relevant tasks that help to stabilize the temporal consistencies as well.  Perhaps, we could explore optical flow as a first approach to tackle since it is similar to the spatial boundaries.

\section{Conclusions}
\label{sec:Conclusions}
In this paper, we incorporated auxiliary contour-based tasks to address the loss of spatial precision. 
This problem is commonly in bounding segmented objects. 
Thus, we propose to use edge detection and semantic contour tasks to reinforce the semantic information on the boundary objects. 
We also proposed using quantized distance transform to add geometric information into the internal representation of deep neural networks (\ie, hourglass models).
We observed (by empirical experiments) that the latent space behavior in hourglass models is clustering when adding complementary information (due to auxiliary tasks). 
Note that the latent space does not present a random distribution. 
Instead, better-distributed clusters produce, in turn, better segmentation results.
We also showed that when using all the tasks, the activation maps (regions used by the networks to perform the segmentation prediction) adjust better to the edges of objects. 
Although the activation maps vary depending on the input image, the latent space's behavior produces an improvement in the quality of segmentation.
Additionally, we verify (empirically) that the improvement produced by using multiple tasks addresses the problem of loss of spatial precision in segmentation. 
In other words, we verified (by using trimap) that using the clustered latent space improves the edges of the segmented objects and, consequently, the final segmentation.
We also interpret that by adding contour-based auxiliary tasks, the models obtain a more powerful generalization.
In order not to limit our study (by using three models), we compared the results of the different hourglass models (with and without MTL) existing in the literature on other datasets (Cityscapes and Freiburg Forest).
Finally, the empirical exploration showed that it is possible to better fit the models (obtain a latent space with cluster behavior) for the semantic segmentation task when we use complementary information (by adding contour-based auxiliary tasks) in the training phase. 

\input{appendix}

\bibliographystyle{IEEEtran}
\footnotesize
\bibliography{abrv,IEEE_references}

\vspace*{-35pt} 
\begin{IEEEbiography}[{\includegraphics[width=1in,height=1.1in,clip,keepaspectratio]{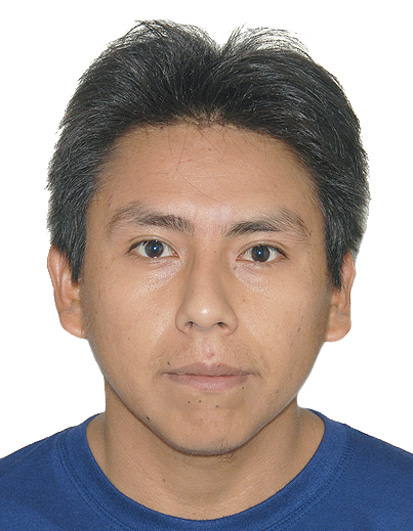}}]{Darwin Saire} received his B.Eng.\ degree in Computer Engineering from Universidad de San Agustin (UNSA), Arequipa in 2013.  He completed his M.Sc.\ degree in Computer Engineering from University of Campinas, Brazil in 2017.  He is currently a Ph.D.\ candidate at the Institute of Computing, University of Campinas, Brazil.  His research interests are computer vision, pattern recognition, image processing, machine learning and deep learning.
\end{IEEEbiography}
\vspace*{-35pt} 
\begin{IEEEbiography}[{\includegraphics[width=1in,height=1.1in,clip,keepaspectratio]{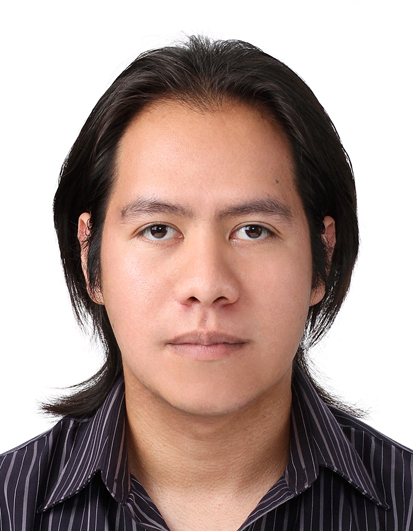}}]{Ad\'{i}n Ram\'{i}rez Rivera} (S'12, M'14) received his B.Eng.\ degree in Computer Engineering from Universidad de San Carlos de Guatemala (USAC), Guatemala in 2009.  He completed his M.Sc.\ and Ph.D.\ degrees in Computer Engineering from Kyung Hee University, South Korea in 2013.  He is currently an Assistant Professor at the Institute of Computing, University of Campinas, Brazil.  His research interests are video understanding (including video classification, semantic segmentation, spatiotemporal feature modeling, and generation), and understanding and creating complex feature spaces.
\end{IEEEbiography}
\vspace*{-10pt}

\EOD
\end{document}

%% file: appendix.tex
\appendices
\counterwithin{figure}{section}
\counterwithin{table}{section}
\counterwithin{equation}{section}

\section{Final Representations}
\label{sec:apx_final_representation}
The methodology proposes to combine information on similar tasks using supervised learning. 
Then, we need to know the operations used to obtain the comparison masks in the different tasks.
Keep in mind; all datasets perform the same preprocessing to obtain the edges and the quantized distance transform.
Thus, to obtain the objects' edges, we take the instances' masks, and we look for a difference of instance labeling. 
For this, we used D-4 connectivity (up, down, right, left)~\cite{Gonzalez2006}, and to better highlight the boundaries, we use the morphology operation of dilation~\cite{Gonzalez2006}, with a structural element of $2$-size and disk-shaped.

On the other hand, for adding geometric information, we extract the distance of pixels to objects' boundaries (\ie, distance transform~\cite{Gonzalez2006}).
Using this distance transform as a task learning gives us the following advantages:
i)~we can easily extract it from the instance masks, and
ii)~the quantized distance transform can be easily trained with the existing loss functions.
Note that this representation, based on the distance transform, allows us to infer the complete shape of an object instance even with incomplete information (\ie, when a part of the object is shown). 

\begin{figure*}[tb]
\centering
\subfloat[Image]{\includegraphics[width=1.7in, height=2.5cm]{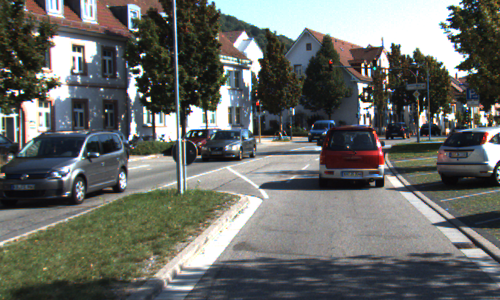}%
  \label{fig:energy_img}}
\hfil
\subfloat[Distance Transform]{\includegraphics[width=1.7in, height=2.5cm]{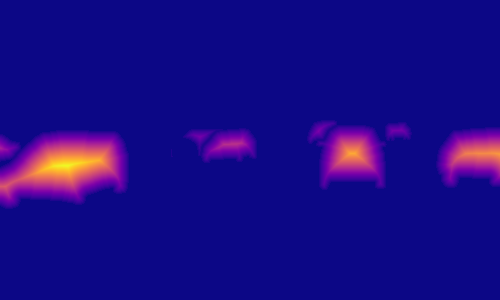}%
  \label{fig:dist_tranf}}
\hfil
\subfloat[Truncate Distance]{\includegraphics[width=1.7in, height=2.5cm]{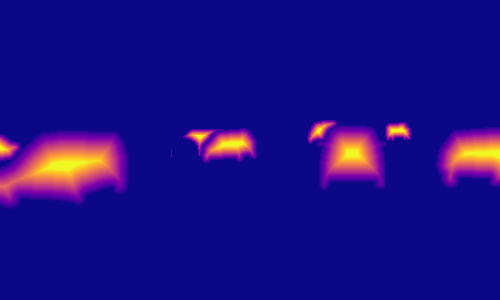}%
  \label{fig:dist_trunc}}
\hfil
\subfloat[Quantized Distance]{\includegraphics[width=1.7in, height=2.5cm]{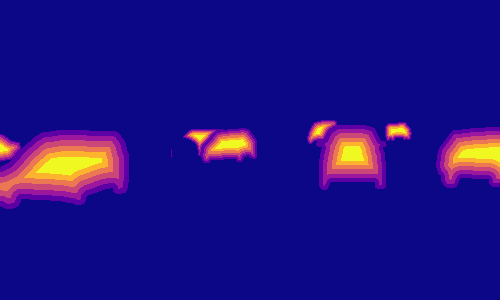}%
  \label{fig:dist_quant}}
\caption{%
  Intending to merge semantic and geometric information, we use a (d)~quantized distance obtained from an (b)~Euclidean distance transform. 
  To enhance the distance transform energy, (c)~we use truncation and normalization. 
  Note that this quantized distance is easy to combine with the multi-label cross-entropy loss function.}
\label{fig:energy_level}
\end{figure*}

The distance transform produces a wide range of values when objects have different shapes and sizes, see Fig.~\ref{fig:dist_tranf}.
For this reason, we truncate the transformation given a threshold $R$, thus guaranteeing a limited range of values, see Fig.~\ref{fig:dist_trunc}.

Therefore, similar to models~\cite{Hayder2017,Bischke2019}, we define $Q$ as the set of pixels on the object boundary the object and $\text{IS}_i$ the set of pixels belonging to instance mask $i$. 
For every pixel $p$, we compute a truncated distance $D_t(p)$ to $Q$ as,
\begin{equation} \label{eq:dist_transf}
D_t(p) = \gamma_p \min \big( \min \lceil d(p,q) \rceil, R \big), \quad \forall\,q \in Q
\end{equation}
where $d(p, q)$ represent the Euclidean distance between pixel $p$ and $q$, $\lceil z \rceil$ give us the nearest integer larger than $z$, and $R$ is the truncation threshold.
Finally, the function $\gamma_p$ denotes if the pixel $p$ is inside or outside of an $\text{IS}_i$ instance mask
\begin{equation}
\gamma_p =\begin{cases}
1, & \text{if} \ p \in \text{IS}_i,\\
0, & \text{otherwise}.\\
\end{cases}
\end{equation}

To facilitate the energy labeling (\ie, continuous distance values), we quantify these values in $K$ uniform bins by one-hot encode the distance map into a binary vector representation $b(p)$ as~\cite{Hayder2017}
\begin{equation} \label{eq:dist_quantized}
D_q(p)\sum_{k=1}^K r_n b_k(p), \qquad \sum_{k=1}^K b_k(p) = 1,
\end{equation}
where $r_n$ is distance value corresponding to bin $k$. The $K$ binary maps are the classification maps for each of the $k$-th edge distance.
We can see an example in Fig~\ref{fig:dist_quant}.

This quantized distance transform operation is not new, having been explored in previous models~\cite{Hayder2017,Bischke2019}.
However, contrary to others that use this transformation in a bounding box~\cite{Hayder2017} or for a single class (\ie, buildings)~\cite{Bischke2019}, our technique applies the transformation for instances that belong to different classes.

\section{Class Imbalance}
\label{sec:apx_imbalance}

In this work, we have class imbalance during training.
The dataset imbalance causes (i)~inefficient training, because it has few samples (of some kinds) in the training stage, the network may not observe all the samples; and, by having a small number of samples, (ii)~the network can fall into overfitting and degenerate the model.
To address this problem, we use median frequency (counting) balancing~\cite{Eigen2016}, defined by
\begin{equation} \label{eq:class_balancing}
\tau_c = \frac{\bar{f}}{f(c)},
\end{equation}
where $f(c)$ is the number of pixels of class $c$ divided by the total number of pixels in images where $c$ is present, and $\bar{f}$ is the median of these frequencies (counting).
Finally, we use this class weighted in Sections~\ref{sec:apx_train_ss} with $\alpha_i$ and~\ref{sec:apx_train_el} with $\mu_i$.

\section{Learning Multi-Task Framework}
\label{sec:apx_training}
We used several hourglass networks based on the MTL approach, where tasks help each other adjust their parameters. 
With this approach, we get a good delimitation of the objects' edge by sharing the information extracted from all the tasks (\ie, share the parameters).  
In the last two layers of the decoding stage, we extract specific information to discriminate each task.
Next, we explain details about the output learning using MTL for edge detection, semantic segmentation, semantic contours, and truncated distance transform (energy level).

\subsection{Edge Detection Training}
\label{sec:apx_train_bound}
In the first specific decoding stage, we learn to detect the edges of each instance object.
In order to handle the imbalance between the two binary classes (edge, no edge), we used the HED-loss function~\cite{Xie2015} a class-balanced cross-entropy function.  
Then we consider the edge-class objective function as
\begin{equation} \label{eq:hed_loss}
\begin{aligned}
\mathcal{L}_{\mathit{c}} =& -\beta \sum_{i \in Y_+} \log P \left( y_i=1 \given X; \theta \right) \\
& - (1-\beta) \sum_{j \in Y_-} \log P \left( y_j=0 \given X; \theta \right),
\end{aligned}
\end{equation}
where $y_i$ and $y_j$ are the indexed predicted edge (output) for the $i$-th and $j$-th pixel, respectively. 
Here, $\theta$ represents the network parameters to be optimized in the edge stage.
The proportion of positive (edge) and negative (no edge) classes on the ground-truth edges $Y$ are $\beta = |Y_+|/|Y|$ and $1-\beta = |Y_-|/|Y|$, where $Y = Y_+ \cup Y_-$.
Moreover, $P$ is the probability that a pixel contained edges (output of edge stage), this is defined by a sigmoid function, such that
\begin{equation} \label{eq:prob_output}
P = P \left( y_i \given X; \theta \right) =  \sigma(y_i) \in [0,1].
\end{equation}

Although the HED-loss function proved to be useful in training for edge detection.
Training time can be reduced and edges further penalized by maximizing intersection-over-union~\cite{Csurka2013}. 
Then, we consider the objective function,
\begin{equation}\label{eq:iou_loss}
\mathcal{L}_{\mathit{iou}} = 1 - \frac{P \cap Y}{P \cup Y} = 1 - \frac{\sum_{v \in Y} P_v  Y_v}{\sum_{v \in Y} P_v  + Y_v - P_v Y_v}.
\end{equation}

Finally, for edge detection, we combine both loss functions to obtain our final objective function, defined by
\begin{equation} \label{eq:edge_loss}
\mathcal{L}_E = \psi_1 \mathcal{L}_{\mathit{c}} + \psi_2 \mathcal{L}_{\mathit{iou}},
\end{equation}
where $\psi_1$ and $\psi_2$ are hyper-parameters that define the contribution of each loss to the learning process.

\subsection{Semantic Segmentation Training}
\label{sec:apx_train_ss}

In the second specific decoding stage, we learn to classify each object in pixel-wise level (\ie, semantic segmentation).
We use a multi-label balanced cross-entropy loss function to address the problem of imbalance. 
Thus we define this function as,
\begin{equation} \label{eq:cross_loss_s}
\mathcal{L_\mathit{cross\ ss}} = -\frac{1}{N}\sum_{i=1}^N \alpha_i \log P(s = s_i \given X; \phi),
\end{equation}
where $s_i$ is the indexed predicted classification (output) for the $i$-th class from the set of ground-truth $S$ on semantic segmentation, additionally, $N$ is the number of classes, and $\phi$ denotes the network parameters to optimize in the semantic segmentation stage. 
Also, $P(\cdot)$ is the probability that a pixel belongs to the $i$th class. 
Similar to~\eqref{eq:prob_output}, this function is defined by a sigmoid activation function.

Besides, similar to the previous section, we use a target function intersection-over-union to penalize the boundary of segmentation.
Contrary to $\mathcal{L_\mathit{iou}}$~\eqref{eq:iou_loss}, at this stage, we use a multi-label function, defined by
\begin{equation}\label{eq:multi_iou_s}
\mathcal{L}_{\mathit{iou\ ss}} = 1 - \sum_{i=1}^N\frac{P_i \cap S_i}{P_i \cup S_i}.
\end{equation}

Subsequently, we combine both loss functions in our final objective function for semantic segmentation. 
Thus, we define the function as,
\begin{equation} \label{eq:ss_loss}
\mathcal{L}_S = \psi_3 \mathcal{L}_{\mathit{cross\ ss}} + \psi_4 \mathcal{L}_{\mathit{iou\ ss}},
\end{equation}
where $\psi_3$ and $\psi_4$ are hyper-parameters used to control the influence of each part of the function (\ie, weighted sum).
Keep in mind that the semantic contour task uses the same loss functions $\mathcal{L}_S$ with the name $\mathcal{L}_C$ but with hyper-parameters $\omega$.

\subsection{Energy Level Training}
\label{sec:apx_train_el}
In the last specific decodification stage, we learn to classifier the bins of each level of the truncated distance transform.
In other words, train the network (with $\varphi$ parameters) to learn how to classify each level of the discretized distance transform (\ie, $K$ bins classifier).
Thus, similar to the previous section, we use  multi-label balanced cross-entropy loss function,
\begin{equation} \label{eq:cross_loss_e}
\mathcal{L_\mathit{cross\ e}} = -\frac{1}{K}\sum_{i=1}^K \mu_i \log P(k = k_i \given X; \varphi),
\end{equation}
and multi-label intersection-over-union loss function, 
\begin{equation}\label{eq:multi_iou_e}
\mathcal{L}_{\mathit{iou\ e}} = 1 - \sum_{i=1}^K\frac{P_i \cap K_i}{P_i \cup K_i},
\end{equation}
on a set of $K$ bins.

Finally, we merge our loss functions for energy level by,
\begin{equation} \label{eq:e_loss}
\mathcal{L}_D = \psi_5 \mathcal{L}_{\mathit{cross\ e}} + \psi_6 \mathcal{L}_{\mathit{iou\ e}},
\end{equation}
where $\psi_5$ and $\psi_6$ are hyper-parameters that define the contribution of each loss to the learning process.

\section{More Visualizations of Latent Space Behavior}
\label{sec:apx_visualization_LS}
\begin{figure*}[tb]
  \centering
  \setlength{\wsz}{2.2in}
  \setlength{\hsz}{1.1in}
  {\captionsetup{justification=centering}
  \subfloat[S \protect\\ $\text{SSI}=0.384$, $\text{DBI}=1.360$, $\text{DBI}=1.360$]{\includegraphics[width=\wsz,height=\hsz]{latent0-S}%
    \label{fig:latent0-S}}
  \hfil
  \subfloat[S+E \protect\\ $\text{SSI}=0.391$, $\text{DBI}=1.141$, $\text{DBI}=1.141$]{\includegraphics[width=\wsz,height=\hsz]{latent0-BS}%
    \label{fig:latent0-SB}}
    \hfil
    \subfloat[S+D \protect\\ $\text{SSI}=0.394$, $\text{DBI}=1.275$, $\text{DBI}=1.275$]{\includegraphics[width=\wsz,height=\hsz]{latent0-SE}%
      \label{fig:latent0-SE}}
  \vfil
  \subfloat[S+E+C \protect\\ $\text{SSI}=0.437$, $\text{DBI}=1.150$, $\text{DBI}=1.149$\label{fig:latent0-SBC}]{\includegraphics[width=\wsz,height=\hsz]{latent0-BCS}}
    \hfil
    \subfloat[S+E+D \protect\\ $\text{SSI}=0.444$, $\text{DBI}=0.920$, $\text{DBI}=0.920$]{\includegraphics[width=\wsz,height=\hsz]{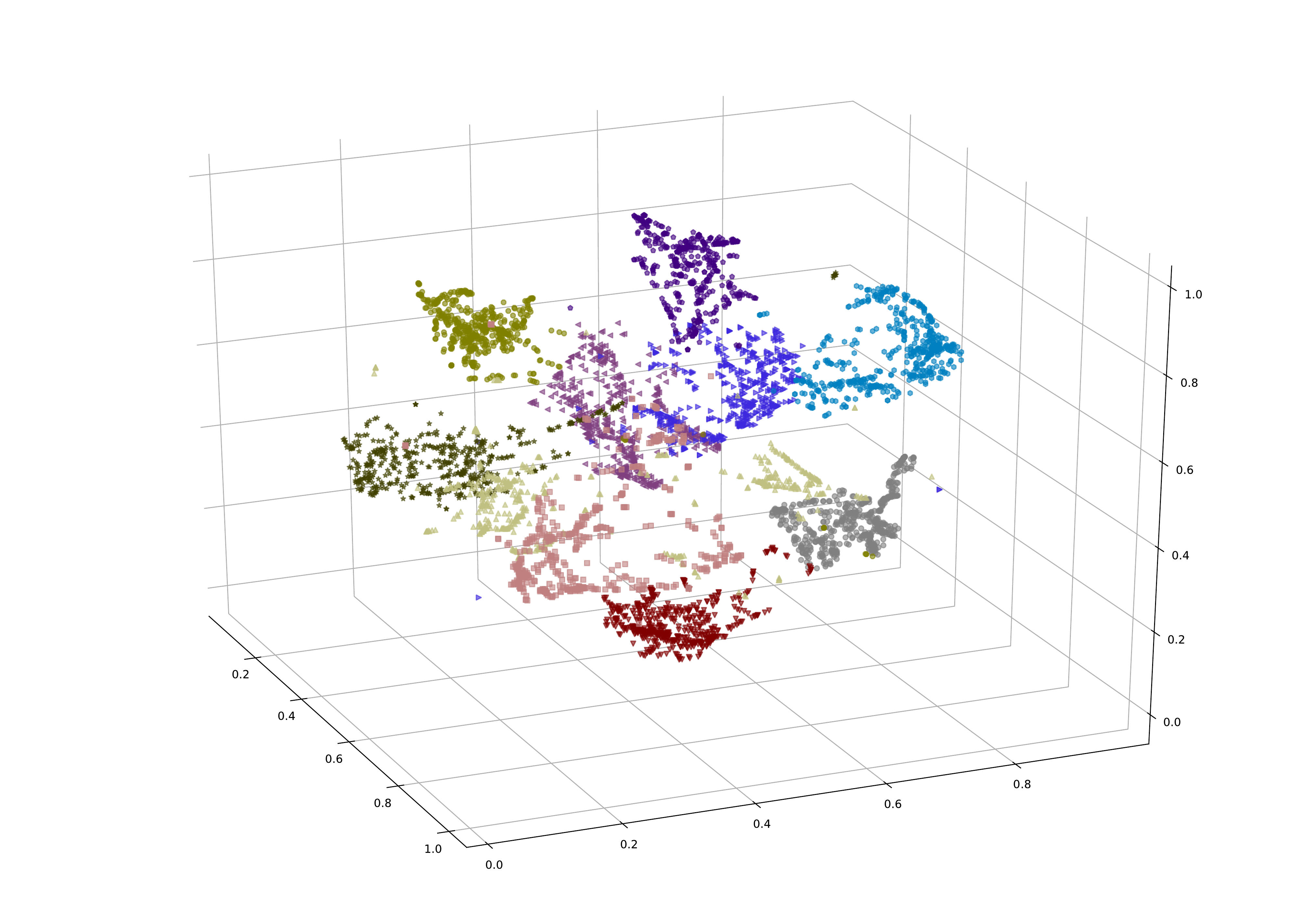}%
      \label{fig:latent0-SBE}}
  \hfil
  \subfloat[S+E+C+D \protect\\ $\text{SSI}=0.636$, $\text{DBI}=0.774$, $\text{DBI}=0.774$]{\includegraphics[width=\wsz,height=\hsz]{latent0-BCSE}%
    \label{fig:latent0-SBCE}}
  }
  \caption[Apx plotting the latent space ]{
    We are plotting of the shared latent space on Camvid testing dataset. Here we combine the different tasks of edge detection~(E), semantic segmentation~(S), semantic contour~(C), and distance transform~(D).
    Note that when adding tasks related to semantic segmentation, \ie, by providing complementary information, maps of similar features (within a multi-task hourglass model) are clustered together in a similar latent space, and they are not spaced arbitrarily.
    We confirm this behavior by using a set of metrics for clustering shown in Table~\ref{tab:ablation2}.
  }
  \label{fig:latent0-space-tasks}
  \vspace*{-10pt}
\end{figure*}

\begin{figure*}[tb]
\centering
\setlength{\wsz}{2.2in}
\setlength{\hsz}{1.1in}
  {\captionsetup{justification=centering}
  \subfloat[S\protect\\ $\text{SSI}=0.325$, $\text{CHI}=1468.3$, $\text{DBI}=1.572$]{\includegraphics[width=\wsz,height=\hsz]{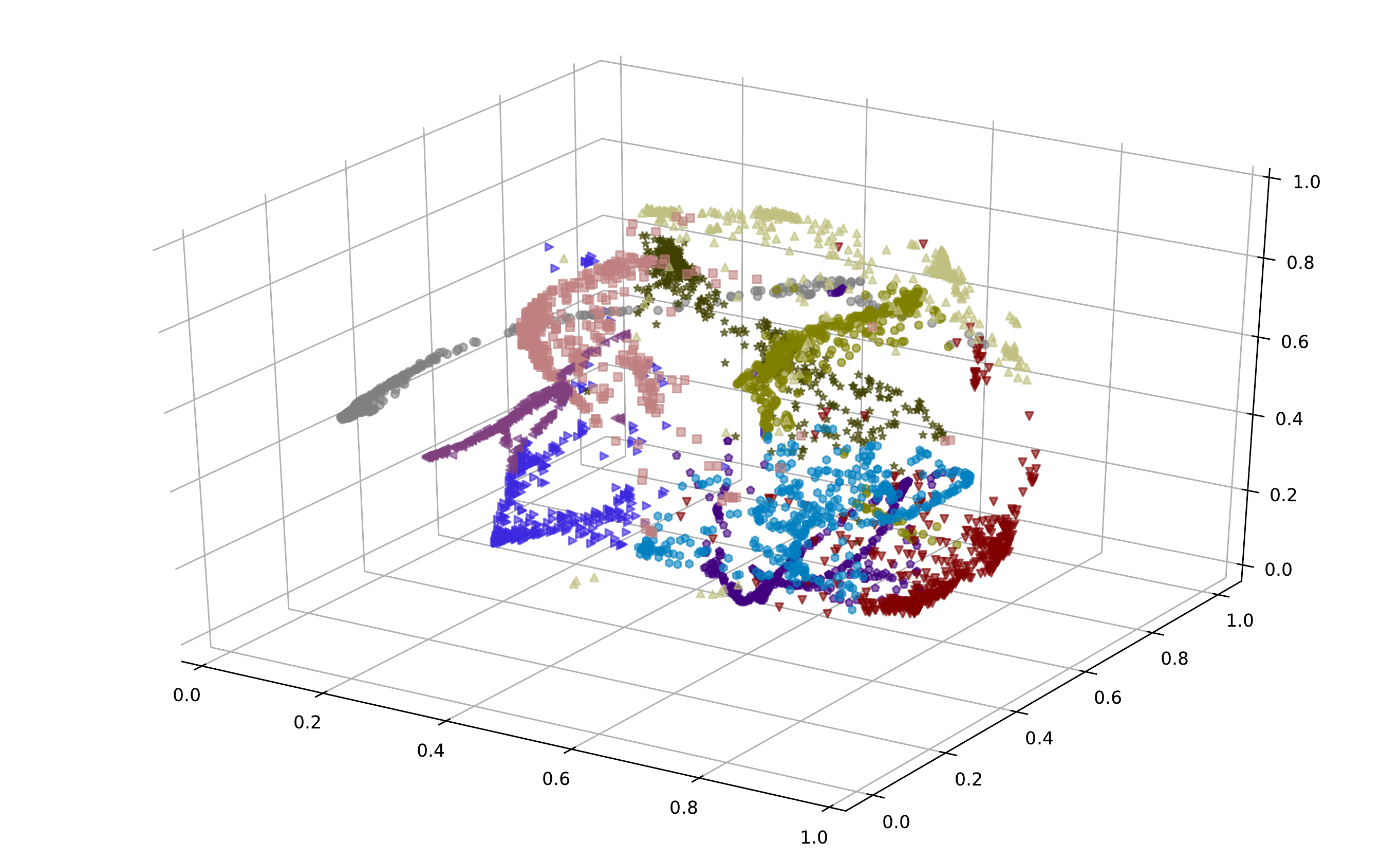}%
    \label{fig:latent1-S}}
  \hfil
  \subfloat[S+E \protect\\ $\text{SSI}=0.342$, $\text{CHI}=1678.9$, $\text{DBI}=1.114$]{\includegraphics[width=\wsz,height=\hsz]{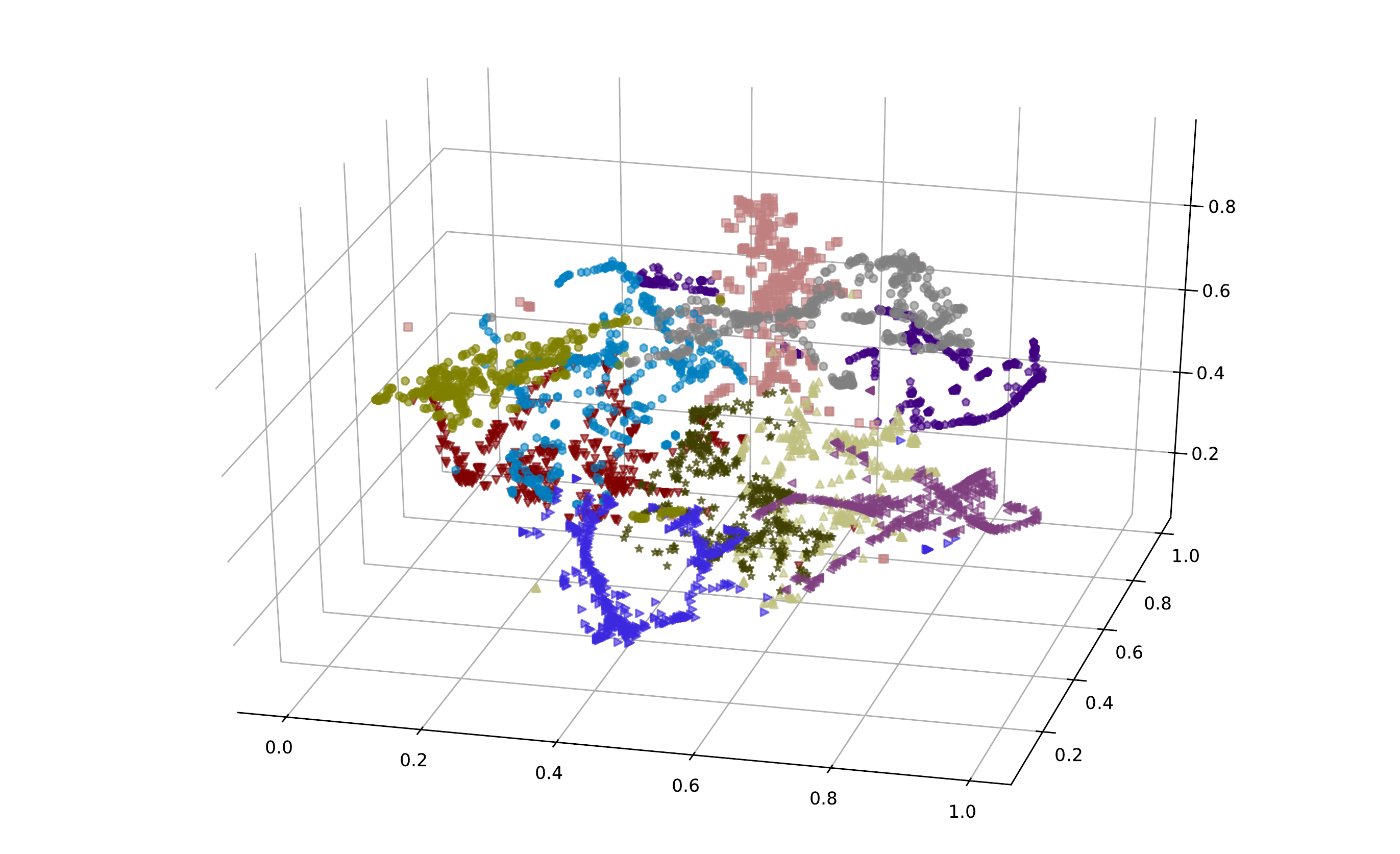}%
    \label{fig:latent1-SB}}
  \hfil
  \subfloat[S+D \protect\\ $\text{SSI}=0.365$, $\text{CHI}=1660.9$, $\text{DBI}=1.275$]{\includegraphics[width=\wsz,height=\hsz]{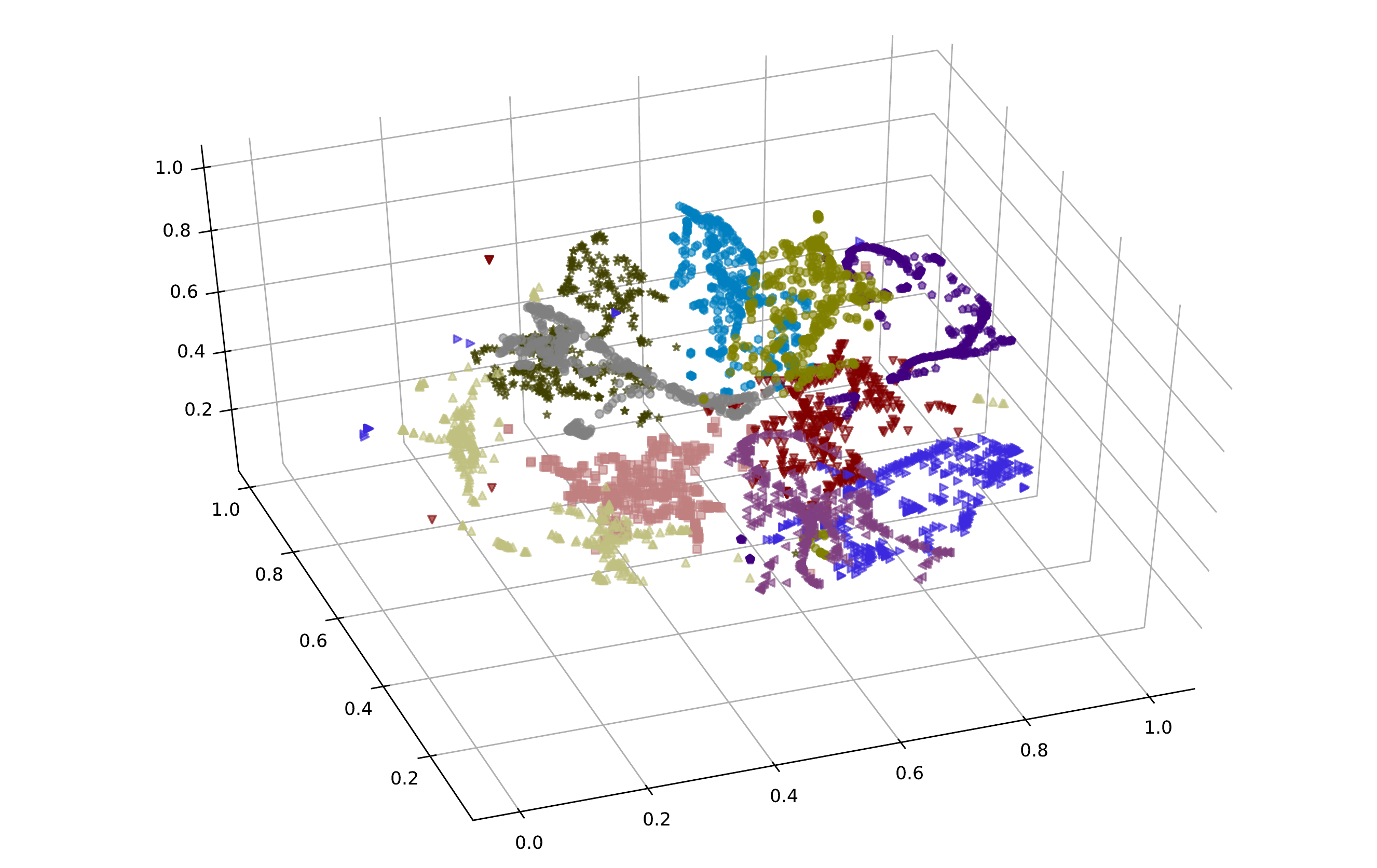}%
    \label{fig:latent1-SE}}
  \vfil
  \subfloat[S+E+C \protect\\ $\text{SSI}=0.413$, $\text{CHI}=1806.5$, $\text{DBI}=0.966$]{\includegraphics[width=\wsz,height=\hsz]{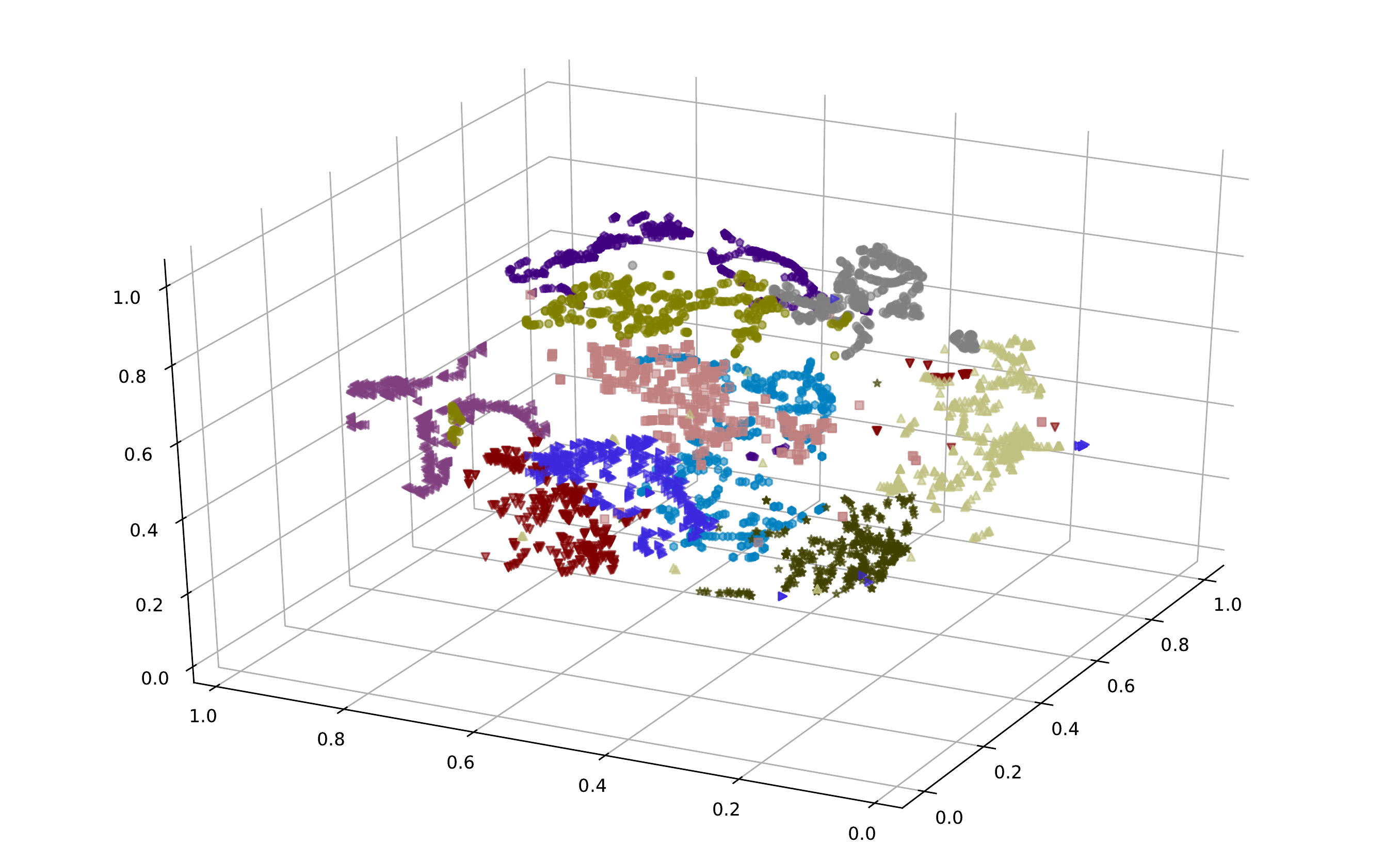}%
    \label{fig:latent1-SBC}}
  \hfil
  \subfloat[S+E+D \protect\\ $\text{SSI}=0.508$, $\text{CHI}=2578.5$, $\text{DBI}=0.922$]{\includegraphics[width=\wsz,height=\hsz]{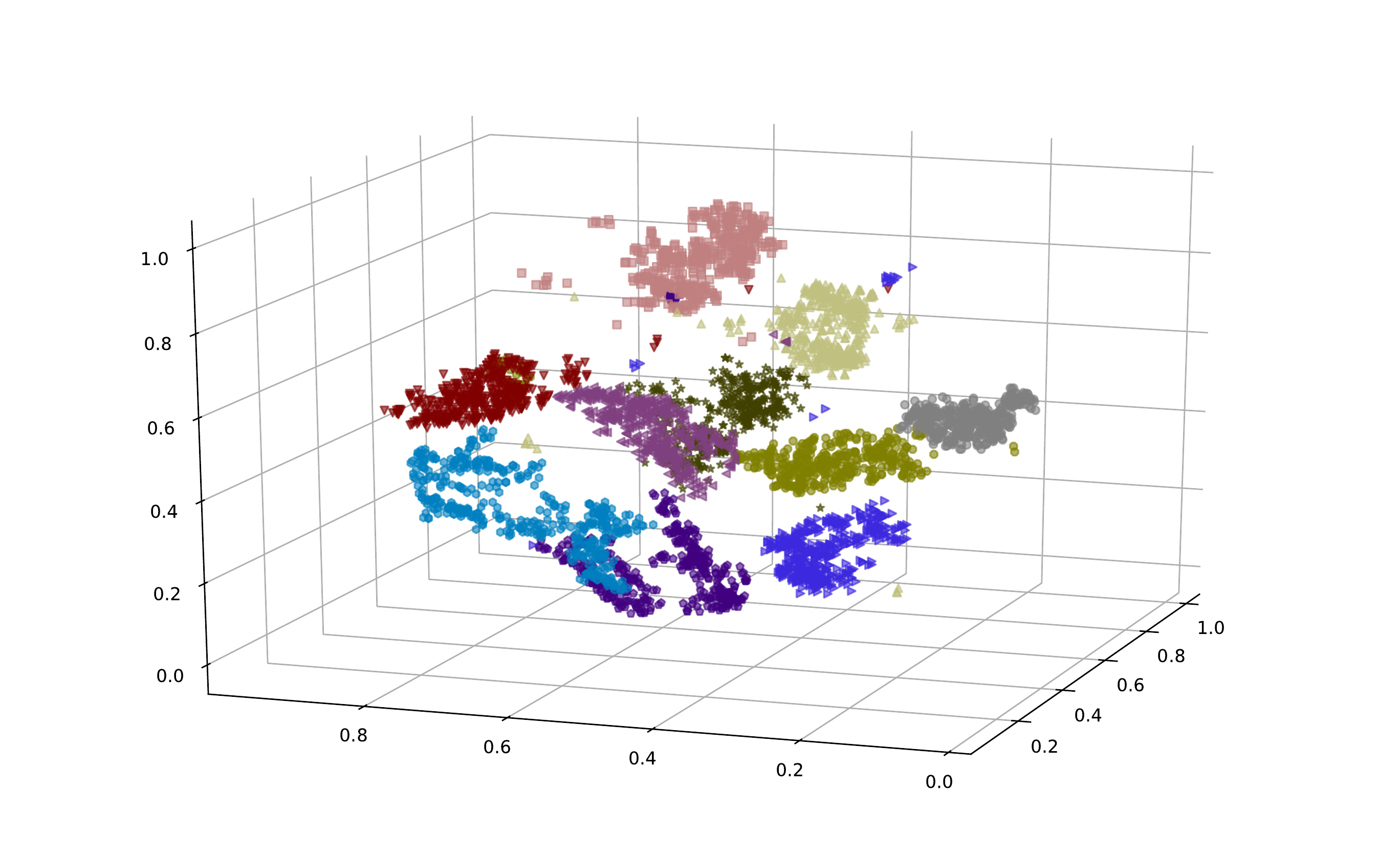}%
    \label{fig:latent1-SBE}}
  \hfil
  \subfloat[S+E+C+D \protect\\ $\text{SSI}=0.510$, $\text{CHI}=2795.3$, $\text{DBI}=0.776$]{\includegraphics[width=\wsz,height=\hsz]{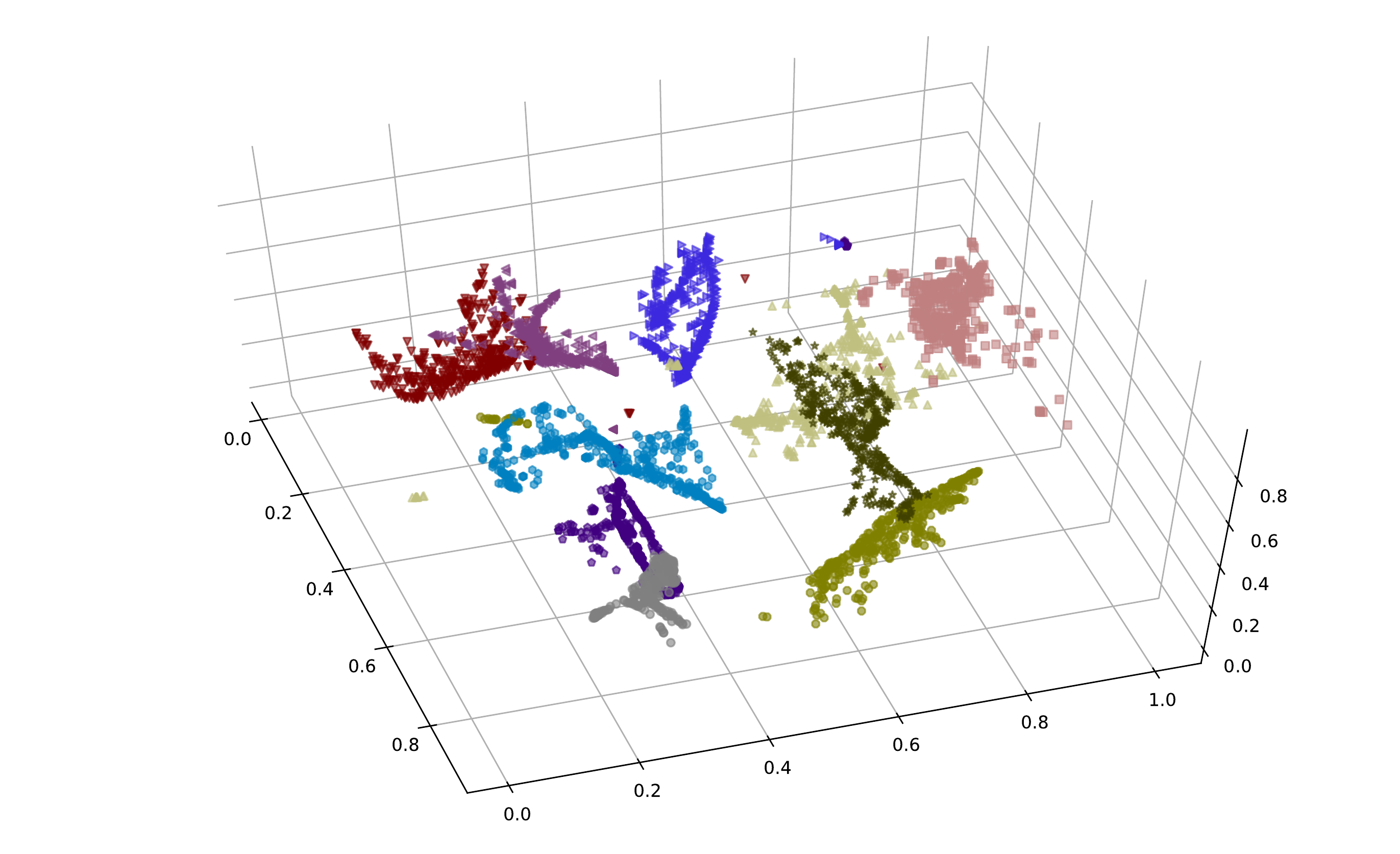}%
    \label{fig:latent1-SBCE}}
  }
\caption[Additional plotting the latent space ]{
  Additional results of the shared latent space for the dataset subset (random labeled-pixels sample on CamVid testing dataset). Merging tasks of edge detection~(E), semantic segmentation~(S), semantic contour~(C), and distance transform~(D).
}
\label{fig:latent1-space-tasks}
\vspace*{-10pt}
\end{figure*}

In this section, we show complementary visualizations in Fig.~\ref{fig:latent0-space-tasks} of the main document Fig.~\ref{fig:latent-space-tasks}.
We present additional plotting of the behavior visualizations of the latent space on a subset of the CamVid dataset (in Fig.~\ref{fig:latent1-space-tasks}). 
We build the subset with $10$ random image samples. 
From these images, we selected $446$ random label-pixels for each class (all on the testing set).

\section{Architectures}
\label{sec:apx_architectures}
The architectures for the semantic segmentation models used in this paper are the same as those used in the original papers. 
We keep the same number of convolution and deconvolution layers for the encoding and decoding stages. 
We maintain the same amount of hidden units for each, that is, channels per layer, and we maintain the same non-linear activation functions and hyperparameters.
The setup of the hourglass models is defined in their respective papers for the architectures we used, namely, FCN8~\cite{Long2016}, ParseNet~\cite{Liu2015}, SegNet~\cite{Badrinarayanan2017}, FastNet~\cite{Oliveira2016}, UNet~\cite{Ronneberger2015}, DeconvNet~\cite{Noh2015}, AdapNet++~\cite{Valada2019}, CGBNet~\cite{Ding2020}, FC-DenseNet67~\cite{Jegou2017}, and ENet~\cite{Paszke2016}.

Finally, for each specific task-block, we use two capable of convolution with kernels of $1\times1$ and $8\times8$, respectively, both with depth (channels) of the same number of classes of the dataset for tasks $S$ and $C$, depth of $1$ for task $B$ and depth of $6$ for task $D$.